\pdfoutput=1
\documentclass[11pt,letterpaper]{article}
\usepackage[T1]{fontenc}
\usepackage[utf8]{inputenc}

\usepackage{amsmath,amssymb,amsthm,mathtools}
\usepackage{newpxtext}                 
\usepackage{newpxmath}                 
\usepackage[scaled=0.90]{FiraSans}     

\usepackage[letterspace=80]{microtype}

\usepackage[letterpaper,left=1.15in,right=1.15in,top=1.1in,bottom=1.15in,headsep=0.32in,footskip=0.45in]{geometry}
\usepackage{xcolor}
\definecolor{ink}{HTML}{1A1F2B}
\definecolor{navy}{HTML}{1F3A63}         
\definecolor{refcool}{HTML}{2C5AA0}      
\definecolor{citewarm}{HTML}{B5403B}     
\definecolor{muted}{HTML}{6B6A66}        
\definecolor{rulegray}{HTML}{D3D6DC}
\definecolor{thmback}{HTML}{F2F6FC}
\definecolor{lemback}{HTML}{F5F6F8}
\definecolor{lembar}{HTML}{6B7A90}
\definecolor{absback}{HTML}{F7F8FA}

\usepackage{graphicx,booktabs,array,tabularx,multicol}
\usepackage{enumitem}
\setlist{itemsep=1pt,parsep=0pt,topsep=3pt,partopsep=0pt}
\usepackage[font=small,labelfont={sf,bf,color=navy},labelsep=period,skip=6pt]{caption}
\usepackage[numbers,square,sort&compress]{natbib}

\usepackage[most]{tcolorbox}
\usepackage{titlesec,titletoc}
\usepackage{fancyhdr}
\usepackage{url}
\usepackage{amsfonts,bm}
\usepackage{nicefrac}
\usepackage{multirow}
\usepackage{makecell}
\usepackage{capt-of}
\usepackage{placeins}
\usepackage{afterpage}
\usepackage{tikz}
\usepackage{pgfplots}
\usepackage{pgfplotstable}
\usepackage{colortbl}
\usepackage{aliascnt}
\usepackage{pifont}

\titleformat{\section}{\sffamily\Large\bfseries\color{navy}}{\textcolor{refcool}{\thesection}}{0.7em}{}
\titlespacing*{\section}{0pt}{3.2ex plus 1ex minus .2ex}{1.5ex plus .2ex}
\titleformat{\subsection}{\sffamily\large\bfseries\color{navy}}{\textcolor{refcool}{\thesubsection}}{0.6em}{}
\titlespacing*{\subsection}{0pt}{2.4ex plus .8ex minus .2ex}{1.0ex plus .2ex}
\titleformat{\subsubsection}{\sffamily\normalsize\bfseries\color{navy}}{\textcolor{refcool}{\thesubsubsection}}{0.5em}{}
\titleformat{\paragraph}[runin]{\sffamily\bfseries\color{navy}}{}{0pt}{}
\titlespacing*{\paragraph}{0pt}{1.4ex plus .5ex minus .2ex}{0.55em}

\titlecontents{section}[2.0em]{\addvspace{0.6pt}\sffamily\fontsize{8.4}{10.2}\selectfont}
  {\contentslabel[\textcolor{refcool}{\bfseries\thecontentslabel}]{2.0em}}{\hspace*{-2.0em}}
  {\titlerule*[0.6pc]{\textcolor{rulegray}{.}}\contentspage}

\fancypagestyle{first}{\fancyhf{}\fancyfoot[C]{\sffamily\footnotesize\color{muted}\thepage}}

\newtheoremstyle{boxed}{0pt}{0pt}{\itshape}{}{\sffamily\bfseries\color{navy}}{.}{0.5em}
  {\thmname{#1}\thmnumber{ #2}\thmnote{\ \textmd{(#3)}}}
\newtheoremstyle{boxedrm}{0pt}{0pt}{\normalfont}{}{\sffamily\bfseries\color{navy}}{.}{0.5em}
  {\thmname{#1}\thmnumber{ #2}\thmnote{\ \textmd{(#3)}}}
\theoremstyle{boxed}
\newtheorem{theorem}{Theorem}
\newtheorem{proposition}[theorem]{Proposition}
\newtheorem{lemma}[theorem]{Lemma}
\newtheorem{corollary}[theorem]{Corollary}

\theoremstyle{boxedrm}
\newtheorem{definition}{Definition}

\newtheorem{remark}[theorem]{Remark}

\tcbset{thmbox/.style={enhanced,breakable,frame hidden,sharp corners,boxrule=0pt,
  left=8pt,right=8pt,top=5pt,bottom=5pt,before skip=9pt,after skip=9pt}}
\tcbset{coolbox/.style={thmbox,colback=thmback,borderline west={2.4pt}{0pt}{refcool}}}
\tcbset{slatebox/.style={thmbox,colback=lemback,borderline west={2.4pt}{0pt}{lembar}}}
\tcbset{warmbox/.style={thmbox,colback=white,borderline west={2.4pt}{0pt}{citewarm}}}
\tcbset{linebox/.style={blanker,breakable,left=8pt,before skip=7pt,after skip=7pt,borderline west={1pt}{0pt}{rulegray}}}
\tcolorboxenvironment{theorem}{coolbox}
\tcolorboxenvironment{proposition}{coolbox}
\tcolorboxenvironment{corollary}{coolbox}
\tcolorboxenvironment{conjecture}{coolbox}
\tcolorboxenvironment{lemma}{slatebox}
\tcolorboxenvironment{claim}{slatebox}
\tcolorboxenvironment{fact}{slatebox}
\tcolorboxenvironment{observation}{slatebox}
\tcolorboxenvironment{definition}{warmbox}
\tcolorboxenvironment{assumption}{warmbox}
\tcolorboxenvironment{remark}{linebox}
\tcolorboxenvironment{example}{linebox}
\tcolorboxenvironment{problem}{linebox}
\tcolorboxenvironment{question}{linebox}

\makeatletter
\renewenvironment{proof}[1][\proofname]{\par\pushQED{\qed}\normalfont\topsep6\p@\@plus6\p@\relax
  \trivlist\item[\hskip\labelsep{\sffamily\bfseries\color{navy}#1.}]\ignorespaces}{\popQED\endtrivlist\@endpefalse}
\makeatother

\makeatletter
\newif\ifwithchecklist
\usetikzlibrary{arrows.meta,positioning,calc,fit,backgrounds,
  decorations.pathreplacing,shapes.geometric,patterns}
\usepgfplotslibrary{groupplots,fillbetween}
\pgfplotsset{compat=1.18}
\definecolor{CoolInk}{HTML}{1C63B7}
\definecolor{CoolMid}{HTML}{79A6DC}
\definecolor{CoolFill}{HTML}{E9F0FA}
\definecolor{WarmInk}{HTML}{D95F02}
\definecolor{WarmMid}{HTML}{F2A46B}
\definecolor{WarmFill}{HTML}{FDEEE3}
\definecolor{SageInk}{HTML}{1B9E77}
\definecolor{SageMid}{HTML}{85CDB6}
\definecolor{SageFill}{HTML}{E3F4EE}
\definecolor{Ink}{HTML}{222222}
\definecolor{MidGray}{HTML}{7A7A7A}
\definecolor{LightGray}{HTML}{C4C4C4}
\definecolor{GridGray}{HTML}{EBEBEB}
\definecolor{FaintGray}{HTML}{F3F3F3}
\definecolor{Paper}{HTML}{F7F3EA}
\tikzset{
  figbase/.style={line width=0.55pt, line cap=round, line join=round,
    >={Stealth[length=1.9mm,width=1.4mm]}, font=\footnotesize,
    execute at begin picture={%
      \def\footnotesize{\fontsize{8}{9.5}\selectfont}%
      \def\scriptsize{\fontsize{7}{8.2}\selectfont}%
      \def\tiny{\fontsize{6.3}{7.3}\selectfont}}},
  box/.style={draw=Ink, rounded corners=2pt, inner xsep=3.5pt, inner ysep=2.5pt,
    align=center, font=\footnotesize},
  coolbox/.style={box, draw=CoolInk, fill=CoolFill},
  warmbox/.style={box, draw=WarmInk, fill=WarmFill},
  sagebox/.style={box, draw=SageInk, fill=SageFill},
  greybox/.style={box, draw=LightGray, fill=FaintGray},
  keybox/.style={box, draw=Ink, fill=Paper, line width=0.9pt, inner xsep=5pt, inner ysep=4pt},
  flow/.style={->, draw=MidGray, line width=0.65pt},
  coolflow/.style={->, draw=CoolInk, line width=0.65pt},
  warmflow/.style={->, draw=WarmInk, line width=0.65pt},
  lab/.style={font=\scriptsize\itshape, text=MidGray},
  tag/.style={font=\scriptsize, text=Ink},
  paneltitle/.style={font=\footnotesize, text=Ink, anchor=west, inner sep=0pt},
  ringdot/.style={circle, inner sep=0pt, minimum size=#1, draw=white, line width=0.7pt},
  ringdot/.default=4.6pt,
}
\pgfplotsset{
  paperplot/.style={
    axis lines=left,
    axis line style={-, draw=MidGray, line width=0.45pt},
    tick style={draw=MidGray, line width=0.4pt},
    major tick length=2.2pt, minor tick length=1.3pt,
    tick align=outside, xtick pos=bottom, ytick pos=left,
    every tick label/.append style={font=\scriptsize, text=Ink!85},
    label style={font=\scriptsize, text=Ink},
    title style={font=\footnotesize, text=Ink, at={(0,1)}, anchor=south west,
      xshift=-2pt, yshift=3pt, align=left},
    legend style={font=\scriptsize, draw=none, fill=none, inner sep=1pt, row sep=-1pt},
    legend cell align=left,
    grid style={draw=GridGray, line width=0.35pt},
    scaled ticks=false,
    clip=false,
  },
  ygridplot/.style={paperplot, ymajorgrids},
  xgridplot/.style={paperplot, xmajorgrids},
}
\newcommand{\ptitle}[2]{\textbf{(#1)}\enspace#2}

\definecolor{RefCool}{HTML}{1C63B7}
\definecolor{CiteWarm}{HTML}{A94F17}
\tcbset{paperstatement/.style={unbreakable, colback=black!7, colframe=white,
  width=\dimexpr\linewidth+8pt\relax, enlarge left by=-4pt, enlarge right by=-4pt,
  boxsep=4pt, boxrule=0pt, left=0pt, right=0pt, top=0pt, bottom=0pt,
  sharp corners, before skip=6pt, after skip=6pt}}
\definecolor{TakeFrame}{HTML}{8E8E8E}
\definecolor{TakeBack}{HTML}{F5F5F3}
\newtcolorbox{takeaway}[1][Takeaway]{enhanced, colback=TakeBack, colframe=TakeFrame,
  boxrule=0.6pt, arc=4pt, outer arc=4pt, left=7pt, right=7pt, top=10pt, bottom=4pt,
  before skip=10pt, after skip=6pt,
  title={#1}, fonttitle=\scshape, coltitle=black!78,
  attach boxed title to top left={xshift=10pt, yshift=-\tcboxedtitleheight/2},
  boxed title style={colback=white, colframe=TakeFrame, boxrule=0.6pt, arc=3pt,
    left=4pt, right=4pt, top=1pt, bottom=1pt}}
\definecolor{HeadBand}{gray}{0.90}
\newcommand{\headtop}{\specialrule{\heavyrulewidth}{0pt}{0pt}\rowcolor{HeadBand}}
\newcommand{\headstrut}{\rule[-1.1ex]{0pt}{3.4ex}}
\newcommand{\headbottom}{\specialrule{\lightrulewidth}{0pt}{\belowrulesep}}
\newcommand{\E}{\mathbb{E}}
\renewcommand{\P}{\mathbb{P}}
\newcommand{\R}{\mathbb{R}}
\newcommand{\N}{\mathbb{N}}
\newcommand{\cA}{\mathcal{A}}
\newcommand{\cD}{\mathcal{D}}
\newcommand{\cF}{\mathcal{F}}
\newcommand{\cJ}{\mathcal{J}}

\newcommand{\one}{\mathbf{1}}
\newcommand{\eps}{\varepsilon}
\newcommand{\KL}{\operatorname{KL}}
\newcommand{\kl}{\operatorname{kl}}
\newcommand{\klinv}{\kl^{-1}_{+}}
\newcommand{\klinvlow}{\kl^{-1}_{-}}
\newcommand{\Lhat}{\widehat{L}}
\newcommand{\Ghat}{\widehat{G}}
\newcommand{\SC}{\mathrm{SC}}
\newcommand{\Cmax}{C_{\max}}

\newcommand{\pen}[1]{\log\!\bigl(2\sqrt{#1}\,#1(#1+1)/\delta\bigr)}
\newcommand{\penx}[2]{\log\!\bigl(2\sqrt{#1}\,#1(#1+1)/#2\bigr)}

\newcommand{\nGenQwen}{49{,}923}
\newcommand{\nGenGpt}{38{,}871}
\newcommand{\nGenTotal}{88{,}794}
\newcommand{\selGG}{5}
\newcommand{\calGG}{0.039}
\newcommand{\heldGG}{0.039}
\newcommand{\certGG}{0.109}

\newcommand{\ratioCalGG}{2.8}
\newcommand{\MGG}{4}
\newcommand{\ncalGG}{659}
\newcommand{\selQG}{17}
\newcommand{\calQG}{0.062}
\newcommand{\heldQG}{0.073}
\newcommand{\certQG}{0.144}

\newcommand{\ratioCalQG}{2.3}
\newcommand{\MQG}{7}
\newcommand{\ncalQG}{659}
\newcommand{\selGM}{5}
\newcommand{\calGM}{0.280}
\newcommand{\heldGM}{0.296}
\newcommand{\certGM}{0.470}

\newcommand{\ratioCalGM}{1.7}
\newcommand{\MGM}{4}
\newcommand{\ncalGM}{250}
\newcommand{\selQM}{9}
\newcommand{\calQM}{0.300}
\newcommand{\heldQM}{0.268}
\newcommand{\certQM}{0.491}

\newcommand{\MQM}{4}
\newcommand{\ncalQM}{250}
\newcommand{\certPilot}{0.219}
\newcommand{\heldPilot}{0.010}
\newcommand{\calPilot}{0.021}
\newcommand{\synthFixedCovMax}{0.388}
\newcommand{\synthMmax}{128}
\newcommand{\synthFixedCovMin}{0.986}
\newcommand{\priorUniCheap}{0.073}
\newcommand{\priorCplxCheap}{0.067}

\newcommand{\frontQGesCostShare}{21}
\newcommand{\nstarFoura}{8{,}917}
\newcommand{\nstarFourb}{4{,}076}
\newcommand{\nstarFourc}{1{,}387}
\newcommand{\nstarFourd}{571}
\newcommand{\nstarFoure}{201}
\newcommand{\nstarTwentyFoura}{9{,}546}
\newcommand{\nstarTwentyFourb}{4{,}388}
\newcommand{\nstarTwentyFourc}{1{,}507}
\newcommand{\nstarTwentyFourd}{626}
\newcommand{\nstarTwentyFoure}{223}

\newcommand{\nstarGrowMin}{7}
\newcommand{\nstarGrowMax}{11}
\newcommand{\gridWidthNats}{0.007}
\newcommand{\klTwoInv}{3.26}
\newcommand{\misWilsonGB}{8.7}

\newcommand{\misPBGB}{0.0}
\newcommand{\optimismGB}{0.005}
\newcommand{\MmisGB}{4}
\newcommand{\misWilsonQB}{8.1}

\newcommand{\misPBQB}{0.0}

\newcommand{\MmisQB}{5}
\newcommand{\misWilsonGGrid}{22.5}

\newcommand{\misPBGGrid}{0.0}

\newcommand{\MmisGGrid}{25}
\newcommand{\misWilsonQGrid}{21.6}

\newcommand{\misPBQGrid}{0.0}
\newcommand{\optimismQGrid}{0.020}
\newcommand{\MmisQGrid}{27}
\newcommand{\hoeffWidth}{0.10}
\newcommand{\gridSelQGrid}{P1/\allowbreak T1.0/\allowbreak b=5}
\newcommand{\gridCertQGrid}{0.198}

\newcommand{\gridHeldQGrid}{0.072}
\newcommand{\gridMQGrid}{27}

\newcommand{\gridSpreadQGrid}{0.006}
\newcommand{\gridSelGGrid}{P1/\allowbreak T1.0/\allowbreak b=9}
\newcommand{\gridCertGGrid}{0.134}

\newcommand{\gridMGGrid}{25}

\newcommand{\sepNQG}{935}

\newcommand{\sepNQM}{397}

\newcommand{\margSepA}{16{,}738}
\newcommand{\margSepQG}{21{,}273}
\newcommand{\margSepQM}{9{,}366}

\newcommand{\marPosQM}{70.8}
\newcommand{\marNegQM}{27.2}
\newcommand{\marTieQM}{2.0}

\newcommand{\marVotedOutQM}{6.6}
\newcommand{\marPosGM}{70.6}
\newcommand{\marNegGM}{28.4}
\newcommand{\marTieGM}{1.0}

\newcommand{\marVotedOutGM}{6.6}
\newcommand{\wfRiskFive}{0.080}

\newcommand{\uniRiskFive}{0.070}

\newcommand{\wfBudgetNine}{8.21}

\newcommand{\wfULUniNine}{0.150}

\newcommand{\wfULWfNine}{0.146}

\newcommand{\wfHeldWfNine}{0.073}

\newcommand{\wfTokSave}{6}
\newcommand{\elocMa}{337}
\newcommand{\elocFlatSamea}{0.203}
\newcommand{\elocFlatAudita}{0.203}
\newcommand{\elocFlatCala}{0.052}
\newcommand{\elocFlatHelda}{0.076}
\newcommand{\elocMb}{1{,}012}
\newcommand{\elocFlatSameb}{0.207}
\newcommand{\elocFlatAuditb}{0.201}
\newcommand{\elocFlatCalb}{0.048}
\newcommand{\elocFlatHeldb}{0.056}
\newcommand{\elocMc}{2{,}787}
\newcommand{\elocFlatSamec}{0.212}
\newcommand{\elocFlatAuditc}{0.198}
\newcommand{\elocFlatCalc}{0.044}
\newcommand{\elocFlatHeldc}{0.072}
\newcommand{\elocLoc}{0.199}
\newcommand{\elocMdefault}{67}
\newcommand{\tokMedB}{2242}
\newcommand{\tokPnnB}{3925}
\newcommand{\cmaxZero}{512}

\newcommand{\csCovMin}{0.000}
\newcommand{\csCovMax}{0.012}
\newcommand{\ebGG}{0.067}
\newcommand{\klFixedGG}{0.066}
\newcommand{\ebQG}{0.098}
\newcommand{\klFixedQG}{0.096}
\newcommand{\ebGM}{0.375}
\newcommand{\klFixedGM}{0.369}
\newcommand{\ebQM}{0.382}
\newcommand{\klFixedQM}{0.390}
\newcommand{\ctrlGB}{5.4}

\newcommand{\ctrlQB}{4.2}

\newcommand{\ctrlGGrid}{5.6}

\newcommand{\ctrlQGrid}{6.8}
\newcommand{\ctrlMaxQGrid}{9.7}
\newcommand{\mixCertQGrid}{0.193}
\newcommand{\mixKQGrid}{3}
\newcommand{\trajGGatNinetySix}{0.193}
\newcommand{\stopPathGG}{153}
\newcommand{\stopPathQG}{493}

\newcommand{\pickLamZero}{$b=17$}
\newcommand{\pickLamTenth}{ES(2,5)}
\newcommand{\pickLamOne}{$b=1$}
\newcommand{\wfErrWfNine}{42}
\newcommand{\wfErrUniNine}{44}
\newcommand{\apiCost}{14.28}
\newcommand{\apiTokM}{11.8}
\newcommand{\gpuMinMeasured}{43.5}

\newcommand{\nstarAtGG}{1{,}846}
\newcommand{\nstarAtQG}{1{,}088}
\newcommand{\timeTermGG}{16.9}
\newcommand{\timeUnifGG}{13.0}
\newcommand{\momentGG}{3.9}
\newcommand{\logMdeltaGG}{4.4}
\newcommand{\ctrlMeanMin}{4.2}
\newcommand{\ctrlMeanMax}{6.8}
\newcommand{\gridLocCertQGrid}{0.199}

\newcommand{\lnMedian}{545}
\newcommand{\lnMean}{898}
\newcommand{\lnPnn}{5{,}577}

\newcommand{\lnReps}{2{,}000}

\newcommand{\cmark}{\textcolor{CoolInk}{\ding{51}}}
\newcommand{\xmark}{\textcolor{WarmMid}{\ding{55}}}
\renewtcolorbox{takeaway}[1][Takeaway]{enhanced,breakable,frame hidden,sharp corners,boxrule=0pt,
  colback=absback,borderline west={2.4pt}{0pt}{citewarm},left=9pt,right=9pt,top=6pt,bottom=6pt,
  before skip=10pt,after skip=8pt,
  before upper={{\sffamily\bfseries\footnotesize\color{citewarm}\lsstyle\MakeUppercase{#1}}\par\vspace{3pt}}}
\definecolor{HeadBand}{HTML}{ECEFF3}
\makeatother

\usepackage[colorlinks=true,linkcolor=refcool,citecolor=citewarm,urlcolor=refcool,
  pdftitle={Anytime-Valid PAC-Bayes Certificates for Adaptive Test-Time Scaling},pdfauthor={Xiaoyu Li, Jiangxuan Long},pdfsubject={},pdfkeywords={}]{hyperref}
\usepackage[capitalize,nameinlink,noabbrev]{cleveref}

\crefname{equation}{Eq.}{Eqs.}
\Crefname{equation}{Equation}{Equations}
\crefname{appendix}{Appendix}{Appendices}
\Crefname{appendix}{Appendix}{Appendices}
\crefname{table}{Table}{Tables}
\crefname{figure}{Figure}{Figures}
\crefname{corollary}{Corollary}{Corollaries}
\crefname{proposition}{Proposition}{Propositions}
\crefname{lemma}{Lemma}{Lemmas}
\crefname{remark}{Remark}{Remarks}
\crefname{definition}{Definition}{Definitions}

\begin{document}
\thispagestyle{first}

\noindent{\color{refcool}\rule{\linewidth}{2.2pt}}\par\vspace{7pt}
\noindent{\sffamily\footnotesize\color{muted}\lsstyle NeurIPS 2026 $\cdot$ Extended Version\quad\textbullet\quad \ifcase\month\or January\or February\or March\or April\or May\or June\or July\or August\or September\or October\or November\or December\fi\ \number\year}\par\vspace{8pt}
\noindent{\sffamily\bfseries\fontsize{23}{28}\selectfont\color{navy}Anytime-Valid PAC-Bayes Certificates\\ for Adaptive Test-Time Scaling\par}\vspace{7pt}
\vspace{6pt}
\noindent\hspace*{0\linewidth}\begin{tabular*}{0.74\linewidth}{@{}l@{\extracolsep{\fill}}l@{}}
{\sffamily\bfseries\color{ink}Xiaoyu Li$^{*}$} & {\sffamily\bfseries\color{ink}Jiangxuan Long$^{*}$} \\[1pt]
{\small University of New South Wales} & {\small The University of Hong Kong} \\[1pt]
{\sffamily\footnotesize\href{mailto:xiaoyu.li2@unsw.edu.au}{\color{muted}xiaoyu.li2@unsw.edu.au}} & {\sffamily\footnotesize\href{mailto:jxlong@connect.hku.hk}{\color{muted}jxlong@connect.hku.hk}}
\end{tabular*}

\begingroup\renewcommand\thefootnote{}\footnotetext{$^{*}$Equal contribution. A preliminary version of this paper was accepted at the 40th Conference on Neural Information Processing Systems (NeurIPS 2026).}\endgroup

\vspace{5pt}

\begin{tcolorbox}[enhanced,colback=absback,frame hidden,sharp corners,boxrule=0pt,
  borderline west={3pt}{0pt}{refcool},left=11pt,right=11pt,top=6pt,bottom=6pt]
{\sffamily\bfseries\footnotesize\color{navy}\lsstyle ABSTRACT}\par\vspace{3pt}
\small Tuning a test-time scaling pipeline by sweeping prompts, budgets, verifiers, and stopping rules on a single labeled calibration set invalidates fixed-recipe error bars for the winner. Across 25-recipe sweeps of gpt-4.1-mini on GSM8K, a nominal 95\% Wilson upper bound for the selected \emph{recipe} falls below its held-out risk in \misWilsonGGrid\% of splits. Modeling each recipe as a stochastic predictor with random cost and a \emph{randomized recipe} as a posterior $Q$ over recipes, we establish an anytime-valid PAC-Bayes-kl certificate from i.i.d.\ calibration data relative to an independently declared prior $P$. The bound holds simultaneously across all sample sizes, posteriors $Q\ll P$, and data-dependent stopping times at selection penalty $\KL(Q\Vert P)$, accommodating hierarchical recipe families and dyadic tail bounds for uncapped token costs. This penalty is fundamental: under independent Bernoulli losses, valid per-recipe certificates obey a Kraft inequality whose expected kl-radius is lower-bounded by the Shannon entropy of the reported recipe up to $O(\log n)$, while even unrestricted certificates pay $\log M - O(\log n)$ over $M$ exchangeable recipes. Because plurality voting concentrates exponentially on the generator's mode whether right or wrong, certifying accuracy requires ground-truth labels; the same large-deviation rate yields an oracle water-filling rule for compute allocation. On GSM8K, gpt-4.1-mini over budgets $\{1,3,5,9\}$ achieves held-out risk \heldGG{} with a 95\% certificate of \certGG{}, and empirical coverage holds across all 12{,}000 resplits of four sweeps.\par

\end{tcolorbox}

\vspace{2pt}
\begin{tcolorbox}[enhanced,colback=white,colframe=rulegray,boxrule=0.5pt,arc=2pt,left=10pt,right=10pt,top=5pt,bottom=4pt]
{\sffamily\bfseries\footnotesize\color{navy}\lsstyle CONTENTS}\par
\setlength{\multicolsep}{4pt}
\begin{multicols}{2}
{\hypersetup{linkcolor=ink}\makeatletter\@starttoc{toc}\makeatother}
\end{multicols}
\end{tcolorbox}
\suppressfloats[t]
\clearpage

\suppressfloats[t]
\section{Introduction}\label{sec:intro}

Inference in modern language models is no longer a fixed forward pass; it is an engineered computation. By sampling multiple reasoning traces and aggregating by plurality, routing across budgets, halting early once a clear lead emerges, or scoring candidates with a verifier \citep{cobbe2021,wang2023,snell2024}, practitioners trade inference compute directly against accuracy and latency. Deploying such a system requires committing to a complete inference \emph{recipe}: prompts, decoding parameters, budget schedules, verifiers, parsers, and tie-breakers. In practice, engineers choose that recipe empirically, sweeping candidate configurations on a labeled calibration set and reporting whichever performs best.

The statistical difficulty is that the same calibration examples routinely serve both to select the winning recipe and to certify its accuracy. Across dozens of candidates, the empirical winner systematically harvests favorable sampling fluctuations; attaching a fixed-recipe confidence interval to that winner mistakes selection luck for genuine performance. On recorded \texttt{gpt-4.1-mini} outputs over 25 GSM8K recipes, a nominal 95\% Wilson upper bound for the selected recipe misses its held-out empirical risk in \misWilsonGGrid\% of splits (\cref{sec:miscoverage}). Sequential practice compounds the problem: practitioners rarely fix a calibration horizon in advance, instead inspecting learning curves example by example until performance stabilizes or a labeling budget is exhausted. The central question of this work is direct: \emph{how can we certify the true error rate and compute cost of a test-time scaling pipeline after arbitrary data-dependent tuning and optional stopping?}

\begin{figure}[t]
\centering
\begin{tikzpicture}[figbase, x=1pt, y=1pt,
  axis/.style={draw=MidGray, line width=0.45pt, -{Stealth[length=1.5mm, width=1.1mm]}},
  note/.style={font=\scriptsize, inner sep=1pt},
  node/.style={rounded corners=2pt, fill=white, font=\scriptsize, inner xsep=3.5pt, inner ysep=2.2pt},
  declare function={
    osc(\x,\b,\a,\p)=\b+\a*exp(-(\x-10)/28)*sin(deg((\x-10)/7+\p));
    sel(\x)=osc(\x,50,12,1.6)-9*exp(-((\x-84)/10)^2);
  }]
\pgfmathsetmacro{\ytau}{sel(84)}
\node[paneltitle] at (0,100) {\ptitle{a}{Peeking selects luck}};
\draw[axis] (6,16) -- (124,16);
\draw[axis] (6,16) -- (6,90);
\node[lab, anchor=north] at (121,14.5) {$n$};
\node[lab, anchor=south west] at (8,84) {risk};
\foreach \b/\a/\p in {58/12/0.4, 62/11/3.4, 55/13/2.3} {
  \draw[LightGray, line width=0.55pt, domain=10:84, samples=70, smooth] plot (\x, {osc(\x,\b,\a,\p)});
  \pgfmathsetmacro{\yend}{osc(84,\b,\a,\p)}
  \fill[MidGray!55] (84,\yend) circle (1.1pt);
}
\draw[WarmInk, line width=0.9pt, domain=10:84, samples=90, smooth] plot (\x, {sel(\x)});
\draw[Ink, line width=0.5pt, dash pattern=on 2.2pt off 1.6pt] (6,50) -- (124,50);
\node[note, text=Ink, anchor=south east] at (124,50.8) {true risk};
\draw[MidGray, line width=0.45pt, densely dotted] (84,16) -- (84,64);
\draw[MidGray, line width=0.45pt] (84,16) -- (84,13.8);
\node[note, text=Ink, anchor=north] at (84,13.5) {$\tau$};
\draw[MidGray, line width=0.4pt, densely dotted] (85.5,\ytau) -- (94,\ytau);
\fill[WarmInk] (84,\ytau) circle (1.45pt);
\draw[WarmInk, line width=1.2pt] (89,\ytau-7) -- (89,\ytau+7);
\draw[WarmInk, line width=0.6pt] (87,\ytau-7) -- (91,\ytau-7) (87,\ytau+7) -- (91,\ytau+7);
\draw[CoolInk, line width=1.2pt] (96,\ytau) -- (96,66);
\draw[CoolInk, line width=0.6pt] (94,\ytau) -- (98,\ytau) (94,66) -- (98,66);
\node[note, text=WarmInk, anchor=north] at (89,\ytau-8.5) {error bar \ding{55}};
\node[note, text=CoolInk, anchor=south] at (96,67) {certificate \ding{51}};
\begin{scope}[xshift=136pt]
\node[paneltitle] at (0,100) {\ptitle{b}{Every selection and stop}};
\fill[CoolFill] (14,16) rectangle (124,84);
\foreach \i in {0,...,11} \foreach \k in {0,...,7}
  \fill[CoolMid!65] ({18.6+9*\i},{20.3+9*\k}) circle (0.4pt);
\draw[CoolInk, line width=0.6pt] (14,16) rectangle (124,84);
\node[note, text=CoolInk, anchor=south east] at (124,85.5) {$\P(\mathcal{E}_\delta)\ge1-\delta$};
\node[lab, anchor=north] at (69,14.5) {calibration time $n$};
\node[lab, rotate=90, anchor=south] at (12.5,50) {posterior $Q$};
\draw[Ink, line width=0.6pt] (34,16) -- (34,84);
\node[note, text=Ink, fill=CoolFill, anchor=north west] at (35.5,82.5) {fixed $n$};
\draw[Ink, line width=0.6pt] (14,28) -- (124,28);
\node[note, text=Ink, fill=CoolFill, anchor=south east] at (122.5,18.0) {fixed recipe};
\draw[WarmInk, line width=1pt] (14,46) -- (52,46) -- (52,68) -- (84,68) -- (84,56) -- (106,56);
\node[star, star points=5, star point ratio=2.25, fill=WarmInk, inner sep=0pt, minimum size=6.2pt] at (106,56) {};
\node[note, text=WarmInk, fill=CoolFill, anchor=north, align=center] at (88,53.2) {selection-and-stopping\\rule $(\tau,Q_\tau)$};
\end{scope}
\begin{scope}[xshift=272pt]
\node[paneltitle] at (0,100) {\ptitle{c}{Selection penalty}};
\node[node, draw=Ink, line width=0.6pt] (root) at (64,85) {prior $P$};
\node[node, draw=LightGray, line width=0.6pt, text=MidGray] (j1) at (18,56) {$j{=}1$};
\node[node, draw=CoolInk, line width=0.7pt, fill=CoolFill, text=CoolInk] (j2) at (64,56) {$j{=}2$};
\node[node, draw=LightGray, line width=0.6pt, text=MidGray] (j3) at (110,56) {$j{=}3$};
\draw[LightGray, line width=0.6pt] (18,68) -- (110,68);
\draw[->, LightGray, line width=0.6pt] (18,68) -- (j1.north);
\draw[->, LightGray, line width=0.6pt] (110,68) -- (j3.north);
\draw[->, CoolInk, line width=0.9pt] (root.south) -- (j2.north);
\node[note, text=CoolInk, anchor=west] at (65.5,73.6) {$\log(1/\pi_j)$};
\foreach \x/\h in {48/4, 56/11, 64/16, 72/8, 80/3} {\fill[CoolMid] (\x-3.1,28) rectangle (\x+3.1,28+\h);}
\draw[MidGray, line width=0.45pt] (42,28) -- (86,28);
\draw[Ink, line width=0.5pt, dash pattern=on 2.2pt off 1.6pt] (42,36.4) -- (84.6,36.4);
\node[note, text=Ink, anchor=east] at (41.5,36.4) {$P_j$};
\node[note, text=white, anchor=south] at (64,28.4) {$Q$};
\draw[->, CoolInk, line width=0.6pt] (j2.south) -- (64,46);
\node[note, text=CoolInk, anchor=west] at (86.5,37.5) {$\KL(Q\Vert P_j)$};
\node[font=\scriptsize, text=Ink, fill=Paper, rounded corners=2pt, inner xsep=4pt, inner ysep=2pt, anchor=north]
  at (64,20) {upper bound $=\log(1/\pi_j)+\KL(Q\Vert P_j)$};
\end{scope}
\end{tikzpicture}%
\caption{\textbf{Certifying a recipe after adaptive selection and stopping.} \textbf{(a)}~Selecting an empirically favorable recipe makes a fixed-recipe error bar systematically optimistic. \textbf{(b)}~A single event of probability at least $1-\delta$ covers every calibration size $n$ and randomized recipe $Q\ll P$ simultaneously, and therefore covers any data-dependent stopping time and posterior $(\tau,Q_\tau)$ (\cref{thm:main,cor:rule}). \textbf{(c)}~A hierarchical prior weights recipe families before weighting configurations within each family, bounding the selection penalty by the family code length plus the within-family divergence (\cref{cor:structured}).}\label{fig:hero}
\end{figure}
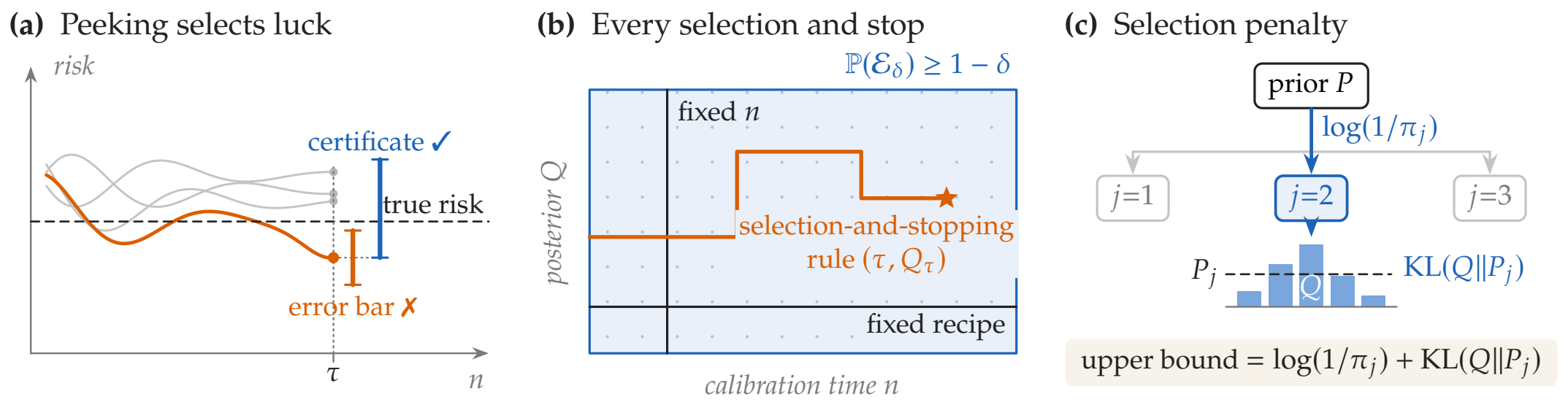

We show that a single commitment made before seeing calibration data suffices: a prior distribution $P$ over the recipe space (\cref{fig:hero}). Given i.i.d.\ labeled calibration examples, \cref{thm:main} establishes an anytime-valid PAC-Bayes-kl certificate that bounds population error rate and expected cost simultaneously across every calibration size $n\ge1$ and every \emph{randomized recipe} $Q\ll P$, a deployment distribution over recipes that includes deterministic selection ($Q=\delta_a$) as a point mass. Because coverage holds uniformly over all pairs $(n,Q)$ on one high-probability event, the practitioner may inspect the calibration stream in any order, expand sweeps across hierarchical recipe families, trade accuracy against token cost on the fly, and stop at any data-dependent time $\tau$. The statistical price of this freedom is exact and transparent: the relative-entropy \emph{selection penalty} $\KL(Q_\tau\Vert P)$. When per-example token cost has no known hard cap, the same construction certifies clipped expectations and tail-exceedance probabilities simultaneously across a dyadic scale of thresholds.

Building on time-uniform PAC-Bayes theory \citep{chugg2023}, this variational formulation extends weighted unions of confidence sequences on countable grids \citep{howard2021,waudbysmith2024} to posteriors over arbitrary recipe spaces. Crucially, the penalty $\KL(Q\Vert P)$ is not an artifact of the proof technique; it reflects an intrinsic coding law (\cref{sec:lower}). Under independent Bernoulli losses, we prove that valid per-recipe certificates obey a Kraft inequality on their binary-kl radii up to $O(\log n)$. Consequently, if an adaptive procedure reports $\hat a\sim\rho$, its expected certificate radius cannot be smaller than the Shannon entropy $H(\rho)-O(\log n)$; choosing the prior $P=\rho$ achieves this benchmark up to logarithmic sample-size terms, while any other prior pays the redundancy $\KL(\rho\Vert P)$. Even if a certificate is permitted to inspect the entire calibration matrix jointly across $M$ exchangeable recipes, a worst-case penalty of $\log M-O(\log n)$ with leading constant one remains unavoidable whenever $\log M\gg\log n$.

Self-consistency illustrates why such certificates must rest on ground-truth labels rather than unlabeled model agreement (\cref{sec:sc}). Plurality voting generates multiple reasoning traces and returns the modal answer \citep{wang2023}. Drawing on large-deviation bounds for multinomial voting \citep{corderoencinar2025}, we characterize plurality as a pure \emph{mode amplifier}: increasing the trace budget concentrates the output exponentially fast onto the generator's strict mode, driving error to zero when that mode is correct and driving accuracy to zero when it is wrong. On MATH-500, wrong answers outvote correct ones on more than a quarter of all problems for both \texttt{gpt-4.1-mini} and \texttt{Qwen2.5-7B-Instruct}. Unlabeled consensus certifies stability, not correctness. Yet the same exponential concentration rate governs how test-time compute should be distributed across inputs: minimizing the resulting risk surrogate under an average-budget constraint yields an exact \emph{water-filling} allocation (\cref{thm:alloc}) that concentrates traces on contestable positive margins rather than hopeless or already decisive problems.

We evaluate our framework across \nGenTotal{} reasoning traces from \texttt{gpt-4.1-mini} and \texttt{Qwen2.5-7B-Instruct} on GSM8K and MATH-500 (\cref{sec:experiments}). The experiments quantify post-selection miscoverage under naive bounds, verify the $1/p$ sample-complexity scaling of binary-kl inversion, map certified risk--cost frontiers, demonstrate how hierarchical priors insulate default pipelines against exploratory sweeps of up to \elocMc{} recipes, and show that paired confidence sequences rank competing recipes over twenty times faster than marginal certificates. On the full GSM8K split, \texttt{gpt-4.1-mini} swept over budgets $\{1,3,5,9\}$ achieves a held-out error rate of \heldGG{} with a 95\% anytime certificate of \certGG{}.

\paragraph{Contributions.}
\begin{enumerate}[leftmargin=1.6em,label=\textup{(\roman*)}]
\item \emph{Anytime-valid post-selection certification (\cref{sec:certificate}):} a simultaneous PAC-Bayes-kl certificate for population risk and compute cost across arbitrary stopping rules, randomized recipes, hierarchical priors, and uncapped heavy-tailed token distributions.
\item \emph{Certificates are codes (\cref{sec:lower}):} a Kraft inequality and Shannon-entropy lower bound for per-recipe certificates under independent Bernoulli losses, paired with a leading-constant-one $\log M$ lower bound for arbitrary data-dependent certificates on exchangeable grids.
\item \emph{Mode amplification, water-filling, and empirical validation (\cref{sec:sc,sec:experiments}):} a large-deviation separation between unlabeled vote consensus and labeled accuracy, an oracle water-filling law with a deployable two-stage allocator, and reproducible benchmarks across \nGenTotal{} LLM reasoning traces.
\end{enumerate}

\section{Setup: recipes, randomized recipes, and calibration data}\label{sec:setup}

We evaluate test-time inference pipelines on labeled examples $Z=(X,Y)\sim\cD$. Throughout Sections~\ref{sec:setup} and~\ref{sec:certificate}, our running example is self-consistency on GSM8K with \texttt{gpt-4.1-mini} at temperature $0.7$ over budgets $b\in\{1,3,5,9\}$, where recipe $a_b$ draws $b$ reasoning traces and outputs their plurality answer.

\begingroup
\tcbset{warmbox/.append style={unbreakable}}
\begin{definition}[Test-time recipe]\label{def:recipe}
A \emph{recipe} $a\in\cA$ is a stochastic inference procedure producing prediction $H_a(X,U)$ at cost $C_a(X,U)\ge0$, where $U$ collects inference-time randomness. A recipe fixes every design choice affecting output or cost: prompt template, decoding law, trace budget, verifier, answer parser, aggregation rule, early-exit criterion, and tie-breaker. Its zero--one loss is $\ell_a(Z,U)=\one\{H_a(X,U)\ne Y\}$. Given a known per-example \emph{cost cap} $\Cmax>0$, write $c_a=C_a/\Cmax\in[0,1]$. The population risk and normalized expected cost are $L(a)=\E\,\ell_a$ and $G(a)=\E\,c_a$.
\end{definition}
\endgroup

Bundling parsers, stopping rules, and tie-breakers inside $a$ ensures that each recipe has a well-defined risk--cost pair $(L(a),G(a))$. When decoding enforces a hard token limit via \texttt{max\_tokens}, $\Cmax$ is exact; \cref{sec:price} removes the need for a known cap. At deployment, a practitioner may commit to a single configuration or randomize across several, for example to break empirical ties symmetrically or smooth a discrete risk--cost frontier.

\begingroup
\tcbset{warmbox/.append style={unbreakable}}
\begin{definition}[Randomized recipe]\label{def:search}
A \emph{randomized recipe} is a distribution $Q\in\Delta(\cA)$ over recipes (a PAC-Bayes posterior). Deployment draws $a\sim Q$ independently of the test input $X$ and executes $a$. Its population risk and normalized cost are the linear expectations $L(Q)=\E_{a\sim Q}L(a)$ and $G(Q)=\E_{a\sim Q}G(a)$; the point mass $\delta_a$ recovers a single deterministic recipe.
\end{definition}
\endgroup

\begin{definition}[Calibration data]\label{def:transcript}
The \emph{calibration data} form an i.i.d.\ sequence of rows $R_i=(Z_i,(U_{i,a})_{a\in\cA})$ for $i\ge1$, with $Z_i\sim\cD$ and each $U_{i,a}$ distributed according to recipe $a$'s inference randomness. Within a single row $i$, the seeds $(U_{i,a})_{a\in\cA}$ may be arbitrarily coupled across recipes (common random numbers). After $n$ calibration examples, the empirical risk and normalized cost are
\[
\Lhat_n(a)=\frac1n\sum_{i\le n}\ell_a(Z_i,U_{i,a}),\qquad
\Ghat_n(a)=\frac1n\sum_{i\le n}c_a(Z_i,U_{i,a}),
\]
extended linearly to randomized recipes via $\Lhat_n(Q)=\E_{a\sim Q}\Lhat_n(a)$ and $\Ghat_n(Q)=\E_{a\sim Q}\Ghat_n(a)$. Let $\cF_n=\sigma(R_1,\ldots,R_n)$ denote the filtration generated by the first $n$ rows.\footnote{All random variables are defined on a common probability space, with maps jointly measurable in the recipe, example, and seed. For $(a,R_i)\mapsto(\ell_a,c_a)$, joint measurability holds automatically whenever $\cA$ is countable or seeds are shared.}
\end{definition}

Allowing arbitrary within-row coupling matches standard engineering practice: in our running GSM8K example, we generate nine traces once for each of the $\ncalGG$ calibration problems and evaluate budget $b$ on the prefix of the first $b$ traces. Sharing traces across budgets leaves every marginal distribution $(Z_i,U_{i,a})$ intact while cutting generation cost to that of the largest budget.

\paragraph{The sole pre-data commitment: the prior.}
Before inspecting any calibration inputs, labels, or model traces, the practitioner declares a reference distribution $P\in\Delta(\cA)$, the \emph{prior}, which may depend on domain knowledge or independent historical data. In our four-budget example, the uniform prior $P(a_b)=1/4$ assigns each budget equal weight and charges selection penalty $\log(1/P(a_b))=\log4$ nats; tilting $P$ toward a trusted baseline lowers its penalty accordingly. Nothing else is fixed in advance. As rows arrive, the analyst may inspect $(\cF_n)$ in any order, revise candidate mixtures, and halt whenever the certificate is sufficiently tight or the labeling budget is spent.

\begin{definition}[Selection-and-stopping rule]\label{def:rule}
An \emph{admissible selection-and-stopping rule} is a pair $(\tau,(Q_n)_{n\ge1})$ consisting of an $(\cF_n)$-stopping time $\tau$ and a sequence of $\cF_n$-measurable randomized recipes $Q_n\ll P$. Upon stopping, the practitioner deploys $Q_\tau$.
\end{definition}

Absolute continuity ($Q_n\ll P$) simply requires that deployment place no mass on events ruled out by the prior; obtaining a non-trivial bound requires $\KL(Q_\tau\Vert P)<\infty$, and reporting a single recipe $a$ requires $P(a)>0$. All probability statements below are taken over the full calibration stream $(R_i)_{i\ge1}$, including examples and recipes never explicitly queried; $n$ counts labeled problems and cost is measured in generated tokens (\cref{app:mc} treats multi-run Monte Carlo averaging per input).

\section{One certificate for every selection and stopping rule}\label{sec:certificate}

A \emph{certificate} is a high-probability upper bound on population risk or cost that remains valid after data-dependent selection and optional stopping. Rather than modeling the analyst's search trajectory, we establish a single concentration event that holds simultaneously across all sample sizes $n\ge1$ and all posteriors $Q\ll P$.

Write $\kl(\hat q\Vert q)$ for the binary relative entropy between Bernoulli distributions with means $\hat q$ and $q$, and define its upper inverse on $\hat q\in[0,1]$, $\eps\ge0$ by
$\klinv(\hat q,\eps)=\sup\{q\in[\hat q,1]:\kl(\hat q\Vert q)\le\eps\}$.
All logarithms are in base $e$, so divergences and penalties are measured in nats.

\begin{theorem}[Anytime-valid PAC-Bayes-kl certificate]\label[theorem]{thm:main}
Let $P$ be a prior on $\cA$ chosen independently of the calibration data, and fix $\delta\in(0,1)$. With probability at least $1-\delta$ over the calibration stream, simultaneously for every $n\ge1$ and every randomized recipe $Q\ll P$,
\begin{equation}\label{eq:main}
\kl\bigl(\Lhat_n(Q)\,\Vert\,L(Q)\bigr)\;\le\;\eps_n(Q)\;:=\;\frac{\KL(Q\Vert P)+\pen{n}}{n}.
\end{equation}
The identical bound holds with $(\Ghat_n,G)$ in place of $(\Lhat_n,L)$.
\end{theorem}

Equation~\eqref{eq:main} cleanly separates two costs in its numerator: $\KL(Q\Vert P)$ prices the adaptive choice of posterior $Q$ relative to the prior $P$, while $\pen{n}$ pays for binomial moment concentration and simultaneous validity across all calibration times $n\ge1$. Setting $U^L_n(Q;\alpha)=\klinv(\Lhat_n(Q),\eps_n(Q;\alpha))$, where $\eps_n(Q;\alpha)$ replaces $\delta$ with $\alpha$ in~\eqref{eq:main}, and defining $U^G_n(Q;\alpha)$ analogously from $\Ghat_n(Q)$, post-selection validity follows immediately without any martingale assumptions on how the practitioner explores.

\begin{corollary}[Adaptive selection and stopping]\label[corollary]{cor:rule}
On the $(1-\delta)$-probability event of \cref{thm:main}, every admissible selection-and-stopping rule satisfies $L(Q_\tau)\le U^L_\tau(Q_\tau;\delta)$ whenever $\tau<\infty$; on the corresponding event for cost, $G(Q_\tau)\le U^G_\tau(Q_\tau;\delta)$.
\end{corollary}

Because~\eqref{eq:main} holds for every deterministic pair $(n,Q)$, it necessarily holds at the realized pair $(\tau,Q_\tau)$. In our four-budget running example, an engineer can track all four empirical error curves online, switch favorites freely as evidence accumulates, and stop at will, paying the constant selection penalty $\log4$ throughout. Complete proofs appear in \cref{app:tools,app:event}.

\subsection{Prior geometry, joint risk--cost bounds, and uncapped costs}\label{sec:price}

\paragraph{Point selection, randomization, and hierarchical priors.}
When reporting a single recipe $a$ with $P(a)>0$, the selection penalty reduces to $\KL(\delta_a\Vert P)=\log(1/P(a))$, the Shannon code length of $a$ under $P$ \citep{blum2003,grunwald2007}. A uniform prior over $M$ candidates charges $\log M$ uniformly; tilting $P$ toward simpler or preferred configurations rewards them with shorter code lengths. As we prove in \cref{sec:lower}, code-length penalties are unavoidable up to $O(\log n)$, and the optimal prior forecasts the distribution of the reported recipe (\cref{app:prior} translates this principle into concrete prior-design rules).

Randomizing across near-tied winners actively shrinks the penalty through posterior entropy. On a countable space, $\KL(Q\Vert P)=\E_{a\sim Q}[\log(1/P(a))]-H(Q)$. Under a uniform prior over $M$ recipes, mixing uniformly across $k$ tied winners reduces the penalty from $\log M$ to $\log(M/k)$: on the $\gridMQGrid$-recipe Qwen sweep of \cref{sec:grids}, randomizing over the $\mixKQGrid$ co-optimal recipes tightens the 95\% certificate from $\gridCertQGrid$ to $\mixCertQGrid$.

Practical sweeps are rarely flat lists; they are organized into structured families, such as a trusted baseline prompt-and-temperature pair alongside several experimental templates, verifiers, or routing rules. A \emph{hierarchical prior} assigns probability $\pi_j$ to each family $j\in\cJ$ and a conditional prior $P_j$ within that family, insulating the baseline from exploratory expansion.

\begin{corollary}[Hierarchical priors]\label[corollary]{cor:structured}
Let $\{P_j:j\in\cJ\}$ be a countable collection of priors and $\pi$ a probability mass function on $\cJ$, all declared independently of the calibration data. With probability at least $1-\delta$, simultaneously for every $n\ge1$, every $j\in\cJ$, and every $Q\ll P_j$,
\begin{equation}\label{eq:local}
\kl\bigl(\Lhat_n(Q)\Vert L(Q)\bigr)\;\le\;\frac{\KL(Q\Vert P_j)+\log(1/\pi_j)+\pen{n}}{n}.
\end{equation}
\end{corollary}

Applying \cref{thm:main} to the mixture $P_\pi=\sum_j\pi_jP_j$ and noting $dP_j/dP_\pi\le1/\pi_j$ yields~\eqref{eq:local} for all families simultaneously without an extra union bound. The practitioner chooses the family $j$ after seeing the data, paying the family code length $\log(1/\pi_j)$ plus the within-family divergence $\KL(Q\Vert P_j)$. Within family $j$, a natural data-dependent posterior is the Gibbs law $dQ/dP_j\propto e^{-\eta n\Lhat_n}$ ($\eta>0$), which concentrates smoothly on low-error recipes at inverse temperature $\eta$ with a closed-form divergence (\cref{tab:ledger}).

\paragraph{Joint certification of accuracy and compute.}
Test-time scaling is fundamentally multi-objective: higher accuracy costs tokens and latency. Splitting the failure budget $\delta=\alpha_L+\alpha_G$ between loss and normalized cost certifies the entire risk--cost Pareto frontier at once.

\begin{corollary}[Joint risk--cost certificate]\label[corollary]{cor:joint}
Fix $\alpha_L,\alpha_G>0$ with $\alpha_L+\alpha_G\le\delta$. With probability at least $1-\delta$, every admissible selection-and-stopping rule satisfies, on $\{\tau<\infty\}$,
\[
L(Q_\tau)\;\le\;U^L_\tau(Q_\tau;\alpha_L),
\qquad
G(Q_\tau)\;\le\;U^G_\tau(Q_\tau;\alpha_G).
\]
\end{corollary}

Crucially, because both bounds hold uniformly over $Q\ll P$, the deployer does not need to pre-commit to an exchange rate between errors and tokens. Minimizing the certified Lagrangian
\[
Q_n\in\arg\min_{Q\ll P}\bigl\{U^L_n(Q;\alpha_L)+\lambda\,U^G_n(Q;\alpha_G)\bigr\}
\]
is admissible for any $\cF_n$-measurable price of compute $\lambda\ge0$, allowing a single calibration run to serve deployments with distinct latency or budget constraints.

\paragraph{Uncapped and heavy-tailed costs.}
When reasoning models run without a strict token cap, generation lengths exhibit heavy tails, and no non-trivial distribution-free upper bound on the unclipped mean $\E\,C$ can exist \citep{bahadur1956}. The right certifiable summary is instead the pair of \emph{clipped expectation} and \emph{tail exceedance}. Fixing a reference scale $C_0>0$ independently of the data and a dyadic grid $C_k=C_02^k$ ($k\ge0$) weighted by $\pi_k=1/[(k+1)(k+2)]$, Corollary~\ref{cor:envelope} (\cref{app:event}) simultaneously bounds $\E\min\{C_a/C_k,1\}$ and $\P\{C_a>C_k\}$ across all $n$, $k$, and $Q\ll P$ at an additional penalty of only $\log[2(k+1)(k+2)]=O(\log\log(C_k/C_0))$ nats, permitting data-dependent threshold selection after inspecting the empirical tail.

\begin{table}[!htbp]
\centering\small
\caption{\textbf{Selection penalties for standard prior architectures.} Point-selection rows report $\KL(\delta_a\Vert P)$; the hierarchical row gives the upper bound of \cref{cor:structured}; the uncapped-cost row includes joint selection of the dyadic threshold $C_k$ and functional (clipped mean vs.\ exceedance).}\label{tab:ledger}
\setlength{\tabcolsep}{5pt}
\begin{tabularx}{\linewidth}{@{}>{\raggedright\arraybackslash}Xl@{}}
\headtop
\headstrut \textbf{recipe choice and prior} & \textbf{selection penalty (nats)}\\
\headbottom
single recipe fixed in advance; point-mass prior & $0$\\[3pt]
single recipe from a uniform grid of size $M$ & $\log M$\\[3pt]
unbounded budget ladder $b\ge1$; $P(b)\propto(1{+}b)^{-2}$ & $2\log(1{+}b)+\log(\pi^2/6-1)$\\[3pt]
posterior $Q$ in family $j$; family weights $\pi_j$, prior $P_j$ & $\le\log(1/\pi_j)+\KL(Q\Vert P_j)$\\[3pt]
Gibbs posterior $dQ/dP_j\propto e^{-\eta n\Lhat_n}$ within family $j$ & $-\eta n\Lhat_n(Q)-\log\E_{P_j}e^{-\eta n\Lhat_n}$\\[3pt]
uncapped cost at threshold $C_02^k$; $\pi_k=1/[(k{+}1)(k{+}2)]$ & $\KL(Q\Vert P)+\log[2(k{+}1)(k{+}2)]$\\
\bottomrule
\end{tabularx}
\end{table}

\begin{takeaway}
Commit to a prior $P$ once before viewing calibration data. Every subsequent decision---which families to explore, how to trade cost against accuracy, when to stop, and which posterior $Q_\tau$ to deploy---is protected simultaneously at information price $\KL(Q_\tau\Vert P)$.
\end{takeaway}

\subsection{Sample complexity of binary-kl certification}\label{sec:radius}

Working in the binary-kl geometry rather than additive Hoeffding/Pinsker form is decisive in the low-error regime typical of strong reasoning models. When a recipe makes zero empirical errors ($\Lhat_n(Q)=0$), exact kl inversion yields $1-e^{-\eps_n}\approx\eps_n$, scaling as $O(1/n)$, whereas Pinsker's relaxation $\Lhat_n+\sqrt{\eps_n/2}$ degrades to the slow $O(1/\sqrt{n})$ rate.

To see how many calibration examples are needed to resolve a small true risk $p\in(0,1/2)$ within a factor of two, evaluate~\eqref{eq:main} for one recipe chosen from a uniform grid of size $M$ when $\Lhat_n=p$. Expanding $\kl(p\Vert2p)=(1-\log2)\,p+O(p^2)$ and solving $\kl(p\Vert2p)=\eps_n$ gives the characteristic calibration horizon
\begin{equation}\label{eq:nstar}
n^\star\;\approx\;\frac{\log M+\tfrac52\log n^\star+\log(2/\delta)}{(1-\log 2)\,p}\;\approx\;\klTwoInv\cdot\frac{\log M+\tfrac52\log n^\star+\log(2/\delta)}{p}.
\end{equation}
Two structural lessons emerge immediately (\cref{tab:nstar}). First, relative precision scales inversely with the error rate ($n^\star\propto 1/p$ up to logarithmic factors): certifying a $2\%$ error rate within $2\times$ requires $n^\star=\nstarFourb$ examples ($M=4$), explaining why a 96-example pilot at empirical risk $\calPilot$ yields a conservative certificate of $\certPilot$, whereas along the full GSM8K stream ($p\approx0.04$) the bound tightens from $\trajGGatNinetySix$ at $n=96$ to $\certGG$ at $n=\ncalGG$ (\cref{sec:scale}). Second, sweep breadth enters only logarithmically: sextupling the grid from $M=4$ to $M=24$ increases $n^\star$ by just \nstarGrowMin--\nstarGrowMax\%, and expanding $M$ from $5$ to $27$ at $n=250$ adds a mere $(\log27-\log5)/250\approx\gridWidthNats$ nats to $\eps_n$. In moderate sweeps, generating traces costs far more than the statistical penalty of comparing extra candidates.

\begin{table}[!htbp]
\centering\small
\caption{\textbf{Calibration sample size $n^\star$ required for a certificate within $2\times$ of true risk.} Exact numerical inversion of~\eqref{eq:main} for a single recipe under a uniform prior at $\delta=0.05$ with $\Lhat_n=p$.}\label{tab:nstar}
\begin{tabular}{@{}lccccc@{}}
\headtop
\headstrut \textbf{true risk $p$} & \textbf{0.01} & \textbf{0.02} & \textbf{0.05} & \textbf{0.10} & \textbf{0.20}\\
\headbottom
$n^\star$, uniform grid of $M=4$ & \nstarFoura & \nstarFourb & \nstarFourc & \nstarFourd & \nstarFoure\\
$n^\star$, uniform grid of $M=24$ & \nstarTwentyFoura & \nstarTwentyFourb & \nstarTwentyFourc & \nstarTwentyFourd & \nstarTwentyFoure\\
\bottomrule
\end{tabular}
\end{table}
\FloatBarrier

\subsection{Information-theoretic optimality: the selection penalty cannot be avoided}\label{sec:lower}

Could a cleverer post-selection procedure certify adaptive sweeps without paying a code-length penalty? We establish two complementary lower bounds showing that $\KL(Q\Vert P)$ is fundamental (\cref{app:price}).

\paragraph{Certificates are codes (Theorem~\ref{thm:codes}).}
Fix $n\ge1$ and $\delta\in(0,1/2]$. Consider any family of \emph{per-recipe certificates} $U_a=\phi_a(\ell_{1:n}(a))$ (bounds depending only on recipe $a$'s own loss sequence, encompassing both point-mass PAC-Bayes bounds and weighted unions of confidence sequences) that is simultaneously valid ($\P\{\forall a,\,L(a)\le U_a\}\ge1-\delta$) across all product Bernoulli loss laws. Let $r_a$ denote the least permutation-median binary-kl radius of $U_a$.\footnote{Write $\bar\phi_a(x)$ for a median of $\phi_a$ over uniform permutations of binary sequences of empirical mean $x\in\{0,1/n,\ldots,1\}$, and set $r_a=\min\{n\kl(x\Vert\bar\phi_a(x)):x\in\{0,1/n,\ldots,1\},\ \bar\phi_a(x)<1\}$, taking $\kl=0$ when $\bar\phi_a(x)\le x$ and $\min\emptyset=\infty$.} Then the radii $(r_a)_{a\in\cA}$ must obey a Kraft-type budget, and any reporting rule outputting $\hat a\sim\rho$ faces a Shannon-entropy floor:
\[
\sum_{a\in\cA} e^{-r_a}\;\le\;4\sqrt{2n}\,\delta,
\qquad
\E\, r_{\hat a}\;\ge\;H(\rho)-\log\bigl(4\sqrt{2n}\,\delta\bigr).
\]
Conversely, for any fixed reporting rule with output law $\rho$, \cref{thm:main} instantiated with prior $q$ achieves expected radius $H(\rho)+\KL(\rho\Vert q)+\pen{n}$. Matching the prior to the reporting law ($q=\rho$) is therefore optimal up to $O(\log n)$ at any fixed failure level $\delta$, and any mismatched prior pays the exact Kullback--Leibler redundancy $\KL(\rho\Vert q)$.

\paragraph{Unrestricted certificates on exchangeable grids (Proposition~\ref{prop:lower}).}
What if the certificate $U$ is an arbitrary measurable function of the \emph{entire} $M\times n$ calibration matrix, free to pool information across recipes? Let $n\ge20$ and $2\le M\le e^{n/8}$, suppose $M$ recipes incur i.i.d.\ Bernoulli$(1/2)$ losses, and let $\hat a\in\arg\min_a\Lhat_n(a)$ be an empirical winner. Then any valid bound satisfying $\P\{L(\hat a)\le U\}\ge1-\delta$ must obey, for every $\eta\in(0,1/2]$ with probability at least $1-\delta-\eta$,
\[
n\,\kl\bigl(\Lhat_n(\hat a)\,\Vert\,\min\{U,1\}\bigr)
\;\ge\;\log M-\tfrac12\log n-\log\bigl(4\sqrt2\log(1/\eta)\bigr).
\]
When $\log M\gg\log n$ at fixed $\delta,\eta$, the $\log M$ selection penalty is sharp with leading constant one.

\section{Why self-consistency needs labels}\label{sec:sc}

Because self-consistency operates without ground-truth labels at test time, sampling $b$ reasoning chains and returning their plurality vote \citep{wang2023}, it is tempting to equate high inter-sample agreement with answer correctness \citep{corderoencinar2025,ota2026cite}. Large-deviation analysis reveals why unlabeled consensus cannot substitute for labeled calibration: voting exponentially suppresses sampling variance, but is blind to systematic bias.

For a fixed input $x$ with ground-truth answer $y=y(x)$, let $p_x$ denote the distribution of a single trace's extracted terminal answer. Define the \emph{vote margin}
\[
\Delta_x\;=\;p_x(y)-\max_{a\ne y}p_x(a),
\]
which is strictly positive if and only if the correct answer $y$ is the unique mode of $p_x$. Write $\SC_b(x)$ for plurality vote over $b$ i.i.d.\ traces from $p_x$ under an arbitrary tie-breaking rule, and let $B(u,v)=-\log\bigl(1-(\sqrt u-\sqrt v)^2\bigr)$ denote the Bhattacharyya exponent.

\begingroup
\setlength{\abovedisplayskip}{6pt}
\setlength{\belowdisplayskip}{6pt}
\setlength{\abovedisplayshortskip}{4pt}
\setlength{\belowdisplayshortskip}{4pt}
\tcbset{coolbox/.append style={breakable=false}}
\begin{proposition}[Plurality voting is a mode amplifier]\label[proposition]{prop:margin}
Suppose $p_x$ is supported on at most $m$ distinct answers. If the correct answer $y$ is the strict mode of $p_x$ with runner-up $a_2$, then
\[
\P\{\SC_b(x)\ne y\}\;\le\;(m-1)\,e^{-bs_x},
\qquad s_x=B\bigl(p_x(y),p_x(a_2)\bigr).
\]
Conversely, if an incorrect answer $a^\star\ne y$ is the strict mode of $p_x$, then
\[
\P\{\SC_b(x)=y\}\;\le\;e^{-bB(p_x(a^\star),p_x(y))}.
\]
Averaging over $X\sim\cD$ yields the population risk decomposition
\[
L(\SC_b)\;\le\;\P\{\Delta_X\le0\}+\E\bigl[\min\{1,(m_X-1)e^{-bs_X}\}\,\one\{\Delta_X>0\}\bigr].
\]
\end{proposition}
\endgroup

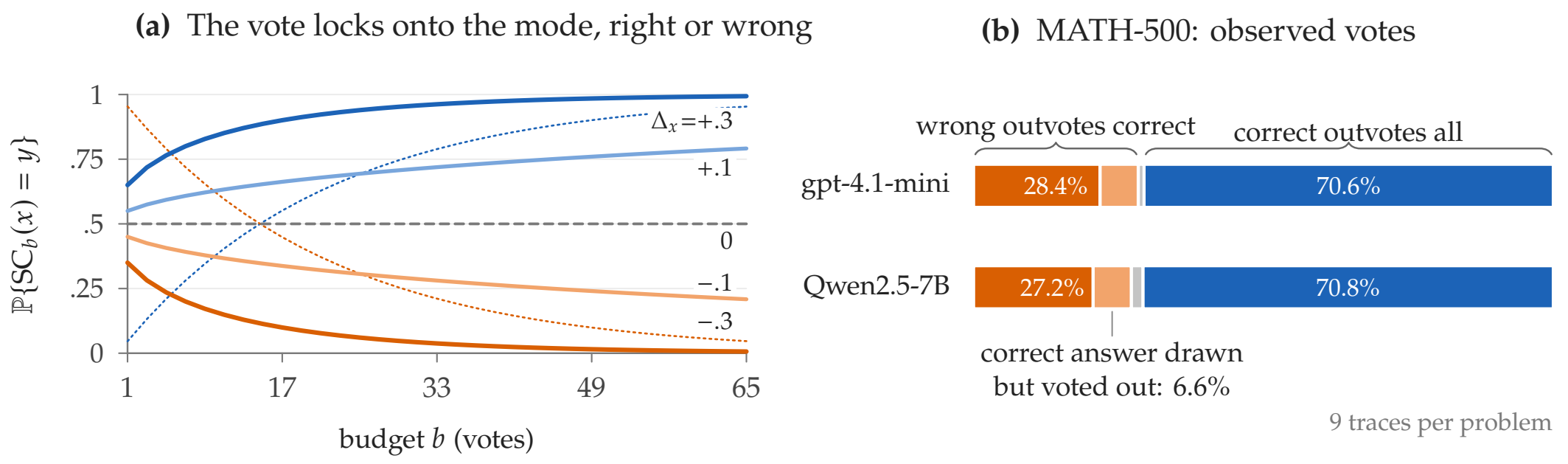
\begin{figure}[!htbp]
\centering
\pgfplotstableread[row sep=crcr]{
b up03 up01 lo01 lo03\\
1 0.046061 0.005013 0.994987 0.953939\\
3 0.131915 0.014962 0.985038 0.868085\\
5 0.210043 0.024813 0.975187 0.789957\\
7 0.281139 0.034565 0.965435 0.718861\\
9 0.345837 0.044219 0.955781 0.654163\\
11 0.404711 0.053777 0.946223 0.595289\\
13 0.458287 0.063239 0.936761 0.541713\\
15 0.507041 0.072607 0.927393 0.492959\\
17 0.551408 0.081881 0.918119 0.448592\\
19 0.591781 0.091062 0.908938 0.408219\\
21 0.628521 0.100151 0.899849 0.371479\\
23 0.661954 0.109150 0.890850 0.338046\\
25 0.692378 0.118058 0.881942 0.307622\\
27 0.720064 0.126878 0.873122 0.279936\\
29 0.745258 0.135609 0.864391 0.254742\\
31 0.768185 0.144253 0.855747 0.231815\\
33 0.789048 0.152810 0.847190 0.210952\\
35 0.808034 0.161282 0.838718 0.191966\\
37 0.825311 0.169669 0.830331 0.174689\\
39 0.841033 0.177973 0.822027 0.158967\\
41 0.855340 0.186193 0.813807 0.144660\\
43 0.868359 0.194331 0.805669 0.131641\\
45 0.880207 0.202388 0.797612 0.119793\\
47 0.890988 0.210364 0.789636 0.109012\\
49 0.900799 0.218260 0.781740 0.099201\\
51 0.909727 0.226078 0.773922 0.090273\\
53 0.917852 0.233817 0.766183 0.082148\\
55 0.925245 0.241479 0.758521 0.074755\\
57 0.931973 0.249064 0.750936 0.068027\\
59 0.938096 0.256573 0.743427 0.061904\\
61 0.943667 0.264007 0.735993 0.056333\\
63 0.948737 0.271367 0.728633 0.051263\\
65 0.953351 0.278654 0.721346 0.046649\\
}{\tblMarginBounds}%
\pgfplotstableread[row sep=crcr]{
b d0 d1 d2 d3 d4\\
1 0.650000 0.550000 0.500000 0.450000 0.350000\\
3 0.718250 0.574750 0.500000 0.425250 0.281750\\
5 0.764831 0.593127 0.500000 0.406873 0.235169\\
7 0.800154 0.608288 0.500000 0.391712 0.199846\\
9 0.828281 0.621421 0.500000 0.378579 0.171719\\
11 0.851316 0.633123 0.500000 0.366877 0.148684\\
13 0.870532 0.643742 0.500000 0.356258 0.129468\\
15 0.886769 0.653504 0.500000 0.346496 0.113231\\
17 0.900621 0.662564 0.500000 0.337436 0.099379\\
19 0.912526 0.671036 0.500000 0.328964 0.087474\\
21 0.922818 0.679003 0.500000 0.320997 0.077182\\
23 0.931759 0.686533 0.500000 0.313467 0.068241\\
25 0.939555 0.693676 0.500000 0.306324 0.060445\\
27 0.946377 0.700476 0.500000 0.299524 0.053623\\
29 0.952363 0.706968 0.500000 0.293032 0.047637\\
31 0.957629 0.713180 0.500000 0.286820 0.042371\\
33 0.962271 0.719138 0.500000 0.280862 0.037729\\
35 0.966372 0.724863 0.500000 0.275137 0.033628\\
37 0.969999 0.730373 0.500000 0.269627 0.030001\\
39 0.973213 0.735685 0.500000 0.264315 0.026787\\
41 0.976065 0.740812 0.500000 0.259188 0.023935\\
43 0.978598 0.745767 0.500000 0.254233 0.021402\\
45 0.980851 0.750561 0.500000 0.249439 0.019149\\
47 0.982857 0.755204 0.500000 0.244796 0.017143\\
49 0.984644 0.759704 0.500000 0.240296 0.015356\\
51 0.986238 0.764071 0.500000 0.235929 0.013762\\
53 0.987660 0.768311 0.500000 0.231689 0.012340\\
55 0.988930 0.772430 0.500000 0.227570 0.011070\\
57 0.990066 0.776436 0.500000 0.223564 0.009934\\
59 0.991081 0.780333 0.500000 0.219667 0.008919\\
61 0.991990 0.784127 0.500000 0.215873 0.008010\\
63 0.992803 0.787822 0.500000 0.212178 0.007197\\
65 0.993532 0.791423 0.500000 0.208577 0.006468\\
}{\tblMarginCurves}%
\begin{tikzpicture}[figbase]
\begin{axis}[ygridplot, name=left, scale only axis, width=158pt, height=66pt,
  xmin=1, xmax=65, ymin=0, ymax=1, xtick={1,17,33,49,65}, ytick={0,0.25,0.5,0.75,1},
  yticklabels={0,.25,.5,.75,1},
  xlabel={budget $b$ (votes)}, ylabel={$\P\{\SC_b(x)=y\}$},
  title={\ptitle{a}{The vote locks onto the mode, right or wrong}}]
  \addplot[CoolInk, line width=0.5pt, densely dotted] table[x=b, y=up03] {\tblMarginBounds};
  \addplot[WarmInk, line width=0.5pt, densely dotted] table[x=b, y=lo03] {\tblMarginBounds};
  \addplot[CoolInk, line width=1.15pt] table[x=b, y=d0] {\tblMarginCurves};
  \addplot[CoolMid, line width=1.0pt] table[x=b, y=d1] {\tblMarginCurves};
  \addplot[MidGray, line width=0.7pt, dash pattern=on 2.2pt off 1.6pt] table[x=b, y=d2] {\tblMarginCurves};
  \addplot[WarmMid, line width=1.0pt] table[x=b, y=d3] {\tblMarginCurves};
  \addplot[WarmInk, line width=1.15pt] table[x=b, y=d4] {\tblMarginCurves};
  \node[font=\tiny, text=Ink, anchor=east, fill=white, inner sep=0.8pt] at (axis cs:64,0.895) {$\Delta_x{=}{+}.3$};
  \node[font=\tiny, text=Ink, anchor=east, fill=white, inner sep=0.8pt] at (axis cs:64,0.72) {${+}.1$};
  \node[font=\tiny, text=Ink, anchor=east, fill=white, inner sep=0.8pt] at (axis cs:64,0.435) {$0$};
  \node[font=\tiny, text=Ink, anchor=east, fill=white, inner sep=0.8pt] at (axis cs:64,0.275) {${-}.1$};
  \node[font=\tiny, text=Ink, anchor=east, fill=white, inner sep=0.8pt] at (axis cs:64,0.125) {${-}.3$};
\end{axis}
\pgfmathsetmacro{\gNever}{\marNegGM-\marVotedOutGM}
\pgfmathsetmacro{\qNever}{\marNegQM-\marVotedOutQM}
\begin{axis}[paperplot, name=right, at={(left.south east)}, anchor=south west, xshift=58pt,
  scale only axis, width=148pt, height=66pt,
  xbar stacked, bar width=11pt, xmin=0, xmax=100, ymin=0.35, ymax=2.9,
  hide x axis, ytick={1,2}, yticklabels={Qwen2.5-7B,gpt-4.1-mini},
  y axis line style={draw=none}, ytick style={draw=none},
  yticklabel style={font=\scriptsize, text=Ink},
  every axis plot/.append style={draw=white, line width=0.8pt},
  title={\ptitle{b}{MATH-500: observed votes}}]
  \addplot[fill=WarmInk]   coordinates {(\qNever,1) (\gNever,2)};
  \addplot[fill=WarmMid]   coordinates {(\marVotedOutQM,1) (\marVotedOutGM,2)};
  \addplot[fill=LightGray] coordinates {(\marTieQM,1) (\marTieGM,2)};
  \addplot[fill=CoolInk]   coordinates {(\marPosQM,1) (\marPosGM,2)};
  \coordinate (gneg) at (axis cs:{\marNegGM/2},2);
  \coordinate (qneg) at (axis cs:{\marNegQM/2},1);
  \coordinate (gpos) at (axis cs:{100-\marPosGM/2},2);
  \coordinate (qpos) at (axis cs:{100-\marPosQM/2},1);
  \draw[decorate, decoration={brace, amplitude=3pt}, line width=0.45pt, draw=Ink!70]
    (axis cs:0.3,2.31) -- (axis cs:{\marNegGM-0.3},2.31)
    node[midway, above=3pt, font=\scriptsize, text=Ink, inner sep=0pt] {wrong outvotes correct};
  \draw[decorate, decoration={brace, amplitude=3pt}, line width=0.45pt, draw=Ink!70]
    (axis cs:{\marNegGM+\marTieGM+0.3},2.31) -- (axis cs:99.7,2.31)
    node[midway, above=3pt, font=\scriptsize, text=Ink, inner sep=0pt] {correct outvotes all};
  \draw[Ink!55, line width=0.35pt] (axis cs:{\qNever+\marVotedOutQM/2},0.76) -- (axis cs:{\qNever+\marVotedOutQM/2},0.52)
    node[below, font=\scriptsize, text=Ink, inner sep=1.2pt, align=center] {correct answer drawn\\but voted out: \marVotedOutQM\%};
  \node[font=\tiny, text=MidGray, anchor=north east, inner sep=0pt] at (axis cs:100,-0.22) {9 traces per problem};
\end{axis}
\foreach \c/\v in {gneg/\marNegGM, qneg/\marNegQM, gpos/\marPosGM, qpos/\marPosQM}
  \node[font=\tiny, text=white, inner sep=0pt] at (\c) {\v\%};
\end{tikzpicture}%
\caption{\textbf{Plurality voting locks onto whichever answer leads.} \textbf{(a)}~Exact plurality accuracy across vote budgets $b$ for binary answer laws at margins $\Delta_x\in\{0,\pm0.1,\pm0.3\}$, alongside the exponential envelopes of \cref{prop:margin} at $\Delta_x=\pm0.3$ (dotted). \textbf{(b)}~Empirical vote outcomes over nine traces per problem on MATH-500: correct answer never sampled (dark orange), sampled at least once but outvoted (light orange), tied for the lead (grey: \marTieGM\% for \texttt{gpt-4.1-mini}, \marTieQM\% for \texttt{Qwen2.5-7B-Instruct}), and strictly leading (blue).}\label{fig:margin}
\end{figure}

Proposition~\ref{prop:margin} exposes a sharp dichotomy (\cref{fig:margin}a).\footnote{The forward bound is the Chernoff form of Theorem~2.2 in \citet{corderoencinar2025}, whose Theorem~2.3 characterizes the exact large-deviation rate; the converse applies their pairwise contest bound (App.~A.4) when a wrong answer is modal. Whereas their work certifies unlabeled mode recovery per instance, we connect mode concentration to labeled population risk and cross-input budget allocation. Note also $B(u,v)\ge(u-v)^2/[2(u+v)]$: when probability mass disperses across many wrong answers, $s_x$ exceeds $\Delta_x^2/2$, capturing the signal-to-noise boost of dispersed errors.} On the positive-margin subpopulation $\{\Delta_X>0\}$, extra compute drives error down exponentially at rate $s_X$. On the negative-margin subpopulation $\{\Delta_X<0\}$, extra compute drives the probability of recovering the truth to zero just as exponentially, locking onto a confident hallucination and explaining why scaling curves can plateau well above zero or even regress with larger budgets \citep{chen2024calls}.

Empirical vote distributions on MATH-500 make this phenomenon concrete (\cref{fig:margin}b). Across nine traces per problem, an incorrect answer outvotes the truth on \marNegGM\% of problems for \texttt{gpt-4.1-mini} and \marNegQM\% for \texttt{Qwen2.5-7B-Instruct}. Tellingly, on \marVotedOutGM\% and \marVotedOutQM\% of all problems, respectively, the model \emph{generates the correct answer at least once} yet votes it down in favor of a more frequent mistake. Unlabeled agreement cannot detect this failure floor $\P\{\Delta_X\le0\}$; labeled calibration measures it directly.

At the same time, the population bound in \cref{prop:margin} reveals how an ideal router should distribute a finite average compute budget $B$ across inputs. Spending extra traces on $\Delta_x\le0$ is futile, while spending many traces when $\Delta_x$ is near $1$ is wasteful. Minimizing the continuous-budget exponential surrogate $(m_X-1)e^{-s_X\beta(X)}\one\{\Delta_X>0\}$ under $\E\beta(X)\le B$ yields an exact \emph{water-filling} solution (\cref{thm:alloc}, \cref{app:voting}): every input funded above the minimum single trace is poured to a common marginal risk reduction $\lambda$, peaking on contestable positive margins (\cref{fig:waterfill}). Estimating $(\Delta_x,s_x,m_x)$ online from a small pilot batch of traces turns this oracle benchmark into a self-contained two-stage allocation recipe (\cref{rem:plugin}) whose realized accuracy and variable token cost fall directly under the joint certificate of \cref{cor:joint}.

\section{Experiments}\label{sec:experiments}

\paragraph{Experimental setup.}
We evaluate two models spanning proprietary API and open-weight local serving: \texttt{gpt-4.1-mini} and \texttt{Qwen2.5-7B-Instruct} \citep{qwen2025} served via vLLM \citep{kwon2023}, using default decoding parameters (temperature $0.7$, top-$p$ $0.95$) unless swept, and hard per-trace caps of 512 tokens on GSM8K \citep{cobbe2021} (all 1{,}319 test problems) and 1{,}024 tokens on MATH-500 \citep{hendrycks2021,lightman2023}. Candidate recipes span nested self-consistency budgets (voting on the first $b$ of $b_{\max}$ shared traces), vote-lead early stopping $\mathrm{ES}(\text{lead},\text{cap})$, prompt-length routing, LLM-as-verifier best-of-3, two-stage water-filling, and multi-template temperature grids (\cref{app:recipes}). For prospective certification, the first half of each dataset in published order forms the calibration stream ($n_{\mathrm{cal}}=\ncalGG$ on GSM8K, $\ncalGM$ on MATH-500, and $250/250$ on the 500-problem multi-template grids) and the second half is held out solely to audit ground-truth empirical risk; random resplits audit post-selection coverage (\cref{sec:miscoverage}), full streams evaluate paired ranking (\cref{sec:ranking}), and unlabeled held-out traces calibrate the water-filling threshold $\lambda$. All certificates are reported at failure level $\delta=0.05$ ($\alpha_L=\delta$ for risk-only bounds; $\alpha_L=\alpha_G=\delta/2$ for joint risk--cost bounds). A single deterministic script regenerates every table and figure from the recorded trace pools (\cref{app:repro}).

\afterpage{\begin{figure}[t]
\centering
\input{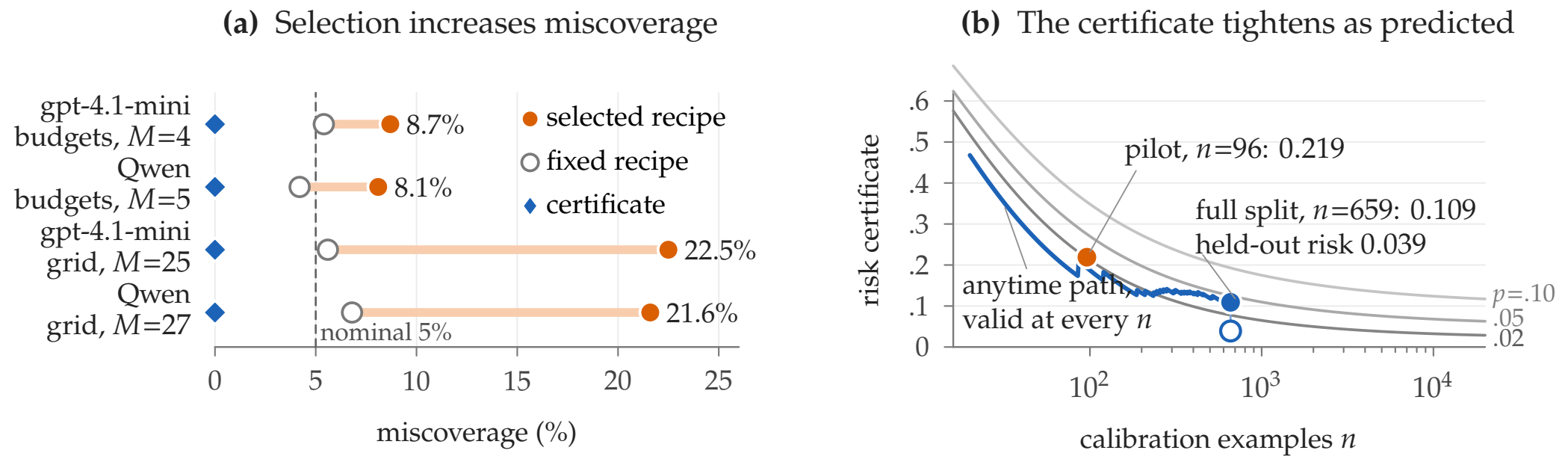}
\caption{\textbf{Naive intervals miscover after selection; anytime certificates track theoretical scaling.} \textbf{(a)}~Empirical miscoverage across 3{,}000 random resplits ($n=150$) per sweep, measuring how often held-out empirical risk exceeds the reported 95\% upper bound. Fixed-recipe Wilson intervals stay near the 5\% nominal level before selection (hollow circles) but fail up to \misWilsonGGrid\% of the time on the selected winner (orange circles); our PAC-Bayes-kl certificate achieves 0\% miscoverage across all 12{,}000 trials (blue diamonds). \textbf{(b)}~Theoretical kl-certificate trajectories at constant empirical risk $p\in\{0.02,0.05,0.10\}$ ($M=4$, grey) versus the live anytime certificate path on GSM8K (\texttt{gpt-4.1-mini}), tightening from pilot scale to \certGG{} at $n=\ncalGG$ above a held-out risk of \heldGG{}.}\label{fig:audits}
\end{figure}}

\subsection{Post-selection miscoverage of fixed-recipe intervals}\label{sec:miscoverage}

To measure the damage caused by post-selection optimism on real model outputs, we draw 3{,}000 random calibration splits of size $n=150$ from each recorded trace pool, select the recipe minimizing calibration error, and check whether its nominal 95\% upper bound covers empirical risk on the held-out complement (\cref{fig:audits}a).

Evaluated on a single recipe fixed \emph{before} seeing the split, a one-sided 95\% Wilson upper bound behaves as advertised, miscovering at average rates of \ctrlMeanMin--\ctrlMeanMax\% across the four pools (at most \ctrlMaxQGrid\% on any individual recipe). Applied to the empirically selected winner, however, Wilson miscoverage climbs to \misWilsonGB\% and \misWilsonQB\% on the small budget sweeps ($M=\MmisGB,\MmisQB$) and surges to \misWilsonGGrid\% and \misWilsonQGrid\%---more than four times the nominal 5\% budget---on the multi-template recipe grids ($M=\MmisGGrid,\MmisQGrid$), driven by an average winner's curse of \optimismGB--\optimismQGrid{} in empirical risk. By contrast, \cref{thm:main} covers the held-out empirical risk in all 12{,}000 trials. Notably, a naive fixed-recipe Hoeffding bound (margin $\hoeffWidth$ at $n=150$) happens not to miscover at $M\le27$ only because its worst-case variance slack accidentally masks small-grid selection bias; in our controlled synthetic stress test at true risk $1/2$ (\cref{app:synthetic}), selection optimism grows as $\sqrt{\log M/n}$ and collapses fixed-recipe Hoeffding coverage to $\synthFixedCovMax$ by $M=\synthMmax$, while the PAC-Bayes certificate maintains 100\% coverage throughout.

\subsection{Certificate tightness, sample scaling, and finite-grid baselines}\label{sec:scale}

\afterpage{\begin{table}[t]
\centering\small
\caption{\textbf{Prospective full-scale certifications.} Risk certificates ($\delta=0.05$, uniform prior over $M$ candidates) for the recipe minimizing calibration risk with a token-cost tie-break. Row~1 reports a 96-example GSM8K pilot using an alternative prompt, parser, and top-$p$ (\cref{app:recipes}).}\label{tab:scale}
\setlength{\tabcolsep}{4.5pt}
\begin{tabular}{@{}llrrcccc@{}}
\headtop
\headstrut \textbf{dataset} & \textbf{model} & \textbf{$M$} & \textbf{$n_{\mathrm{cal}}$} & \textbf{selected} & \textbf{cal.\ risk} & \textbf{held-out risk} & \textbf{certificate}\\
\headbottom
GSM8K (pilot) & gpt-4.1-mini & 4 & 96 & $b=3$ & \calPilot & \heldPilot & \certPilot\\
GSM8K & gpt-4.1-mini & \MGG & \ncalGG & $b=\selGG$ & \calGG & \heldGG & \textbf{\certGG}\\
GSM8K & Qwen2.5-7B & \MQG & \ncalQG & $b=\selQG$ & \calQG & \heldQG & \certQG\\
MATH-500 & gpt-4.1-mini & \MGM & \ncalGM & $b=\selGM$ & \calGM & \heldGM & \certGM\\
MATH-500 & Qwen2.5-7B & \MQM & \ncalQM & $b=\selQM$ & \calQM & \heldQM & \certQM\\
\bottomrule
\end{tabular}
\end{table}}

\paragraph{Anytime trajectories and optional stopping.}
Along the full GSM8K calibration stream, the \texttt{gpt-4.1-mini} certificate tightens smoothly from $\trajGGatNinetySix$ at $n=96$ to $\certGG$ at $n=\ncalGG$ ($\ratioCalGG\times$ calibration risk, matching held-out risk $\heldGG$), while \texttt{Qwen2.5-7B-Instruct} reaches $\certQG$ ($\ratioCalQG\times$ calibration risk; \cref{tab:scale,fig:audits}b). Both numbers align quantitatively with the $1/p$ scaling law of \cref{sec:radius}: exact inversion of~\eqref{eq:main} places the $2\times$-relative-tightness horizon at $n^\star=\nstarAtGG$ and $\nstarAtQG$, respectively, whereas on MATH-500 (where baseline error rates sit near $0.30$), just $n=\ncalGM$ examples suffice to bring both certificates within $\ratioCalGM\times$ of empirical risk ($\certGM$ and $\certQM$). Because validity holds at every prefix $n$, an engineer targeting a certified error ceiling of $0.15$ on GSM8K can legally halt calibration after $n=\stopPathGG$ examples on \texttt{gpt-4.1-mini} and $n=\stopPathQG$ on \texttt{Qwen2.5-7B-Instruct}.

\paragraph{Finite-grid confidence sequences versus PAC-Bayes posteriors.}
When the candidate space is a fixed finite grid of deterministic recipes, taking a union bound at level $\delta/M$ over sharp betting or empirical-Bernstein confidence sequences \citep[Thm.~2]{waudbysmith2024} also protects adaptive selection and optional stopping, yielding tighter terminal bounds on \cref{tab:scale}: $\ebGG$, $\ebQG$, $\ebGM$, and $\ebQM$ at $n_{\mathrm{cal}}$. Decomposing~\eqref{eq:main} pinpoints the exact source of this numerical gap. It does not stem from the PAC-Bayes change of measure: at a pre-committed sample size $n_{\mathrm{cal}}$, the fixed-horizon PAC-Bayes-kl bound paying only $\log(M/\delta)=\logMdeltaGG$ nats delivers $\klFixedGG$, $\klFixedQG$, $\klFixedGM$, and $\klFixedQM$, matching empirical-Bernstein across all four benchmarks. Rather, the gap comes almost entirely from our elementary closed-form time-uniformity and moment term $\pen{n}$ ($\timeTermGG$ nats at $n=\ncalGG$: $\timeUnifGG$ nats from the $\delta/[n(n+1)]$ union schedule plus $\momentGG$ nats from Maurer's $2\sqrt{n}$ moment bound). Replacing this term with stitched or coin-betting PAC-Bayes boundaries \citep{chugg2023,jang2023} compresses the finite-sample constant while preserving every post-selection guarantee of \cref{sec:certificate}. What the PAC-Bayes variational layer uniquely contributes is structural generality: it scores randomized posteriors via $\KL(Q\Vert P)$ (rewarding tie-breaking entropy), scales seamlessly to continuous or unbounded recipe spaces, composes additively across hierarchical families, and admits the closed-form $1/p$ sample-complexity law of~\eqref{eq:nstar}.

\subsection{Multi-dimensional sweeps, risk--cost frontiers, and hierarchical priors}\label{sec:grids}

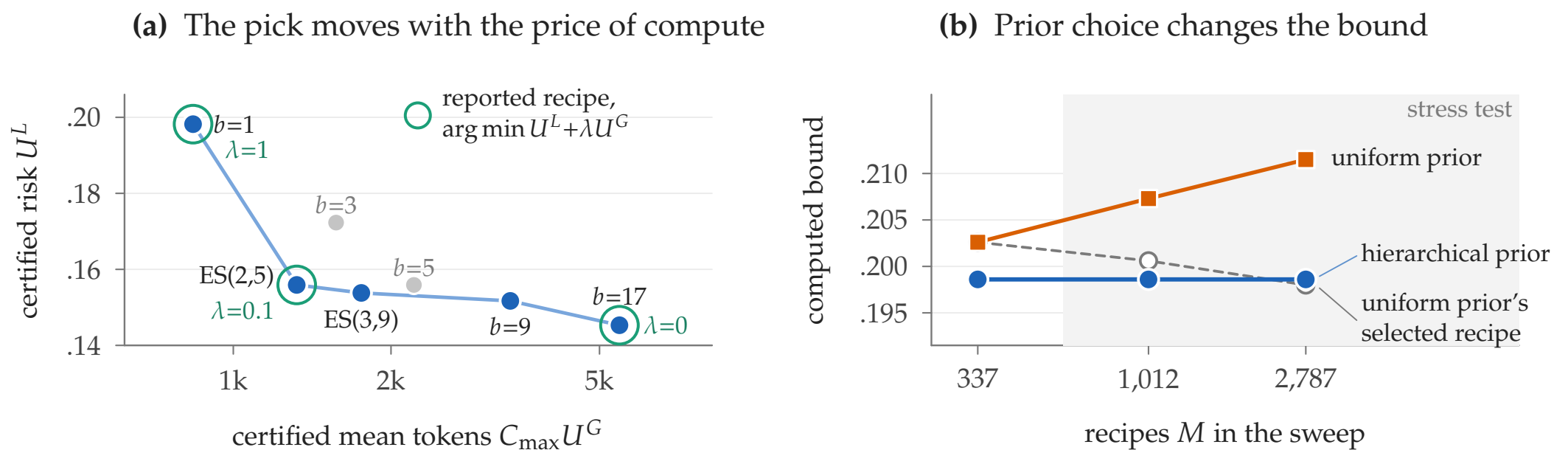
\begin{figure}[t]
\centering
\pgfplotstableread[row sep=crcr]{
UGtok UL tok cal held label front\\
837.9 0.1982 257.1 0.1017 0.0970 {b{=}1} 1\\
1321.4 0.1559 582.1 0.0698 0.0803 {ES(2,5)} 1\\
1571.1 0.1723 766.9 0.0819 0.0848 {b{=}3} 0\\
1755.4 0.1538 908.8 0.0683 0.0727 {ES(3,9)} 1\\
2211.8 0.1559 1276.6 0.0698 0.0833 {b{=}5} 0\\
3378.3 0.1517 2297.5 0.0668 0.0727 {b{=}9} 1\\
5457.9 0.1453 4337.9 0.0622 0.0727 {b{=}17} 1\\
}{\tblFrontierQwengsm}%
\pgfplotstableread[row sep=crcr]{
M flataudit flatsame loc flatcal flatheld\\
337 0.2026 0.2026 0.1986 0.0520 0.0760\\
1012 0.2006 0.2073 0.1986 0.0480 0.0560\\
2787 0.1980 0.2115 0.1986 0.0440 0.0720\\
}{\tblEloc}%
\begin{tikzpicture}[figbase]
\begin{axis}[ygridplot, name=pc, scale only axis, width=150pt, height=64pt,
  xmode=log, xmin=620, xmax=8200, ymin=0.140, ymax=0.206,
  xtick={1000,2000,5000}, xticklabels={1k,2k,5k},
  ytick={0.14,0.16,0.18,0.20}, yticklabels={.14,.16,.18,.20},
  xlabel={certified mean tokens $\Cmax U^G$}, ylabel={certified risk $U^L$},
  title={\ptitle{a}{The pick moves with the price of compute}}]
  \addplot[CoolMid, line width=0.9pt, unbounded coords=discard, y filter/.expression={\thisrow{front}==1 ? y : nan}]
    table[x=UGtok, y=UL] {\tblFrontierQwengsm};
  \addplot[only marks, mark=*, mark size=2.3pt, mark options={fill=LightGray, draw=white, line width=0.6pt},
    unbounded coords=discard, y filter/.expression={\thisrow{front}==0 ? y : nan}]
    table[x=UGtok, y=UL] {\tblFrontierQwengsm};
  \addplot[only marks, mark=*, mark size=2.6pt, mark options={fill=CoolInk, draw=white, line width=0.7pt},
    unbounded coords=discard, y filter/.expression={\thisrow{front}==1 ? y : nan}]
    table[x=UGtok, y=UL] {\tblFrontierQwengsm};
  \node[font=\tiny, text=MidGray, anchor=south, inner sep=2pt] at (axis cs:1571.1,0.1723) {$b{=}3$};
  \node[font=\tiny, text=MidGray, anchor=south, inner sep=2pt] at (axis cs:2211.8,0.1559) {$b{=}5$};
  \node[circle, inner sep=0pt, minimum size=9.6pt, draw=SageInk, line width=0.75pt] at (axis cs:837.9,0.1982) {};
  \node[circle, inner sep=0pt, minimum size=9.6pt, draw=SageInk, line width=0.75pt] at (axis cs:1321.4,0.1559) {};
  \node[circle, inner sep=0pt, minimum size=9.6pt, draw=SageInk, line width=0.75pt] at (axis cs:5457.9,0.1453) {};
  \node[font=\tiny, text=Ink, anchor=west, inner sep=1pt, xshift=4pt] at (axis cs:837.9,0.1982) {$b{=}1$};
  \node[font=\tiny, text=Ink, anchor=south east, inner sep=1pt, xshift=-4.5pt, yshift=-2.2pt] at (axis cs:1321.4,0.1559) {ES(2,5)};
  \node[font=\tiny, text=Ink, anchor=north, inner sep=1pt, yshift=-3pt] at (axis cs:1755.4,0.1538) {ES(3,9)};
  \node[font=\tiny, text=Ink, anchor=north, inner sep=1pt, yshift=-3pt] at (axis cs:3378.3,0.1517) {$b{=}9$};
  \node[font=\tiny, text=Ink, anchor=south, inner sep=1pt, yshift=4pt] at (axis cs:5457.9,0.1453) {$b{=}17$};
  \node[font=\tiny, text=SageInk!80!black, anchor=north west, inner sep=1pt, xshift=7pt, yshift=-3pt] at (axis cs:837.9,0.1982) {$\lambda{=}1$};
  \node[font=\tiny, text=SageInk!80!black, anchor=north east, inner sep=1pt, xshift=-4.5pt, yshift=-3pt] at (axis cs:1321.4,0.1559) {$\lambda{=}0.1$};
  \node[font=\tiny, text=SageInk!80!black, anchor=west, inner sep=1pt, xshift=5pt] at (axis cs:5457.9,0.1453) {$\lambda{=}0$};
  \node[circle, inner sep=0pt, minimum size=6.4pt, draw=SageInk, line width=0.75pt] (key) at (axis cs:2250,0.2005) {};
  \node[font=\tiny, text=Ink, anchor=west, inner sep=1pt, xshift=1.5pt, align=left] at (key.east) {reported recipe,\\$\arg\min U^L{+}\lambda U^G$};
\end{axis}
\begin{axis}[ygridplot, name=pd, at={(pc.south east)}, anchor=south west, xshift=56pt,
  scale only axis, width=150pt, height=64pt,
  xmode=log, xmin=250, xmax=11000, ymin=0.1915, ymax=0.2185,
  xtick={337,1012,2787}, xticklabels={{\elocMa},{\elocMb},{\elocMc}},
  ytick={0.195,0.200,0.205,0.210}, yticklabels={.195,.200,.205,.210},
  xlabel={recipes $M$ in the sweep}, ylabel={computed bound},
  title={\ptitle{b}{Prior choice changes the bound}}]
  \fill[FaintGray] (axis cs:584,0.1915) rectangle (axis cs:11000,0.2185);
  \node[font=\tiny, text=MidGray, anchor=north east, inner sep=1.5pt] at (axis cs:11000,0.2185) {stress test};
  \addplot[MidGray, line width=0.7pt, dash pattern=on 2.2pt off 1.6pt, mark=*, mark size=2.1pt,
    mark options={solid, fill=white, draw=MidGray, line width=0.7pt}] table[x=M, y=flataudit] {\tblEloc}
    coordinate[pos=1] (endW);
  \addplot[WarmInk, line width=1.0pt, mark=square*, mark size=2.2pt,
    mark options={fill=WarmInk, draw=white, line width=0.6pt}] table[x=M, y=flatsame] {\tblEloc}
    node[pos=1, anchor=west, font=\tiny, text=Ink, inner sep=1pt, xshift=5pt] {uniform prior};
  \addplot[CoolInk, line width=1.1pt, mark=*, mark size=2.6pt,
    mark options={fill=CoolInk, draw=white, line width=0.7pt}] table[x=M, y=loc] {\tblEloc}
    coordinate[pos=1] (endL);
\end{axis}
\node[font=\tiny, text=Ink, anchor=west, inner sep=1pt] (ll) at ([xshift=13pt, yshift=6pt]endL) {hierarchical prior};
\node[font=\tiny, text=Ink, anchor=west, inner sep=1pt, align=left] (lw) at ([xshift=13pt, yshift=-9pt]endW) {uniform prior's\\selected recipe};
\draw[CoolInk!70, line width=0.35pt] ([xshift=3.5pt, yshift=0.8pt]endL) -- (ll.west);
\draw[MidGray, line width=0.35pt] ([xshift=3.5pt, yshift=-0.8pt]endW) -- (lw.west);
\end{tikzpicture}%
\caption{\textbf{Pricing compute shifts the optimal recipe; hierarchical priors insulate defaults from wide exploration.} \textbf{(a)}~Joint 95\% risk--cost certificates ($\alpha_L=\alpha_G=\delta/2$) across the $\MQG$ Qwen recipes on GSM8K ($n=\ncalQG$); green rings highlight the minimizer of $U^L+\lambda U^G$ as the token price $\lambda$ varies. \textbf{(b)}~Controlled prior ablation expanding exploratory families from $M=\elocMa$ to $\elocMc$ recipes (shaded): a hierarchical prior keeps the default-family winner's bound constant at $\elocLoc$, whereas a flat uniform prior penalizes it continuously (\cref{tab:loc}).}\label{fig:frontier}
\end{figure}

\paragraph{Recipe grids and the certified Pareto frontier.}
Sweeping $M=\gridMQGrid$ heterogeneous Qwen recipes on GSM8K ($n=250$; three prompt templates $\times$ two temperatures $\times$ four budgets, plus early stopping, length routing, and best-of-3 verification) selects \gridSelQGrid{} with a 95\% certificate of $\gridCertQGrid$ (held-out risk $\gridHeldQGrid$), while the 25-recipe \texttt{gpt-4.1-mini} sweep selects \gridSelGGrid{} at $\gridCertGGrid$. On the Qwen grid, configurations from all three prompt templates finish within $\gridSpreadQGrid$ empirical error of the winner, and deploying the uniform randomized mixture over the $\mixKQGrid$ tied leaders tightens the certificate to $\mixCertQGrid$. Incorporating token cost via \cref{cor:joint} exposes a rich trade-off (\cref{fig:frontier}a): on the full Qwen GSM8K stream, early stopping $\mathrm{ES}(3,9)$ matches the held-out accuracy of $b=17$ while consuming just \frontQGesCostShare\% of its tokens, and varying the deployer's compute price $\lambda$ in $\arg\min\{U^L+\lambda U^G\}$ shifts the certified optimum smoothly along the frontier from \pickLamZero{} ($\lambda=0$) through \pickLamTenth{} ($\lambda=0.1$) to \pickLamOne{} ($\lambda=1$).

\paragraph{Insulating default pipelines with hierarchical priors.}
To isolate how prior architecture governs selection penalties as exploration scales, we conduct a controlled prior ablation on the recorded Qwen calibration split (\cref{rem:repairs}). We partition $M=\elocMa$ recipes across six $(\text{template},\text{temperature})$ families, assigning prior weight $\pi_0=0.5$ to the default family (template P0 at temperature $0.7$, $\elocMdefault$ recipes) and $0.1$ to each of five exploratory families, and then progressively augment the exploratory families with seeded trace-subset variants up to $M=\elocMc$ (\cref{fig:frontier}b, \cref{tab:loc}). Under \cref{cor:structured}, the bound for the default-family winner remains strictly invariant at $\elocLoc$ regardless of how many thousands of exploratory configurations are tested alongside it. Under a flat uniform prior, by contrast, the bound on that same baseline configuration degrades steadily from $\elocFlatSamea$ to $\elocFlatSamec$, while the uniform sweep's empirical winner drifts into noisy exploratory subsets whose apparent calibration error drops from $\elocFlatCala$ to $\elocFlatCalc$ even as $\log M$ inflates the penalty.

\subsection{Paired ranking and input-adaptive compute allocation}\label{sec:ranking}

\begin{remark}[Certifying relative rankings via common random numbers]\label[remark]{rem:ranking}
Bounding a recipe's absolute risk $L(a)$ and proving $L(a)<L(a')$ have fundamentally different noise scales. Because all candidates are evaluated on the same problems with shared trace prefixes, paired empirical-Bernstein confidence sequences on per-example loss differences \citep[Thm.~2]{waudbysmith2024} cancel shared problem difficulty and shrink at the rate of \emph{disagreement} between $a$ and $a'$ (\cref{app:paired}). On Qwen, paired sequences certify that $b=9$ strictly outperforms $b=1$ after just $n=\sepNQG$ examples on GSM8K and $n=\sepNQM$ on MATH-500---a more-than-twentyfold reduction over requiring non-overlapping marginal certificates ($n\approx\margSepQG$ and $\margSepQM$). Where true accuracy differences are tiny, as on \texttt{gpt-4.1-mini}, paired sequences appropriately refrain from declaring a winner.
\end{remark}

\paragraph{Water-filling compute allocation and evaluation cost.}
Finally, we evaluate the two-stage plug-in water-filling allocator of \cref{rem:plugin} on Qwen/GSM8K, using unlabeled traces from the second split to calibrate $\lambda$ for a target mean budget and evaluating labeled errors on the calibration split (\cref{rem:repairs}). At target budget $9$, adaptive water-filling commits $\wfErrWfNine$ errors at an average of $\wfBudgetNine$ traces per problem, edging out uniform $b=9$ ($\wfErrUniNine$ errors; identical design-set error rate $\wfHeldWfNine$) and tightening the four-recipe comparative bound from $\wfULUniNine$ to $\wfULWfNine$ while saving \wfTokSave\% of tokens. At target budget $5$, however, the $b_0=3$ pilot stage consumes most of the budget upfront, leaving too little headroom to reallocate and yielding calibration risk $\wfRiskFive$ versus $\uniRiskFive$ for uniform $b=5$, confirming that the pilot stage $b_0$ must sit well below the target average budget. Across the entire study, trace sharing keeps calibration remarkably economical: generating all \nGenQwen{} local Qwen traces took $\gpuMinMeasured$ GPU-minutes on a single NVIDIA L40S, and all \nGenGpt{} \texttt{gpt-4.1-mini} API calls (\apiTokM M tokens) cost \$\apiCost{} (\cref{app:recipes}).

\section{Related work}\label{sec:related}

\begin{table}[t]
\centering\small
\caption{\textbf{Scope of statistical guarantees for test-time scaling.} Adaptive inference pipelines require protection across post-selection reporting, optional stopping, ground-truth labeled risk (\cref{sec:sc}), and general randomized or continuous recipe spaces. Weighted unions of confidence sequences cover the first three axes on countable grids (\cref{sec:scale}); our anytime-valid PAC-Bayes-kl certificate covers all four.}\label{tab:protects}
\setlength{\tabcolsep}{2.6pt}
\begin{tabular}{@{}lcccc@{}}
\headtop
\headstrut \textbf{statistical guarantee} & \bfseries \makecell{post-\\selection} & \bfseries \makecell{optional\\stopping} & \bfseries \makecell{labeled\\risk} & \bfseries \makecell{uncountable\\recipe spaces}\\
\headbottom
fixed-recipe interval (Wilson, Hoeffding) & \xmark & \xmark & \cmark & \xmark\\
union bound; learn-then-test; conformal LM \cite{angelopoulos2025ltt,quach2024} & \cmark & \xmark & \cmark & \xmark\\
weighted union of anytime confidence sequences \cite{howard2021,waudbysmith2024} & \cmark & \cmark & \cmark & \xmark\\
per-instance unlabeled mode certificate \cite{corderoencinar2025,ota2026cite} & \xmark & \cmark & \xmark & \xmark\\
fixed-sample PAC-Bayes-kl bound \cite{seeger2002,maurer2004} & \cmark & \xmark & \cmark & \cmark\\
\textbf{anytime-valid PAC-Bayes-kl} (\cref{thm:main}) & \cmark & \cmark & \cmark & \cmark\\
\bottomrule
\end{tabular}
\end{table}

\paragraph{Test-time compute scaling.}
Chain-of-thought prompting, self-consistency, step- and outcome-level verifiers, tree search, early stopping, and query routing treat inference computation as an elastic resource traded against solution quality \citep{wei2022,cobbe2021,wang2023,lightman2023,li2023,yao2023,aggarwal2023,li2024esc,chen2023frugal,brown2024,snell2024,wu2024}. Recent theoretical work has studied scaling laws and aggregation mechanics \citep{chen2024calls,chen2025,zhou2025}, constrained policy optimization for budget allocation \citep{zhai2026}, risk control against early-stopping disagreement with full-budget voting \citep{chen2026mars}, and per-instance sequential guarantees that plurality voting recovers the model's modal answer without labels \citep{corderoencinar2025,ota2026cite}. Building on the multinomial large-deviation exponents of \citet{corderoencinar2025}, \cref{prop:margin} proves the complementary converse for wrong modes (establishing why unlabeled stability cannot bound task error) and derives our exact water-filling allocation law (\cref{thm:alloc}), while \cref{thm:main} certifies the labeled population risk and token cost of whichever pipeline an adaptive sweep ultimately selects (\cref{tab:protects}).

\paragraph{Post-selection and anytime-valid inference.}
Reusing evaluation data for model selection is a classic source of invalid inference across selective statistics \citep{berk2013}, adaptive data analysis \citep{dwork2015,russo2016}, and distribution-free risk control \citep{vovk2005,bates2021}. Learn-then-test and conformal language modeling \citep{angelopoulos2025ltt,quach2024} calibrate discrete hyperparameter grids at a fixed sample size, whereas time-uniform confidence sequences \citep{howard2021,waudbysmith2024} permit optional stopping and, via weighted union bounds, adaptive selection over countable sets. Our framework lifts anytime post-selection validity from countable grids to randomized posteriors over general recipe spaces, handles heavy-tailed uncapped costs in light of Bahadur--Savage impossibility \citep{bahadur1956} (\cref{cor:envelope}), and pairs marginal certification with common-random-number confidence sequences for fast relative ranking (\cref{rem:ranking}).

\paragraph{PAC-Bayes, description length, and coding lower bounds.}
Our certificate synthesizes classical pillars of PAC-Bayesian and sequential analysis: McAllester's and Catoni's variational change of measure \citep{mcallester1999,catoni2007,germain2015,alquier2024}, Seeger's and Maurer's exact binary-kl binomial moment bound \citep{seeger2002,maurer2004} (extended here to all $n\ge1$), and the time-uniform stitching framework of \citet[Cors.~22 and~25]{chugg2023}. On the lower-bound side, the Minimum Description Length view of PAC-Bayes dates to \citet{blum2003} and \citet{grunwald2007}, and \citet[Thm.~4.4]{langford2005} showed that Occam bounds cannot be uniformly tightened on adversarial independent configurations. \Cref{thm:codes} elevates Langford's construction into a universal Kraft inequality and Shannon-entropy lower bound on the kl-radii of \emph{any} valid per-recipe certificate, while \cref{prop:lower} establishes that even unrestricted certificates inspecting the full calibration matrix must pay $\log M-O(\log n)$ on exchangeable grids.

\section{Discussion}\label{sec:discussion}

Adaptive tuning on a calibration set need not be treated as an informal heuristic that invalidates statistical guarantees, nor must practitioners quarantine a fresh split after every tweak. By declaring a single prior $P$ upfront, iterative pipeline engineering acquires an exact accounting system: every adaptive query, stopping decision, and randomized deployment $Q_\tau$ is valid simultaneously at price $\KL(Q_\tau\Vert P)$. As \cref{thm:codes} reveals, this penalty is governed by the Shannon entropy of the selection rule's own output distribution up to $O(\log n)$, providing a sharp certificate-level counterpart to mutual-information bounds on adaptive selection bias \citep{russo2016}. Three practical principles follow directly from the theory:
\begin{enumerate}[leftmargin=1.6em,label=\textup{(\arabic*)}]
\item \emph{Forecast with hierarchical priors:} concentrate prior mass on a trusted default family and place smaller weights on experimental expansions, so broad exploration never inflates the certificate of a winning baseline (\cref{cor:structured,fig:frontier}b).
\item \emph{Certify the frontier, let deployment price compute:} report joint $(U^L,U^G)$ bounds at $\alpha_L+\alpha_G=\delta$ so downstream operators can choose their own risk--latency operating point $\lambda$ without re-calibrating (\cref{cor:joint,fig:frontier}a).
\item \emph{Decouple absolute certification from relative ranking:} use PAC-Bayes certificates to bound deployment error and paired confidence sequences on shared traces to separate competing candidates at over twentyfold smaller sample sizes (\cref{rem:ranking}).
\end{enumerate}

Finally, the architecture of \cref{sec:certificate} is modular. Corollaries~\ref{cor:rule}--\ref{cor:joint} rely on \cref{thm:main} solely through simultaneous validity over $(n,Q)$, meaning that the variance-adaptive stitched and coin-betting PAC-Bayes processes of \citet[Cors.~14--15]{chugg2023} and \citet{jang2023} can substitute directly for~\eqref{eq:main} whenever tighter finite-sample constants outweigh the analytical simplicity of the binary-kl scaling law~\eqref{eq:nstar}. Paired with a lightweight held-out check against parser bugs or distribution shift, anytime-valid certification turns test-time compute tuning into an auditable engineering discipline.

\clearpage
\appendix
\crefalias{section}{appendix}\crefalias{subsection}{appendix}
\numberwithin{theorem}{section}
\addtocontents{toc}{\protect\addvspace{5pt}\protect\noindent{\protect\sffamily\protect\small\protect\bfseries\protect\color{muted}Appendix}\protect\par}
\noindent{\color{refcool}\rule{\linewidth}{1.6pt}}\par\vspace{8pt}
\noindent{\sffamily\bfseries\LARGE\color{navy}Appendix}\par\vspace{3pt}
\noindent{\sffamily\small\color{muted}Anytime-Valid PAC-Bayes Certificates for Adaptive Test-Time Scaling}\par\vspace{6pt}
\noindent{\color{rulegray}\rule{\linewidth}{0.5pt}}\par\vspace{4pt}
\suppressfloats[t]
The appendix is organized into four parts and concludes with scope and broader impact (\cref{app:limits}). \emph{Foundations:} \cref{app:notation} maps the logical dependencies among all theoretical results, summarizes notation, and provides a dictionary between test-time scaling engineering and PAC-Bayes theory. \emph{Theory:} \cref{app:tools} establishes the three core inequalities underlying our certificates and lower bounds; \cref{app:event} proves \cref{thm:main} and derives Corollaries~\ref{cor:rule}--\ref{cor:joint} together with the uncapped-cost bound (\cref{cor:envelope}); \cref{app:price} proves the coding and exchangeable-grid lower bounds (\cref{thm:codes,prop:lower}); and \cref{app:voting} proves the mode-amplification bounds (\cref{prop:margin}), the water-filling theorem (\cref{thm:alloc}), and the paired ranking construction (\cref{rem:ranking}). \emph{Practice:} \cref{app:practice} translates \cref{thm:codes} into prior-selection guidelines and analyzes inner Monte Carlo evaluation of stochastic recipes. \emph{Experiments:} \cref{app:recipes,app:results} document the full experimental protocol and extended empirical tables.

\section{Map of results, notation, and dictionary}\label{app:notation}

\begin{figure}[!htbp]
\centering
\begin{tikzpicture}[figbase, x=1pt, y=1pt,
  dn/.style={font=\scriptsize, inner xsep=2.5pt, inner ysep=2pt, minimum height=19pt},
  ext/.style={greybox, dn}, key/.style={keybox, dn}, rd/.style={coolbox, dn},
  lim/.style={warmbox, dn}, al/.style={sagebox, dn}]
  \node[ext, minimum width=72pt] (L1) at (40,112) {Lem.~\ref{lem:dv}\\ change of measure};
  \node[ext, minimum width=72pt] (L2) at (40,84) {Lem.~\ref{lem:maurer}\\ moment bound, all $n$};
  \node[ext, minimum width=72pt] (H) at (40,-8) {Chernoff bound};
  \node[ext, minimum width=72pt] (L3) at (40,-36) {Lem.~\ref{lem:binom}\\ binomial point mass};
  \node[ext, minimum width=70pt] (W) at (330,140) {empirical-Bernstein\\ confidence sequence};
  \node[key, minimum width=70pt] (T1) at (140,98) {Thm.~\ref{thm:main}\\ anytime-valid bound};
  \node[rd, minimum width=66pt] (C2) at (238,140) {Cor.~\ref{cor:rule}\\ selection and stopping};
  \node[rd, minimum width=66pt] (C4) at (238,112) {Cor.~\ref{cor:joint}\\ risk--cost};
  \node[rd, minimum width=66pt] (C3) at (238,84) {Cor.~\ref{cor:structured}\\ hierarchical prior};
  \node[rd, minimum width=66pt] (C5) at (238,56) {Cor.~\ref{cor:envelope}\\ unknown cost cap};
  \node[al, minimum width=66pt] (R11) at (238,28) {Rem.~\ref{rem:plugin}\\ plug-in recipe};
  \node[rd, minimum width=70pt] (R6) at (330,112) {Rem.~\ref{rem:ranking}\\ ranking};
  \node[lim, minimum width=70pt] (P9) at (140,-8) {Prop.~\ref{prop:margin}\\ vote margin};
  \node[al, minimum width=66pt] (T10) at (238,-8) {Thm.~\ref{thm:alloc}\\ water-filling};
  \node[lim, minimum width=70pt] (P8) at (140,-36) {Prop.~\ref{prop:lower}\\ $\log M$ is sharp};
  \node[lim, minimum width=66pt] (T7) at (238,-36) {Thm.~\ref{thm:codes}\\ certificates are codes};
  \draw[flow] (L1) -- (T1);
  \draw[flow] (L2) -- (T1);
  \draw[flow] (T1.east) -- (C2.west);
  \draw[flow] (T1.east) -- (C4.west);
  \draw[flow] (T1.east) -- (C3.west);
  \draw[flow] (T1.east) -- (C5.west);
  \draw[flow] (T1.east) -- (R11.west);
  \draw[flow] (C2) -- (C4);
  \draw[flow] (W) -- (R6);
  \draw[flow] (H) -- (P9);
  \draw[flow, dashed] (P9) -- (T10) node[midway, above, lab, inner sep=1.5pt] {surrogate};
  \draw[flow] (T10) -- (R11);
  \draw[flow] (L3) -- (P8);
  \draw[flow, rounded corners=3pt] (L3.south) -- (40,-54) -- (238,-54) -- (T7.south);
  \draw[flow, rounded corners=3pt] (C3.east) -- (330,84) -- (330,-36) -- (T7.east);
  \draw[flow, dashed] (T7.west) -- (P8.east);
  \node[lab, anchor=north, inner sep=1.5pt] at (189.5,-38.5) {any cert.};
\end{tikzpicture}%
\caption{\textbf{Logical dependencies among theoretical results.} Solid arrows $A\to B$ indicate that the proof of $B$ invokes $A$. Dashed arrows indicate structural links: \cref{prop:margin} supplies the exponential risk surrogate minimized in \cref{thm:alloc}, and \cref{prop:lower} extends the exchangeable specialization of \cref{thm:codes} to arbitrary full-data certificates. Paired ranking (\cref{rem:ranking}) builds on \citet[Thm.~2]{waudbysmith2024}.}\label{fig:dependency}
\end{figure}
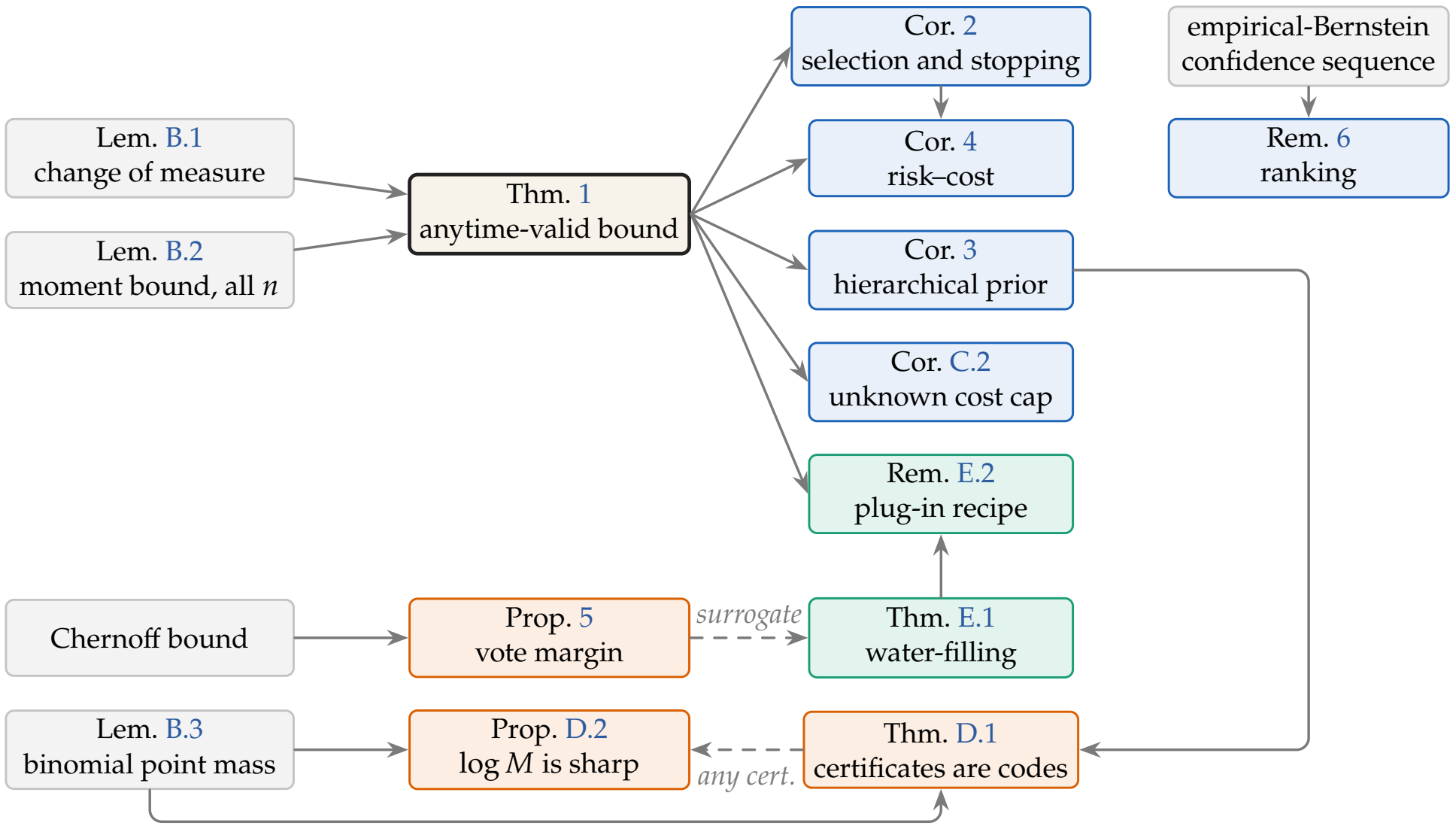

\Cref{fig:dependency} charts the deductive structure of the paper, and \cref{tab:dictionary} aligns practical test-time scaling concepts with their mathematical counterparts. Because \cref{thm:main} holds uniformly across all sample sizes $n\ge1$ and posteriors $Q\ll P$ on a single high-probability event, every adaptive selection and stopping corollary follows deterministically on that event.

\paragraph{Notation.}
Empirical risk $\Lhat_n$ and normalized cost $\Ghat_n$ follow \cref{def:transcript}. Binary relative entropy is $\kl(q\Vert p)=q\log\frac qp+(1-q)\log\frac{1-q}{1-p}$ (extended continuously to $[0,1]^2$), and Kullback--Leibler divergence is $\KL(Q\Vert P)=\int\log\frac{dQ}{dP}\,dQ$ when $Q\ll P$ and $+\infty$ otherwise. Its upper and lower Bernoulli inverses are $\klinv(\hat q,\eps)=\sup\{q\in[\hat q,1]:\kl(\hat q\Vert q)\le\eps\}$ and $\klinvlow(\hat q,\eps)=\inf\{q\in[0,\hat q]:\kl(\hat q\Vert q)\le\eps\}$. All logarithms are natural. For any $Q$-integrable functional $h:\cA\to\R$, we write $h(Q)=\E_{a\sim Q}h(a)$. At failure level $\alpha\in(0,1)$, the kl-radius is $\eps_n(Q;\alpha)=[\KL(Q\Vert P)+\penx{n}{\alpha}]/n$, the risk certificate is $U^L_n(Q;\alpha)=\klinv(\Lhat_n(Q),\eps_n(Q;\alpha))$, and the cost certificate $U^G_n(Q;\alpha)$ is defined identically with $\Ghat_n(Q)$.

\begin{table}[!htbp]
\centering\small
\caption{\textbf{Conceptual dictionary.} Practical test-time scaling notions (left) and their formal PAC-Bayes counterparts (right).}\label{tab:dictionary}
\begin{tabular}{@{}p{0.54\linewidth}p{0.42\linewidth}@{}}
\headtop
\headstrut \textbf{test-time scaling concept} & \textbf{formal object in this paper}\\
\headbottom
fully specified pipeline (prompt, temperature, budget, verifier, stopping rule, parser, tie-breaker) & recipe $a\in\cA$ (\cref{def:recipe})\\
stochastic tie-breaking or ensemble deployment & randomized recipe $Q\in\Delta(\cA)$ (\cref{def:search})\\
pre-committed sweep architecture & prior $P$, or family priors $\{P_j\}$ with weights $\pi_j$\\
flat hyperparameter sweep vs.\ protected baseline family & uniform prior vs.\ hierarchical prior (\cref{cor:structured})\\
monitoring validation curves and halting when satisfied & selection-and-stopping rule $(\tau,(Q_n))$ (\cref{def:rule})\\
statistical cost of adaptive tuning & selection penalty $\KL(Q\Vert P)$ (\cref{tab:ledger})\\
certified deployment error and token bounds & $(U^L_\tau(Q_\tau),\,U^G_\tau(Q_\tau))$ (\cref{cor:rule,cor:joint})\\
sharing generated traces across nested budgets & within-row coupling / common random numbers\\
\bottomrule
\end{tabular}
\end{table}

\FloatBarrier
\section{Auxiliary results}\label{app:tools}

Both our anytime certificate and its information-theoretic converse rest on three self-contained inequalities: Donsker--Varadhan variational duality to lift prior expectations uniformly to all posteriors (\cref{lem:dv}), Maurer's binomial moment bound extended to all $n\ge1$ (\cref{lem:maurer}), and a non-asymptotic binomial point-mass lower bound that converts coverage guarantees into a Kraft budget (\cref{lem:binom}).

\begin{lemma}[Variational change of measure; see \cite{catoni2007}]\label[lemma]{lem:dv}
Let $P$ be a probability measure on $\cA$, let $Q\ll P$, and let $f:\cA\to[0,\infty]$ be measurable with $\E_{a\sim P}e^{f(a)}<\infty$ (hence $f<\infty$ $P$-a.s.\ and $Q$-a.s.). Then $\E_{a\sim Q}f(a)\le\KL(Q\Vert P)+\log\E_{a\sim P}e^{f(a)}$.
\end{lemma}

\begin{lemma}[Exact binary-kl moment bound for all $n\ge1$]\label[lemma]{lem:maurer}
Let $X_1,\ldots,X_n$ be i.i.d.\ random variables in $[0,1]$ with mean $\mu$ and empirical mean $\bar X_n=\frac1n\sum_{i=1}^n X_i$. For every $n\ge1$,
\[
\E\exp\{n\kl(\bar X_n\Vert\mu)\}\;\le\;\xi(n)\;:=\;\sum_{k=0}^n\binom nk\Bigl(\frac kn\Bigr)^k\Bigl(1-\frac kn\Bigr)^{n-k}\;\le\;2\sqrt n.
\]
\end{lemma}
\begin{proof}
\emph{Reduction to Bernoulli laws.} Because $x\mapsto\exp\{n\kl(\bar x\Vert\mu)\}$ is the composition of a nondecreasing convex function with the convex map $x\mapsto n\kl(\bar x\Vert\mu)$, it is convex on $[0,1]^n$. By \citet[Lemma~3]{maurer2004}, the expectation of any convex function of i.i.d.\ $[0,1]$-valued variables is maximized by i.i.d.\ Bernoulli($\mu$) variables.

\emph{Evaluation on Bernoulli variables.} For $X_i\sim\mathrm{Bernoulli}(\mu)$ with $\mu\in(0,1)$, the parameter $\mu$ cancels identically:
\[
\E\exp\{n\kl(\bar X_n\Vert\mu)\}\;=\;\sum_{k=0}^n\binom nk\mu^k(1-\mu)^{n-k}\Bigl(\frac{k}{n\mu}\Bigr)^k\Bigl(\frac{n-k}{n(1-\mu)}\Bigr)^{n-k}\;=\;\xi(n),
\]
while for $\mu\in\{0,1\}$ the left-hand side equals $1\le\xi(n)$.

\emph{Verification of $\xi(n)\le2\sqrt n$ for all $n\ge1$.} Stirling's estimate $\xi(n)\le e^{1/(12n)}\sqrt{\pi n/2}+2$ \citep[Thm.~1, Eq.~(5)]{maurer2004} implies $\xi(n)\le2\sqrt n$ for all $n\ge8$. Direct computation confirms $\xi(n)\le2\sqrt n$ for $1\le n\le7$, with equality at $n=1$:
\[
\text{\small\begin{tabular}{@{}lccccccc@{}}
\headtop
\headstrut \textbf{$n$} & \textbf{1} & \textbf{2} & \textbf{3} & \textbf{4} & \textbf{5} & \textbf{6} & \textbf{7}\\
\headbottom
$\xi(n)$ & 2.000 & 2.500 & 2.889 & 3.219 & 3.510 & 3.775 & 4.018 \\
$2\sqrt n$ & 2.000 & 2.828 & 3.464 & 4.000 & 4.472 & 4.899 & 5.292 \\
\bottomrule
\end{tabular}
}\qedhere
\]
\end{proof}

\begin{lemma}[Non-asymptotic binomial point mass]\label[lemma]{lem:binom}
For every $n\ge1$, $0\le k\le n$, and $p\in(0,1)$, $\P\{\mathrm{Bin}(n,p)=k\}\ge e^{-n\kl(k/n\Vert p)}/\sqrt{2n}$.
\end{lemma}
\begin{proof}
At the endpoints $k\in\{0,n\}$, $\P\{\mathrm{Bin}(n,p)=k\}=e^{-n\kl(k/n\Vert p)}$ holds with prefactor $1\ge1/\sqrt{2n}$. For $0<k<n$, writing $h$ for binary Shannon entropy in nats,
\[
\binom nkp^k(1-p)^{n-k}=\binom nk\,e^{-nh(k/n)}\,e^{-n\kl(k/n\Vert p)},\qquad\binom nk\;\ge\;\frac{e^{nh(k/n)}}{\sqrt{8n\frac kn(1-\frac kn)}}\;\ge\;\frac{e^{nh(k/n)}}{\sqrt{2n}},
\]
where the first inequality is \citet[Sec.~17.5]{cover2006} and the second uses $\frac kn(1-\frac kn)\le\frac14$.
\end{proof}

\section{Anytime-valid certificates}\label{app:event}

\Cref{thm:main} combines \cref{lem:dv,lem:maurer} across calibration times via a convergent union schedule; \cref{cor:rule,cor:structured,cor:joint,cor:envelope} then follow directly from simultaneous coverage without further probabilistic arguments.

\subsection{The simultaneous concentration event (\texorpdfstring{Theorem~\ref*{thm:main}}{Main certificate})}
\begin{proof}
We instantiate the time-uniform PAC-Bayes recipe of \citet[Cors.~22 and~25]{chugg2023} with error allocation $\delta_n=\delta/[n(n+1)]$ at each sample size $n\ge1$.

\emph{Step 1: Fixed-$n$ prior expectation.} Define
\[
S_n\;=\;\E_{a\sim P}\exp\bigl\{n\kl\bigl(\Lhat_n(a)\Vert L(a)\bigr)\bigr\},
\]
which is $\cF_n$-measurable by joint measurability. For each fixed $a\in\cA$, the losses $(\ell_a(Z_i,U_{i,a}))_{i\le n}$ are i.i.d.\ in $[0,1]$ with mean $L(a)$, so Tonelli's theorem and \cref{lem:maurer} yield $\E S_n\le2\sqrt n$. Markov's inequality gives $\P\{S_n>2\sqrt n/\delta_n\}\le\delta_n$.

\emph{Step 2: Uniform lift to all posteriors.} On the event $\{S_n\le2\sqrt n/\delta_n\}$, applying \cref{lem:dv} to $f(a)=n\kl(\Lhat_n(a)\Vert L(a))$ yields, simultaneously for all $Q\ll P$,
\[
\E_{a\sim Q}\,n\kl\bigl(\Lhat_n(a)\Vert L(a)\bigr)\;\le\;\KL(Q\Vert P)+\log\frac{2\sqrt n}{\delta_n}.
\]

\emph{Step 3: Convexity of binary relative entropy.} Joint convexity of $\kl(\cdot\Vert\cdot)$ implies
\[
\kl\bigl(\Lhat_n(Q)\Vert L(Q)\bigr)\;=\;\kl\bigl(\E_Q\Lhat_n(a)\,\Vert\,\E_QL(a)\bigr)\;\le\;\E_Q\kl\bigl(\Lhat_n(a)\Vert L(a)\bigr).
\]

\emph{Step 4: Union over all calibration sizes.} Setting $\delta_n=\delta/[n(n+1)]$ gives $\sum_{n\ge1}\delta_n=\delta$ and $\log(2\sqrt n/\delta_n)=\pen{n}$. A union bound over $n\ge1$ places all events $\{S_n\le2\sqrt n/\delta_n\}$ onto a single event of probability at least $1-\delta$, on which Steps~2 and~3 establish~\eqref{eq:main} simultaneously for all $n\ge1$ and all $Q\ll P$.

Because $c_a\in[0,1]$, the proof for $(\Ghat_n,G)$ is identical. Observe that the argument uses only three structural properties: $P$ is independent of the calibration data (permitting Tonelli's theorem in Step~1), the observable is jointly measurable in the index and calibration row, and for each fixed index the observables are i.i.d.\ in $[0,1]$. \Cref{cor:envelope} exploits this generality on an enlarged index space.
\end{proof}

\begin{remark}[Small-sample exactness, input independence, and prospective vs.\ ablation protocols]\label[remark]{rem:repairs}
\textup{(i)}~\emph{Validity from the first example.} Extending $\xi(n)\le2\sqrt n$ to $1\le n\le7$ in \cref{lem:maurer} ensures that \cref{thm:main} holds non-asymptotically for every $n\ge1$.
\textup{(ii)}~\emph{Why $P$ cannot depend even on unlabeled calibration inputs.} Suppose $X$ has a continuous law, $Y\equiv0$, and recipe $a_S$ predicts $0$ at cost $0$ on the observed calibration inputs $S=\{X_1,\ldots,X_n\}$ while predicting $1$ at cost $\Cmax$ everywhere else. Declaring $P=\delta_{a_S}$ uses no labels, yet yields $\Lhat_n=\Ghat_n=0$ alongside $L=G=1$, violating the bound with probability one.
\textup{(iii)}~\emph{Prospective certifications versus controlled ablations in \cref{sec:experiments}.} Our main budget sweeps and multi-template grids (\cref{tab:scale,sec:grids}) are strictly prospective: the uniform prior $P$ enumerates all pre-generated configurations before calibration. By contrast, two supplementary experiments on the recorded Qwen calibration half isolate the effect of prior and grid design on a fixed dataset: the hierarchical-prior scaling ablation (\cref{fig:frontier}b, \cref{tab:loc}; mass $0.5$ on the default template--temperature family, expanded from $M=\elocMa$ to $\elocMc$ via seeded trace subsets to trace the $\log M$ growth of flat priors) and the four-recipe water-filling comparison (\cref{sec:ranking}). To mark that distinction transparently, we refer to values in those two ablations as computed bounds under the specified priors rather than prospective certificates.
\end{remark}

\subsection{Deriving adaptive selection, hierarchical, and joint bounds (\texorpdfstring{Corollaries~\ref*{cor:rule}--\ref*{cor:joint}}{Corollaries 2--4})}
\begin{proof}[Proof of \cref{cor:rule}]
On the event of \cref{thm:main}, inequality~\eqref{eq:main} holds simultaneously for all deterministic pairs $(n,Q)$ with $Q\ll P$. For any admissible rule and any sample path with $\tau<\infty$, the realized pair $(\tau,Q_\tau)$ belongs to that collection, so $\kl(\Lhat_\tau(Q_\tau)\Vert L(Q_\tau))\le\eps_\tau(Q_\tau)$. Because $q\mapsto\kl(\hat q\Vert q)$ is strictly increasing on $[\hat q,1]$, either $L(Q_\tau)\le\Lhat_\tau(Q_\tau)\le U^L_\tau(Q_\tau;\delta)$ or $L(Q_\tau)\le\klinv(\Lhat_\tau(Q_\tau),\eps_\tau(Q_\tau))=U^L_\tau(Q_\tau;\delta)$.
\end{proof}

\begin{proof}[Proof of \cref{cor:structured}]
Define the mixture prior $P_\pi=\sum_{j\in\cJ}\pi_jP_j$. Whenever $Q\ll P_j$, we have $Q\ll P_\pi$ and Radon--Nikodym derivative $dP_j/dP_\pi\le1/\pi_j$ $P_\pi$-a.s.\ (since $\int h\,dP_\pi\ge\pi_j\int h\,dP_j$ for all measurable $h\ge0$). The chain rule gives
\[
\KL(Q\Vert P_\pi)\;=\;\int\log\Bigl(\frac{dQ}{dP_j}\,\frac{dP_j}{dP_\pi}\Bigr)\,dQ\;\le\;\KL(Q\Vert P_j)+\log\frac1{\pi_j}.
\]
Applying \cref{thm:main} with prior $P_\pi$ yields~\eqref{eq:local} simultaneously across all $j\in\cJ$. For the Gibbs posterior $dQ_{n,j,\eta}/dP_j(a)=e^{-\eta n\Lhat_n(a)}/Z_{n,j}(\eta)$ with partition function $Z_{n,j}(\eta)=\E_{P_j}e^{-\eta n\Lhat_n}\in[e^{-\eta n},1]$, integrating $\log(dQ_{n,j,\eta}/dP_j)=-\eta n\Lhat_n-\log Z_{n,j}(\eta)$ against $Q_{n,j,\eta}$ yields the expression in \cref{tab:ledger}.
\end{proof}

\begin{proof}[Proof of \cref{cor:joint}]
Apply \cref{thm:main,cor:rule} to the loss stream at failure level $\alpha_L$ and to the cost stream at level $\alpha_G$, then combine the two events by a union bound ($\alpha_L+\alpha_G\le\delta$). Any measurable selection from $\arg\min_{Q\ll P}\{U^L_n(Q;\alpha_L)+\lambda U^G_n(Q;\alpha_G)\}$ is $\cF_n$-measurable and hence admissible.
\end{proof}

\subsection{Uncapped heavy-tailed costs (\texorpdfstring{Corollary~\ref*{cor:envelope}}{Cost without a known cap})}

When token generation is uncapped, cost distributions can be heavy-tailed, and guessing a synthetic cap $\Cmax$ is either invalid (if exceeded) or overly loose. Instead, one can certify clipped expectations and exceedance probabilities simultaneously across a dyadic threshold grid chosen \emph{after} observing the empirical tail.

\begin{corollary}[Dyadic clipped-cost and exceedance certificate]\label[corollary]{cor:envelope}
Fix a reference scale $C_0>0$ independently of the calibration data, dyadic thresholds $C_k=C_02^k$ for $k\ge0$, and prior weights $\pi_k=1/[(k+1)(k+2)]$. For each $k\ge0$, write $c^{(k)}_a=\min\{C_a/C_k,1\}$ for clipped normalized cost and $e^{(k)}_a=\one\{C_a>C_k\}$ for tail exceedance. With probability at least $1-\delta$, simultaneously for every $n\ge1$, every $k\ge0$, every $f\in\{c^{(k)},e^{(k)}\}$, and every $Q\ll P$,
\[
\kl\bigl(\widehat F_n(Q)\Vert F(Q)\bigr)\;\le\;\frac{\KL(Q\Vert P)+\log[(k+1)(k+2)]+\log2+\pen{n}}{n}.
\]
\end{corollary}

Selecting the threshold $C_k$ from the data adds at most $\log[(k+1)(k+2)]+\log2\le2\log(2+\log_2(C_k/C_0))+\log2$ nats. Reporting the pair $(\E\min\{C/C_k,1\},\,\P\{C>C_k\})$ is statistically canonical: for nonnegative random variables without a known upper support bound, no non-trivial distribution-free upper confidence bound on the unclipped mean $\E\,C$ exists \citep{bahadur1956}.

\begin{proof}
Consider the product index space $\cA\times\N_0\times\{\mathrm{clip},\mathrm{exc}\}$ endowed with the product prior $P\otimes\pi\otimes\mathrm{Unif}\{\mathrm{clip},\mathrm{exc}\}$, noting $\sum_{k\ge0}\pi_k=\sum_{k\ge0}(\frac1{k+1}-\frac1{k+2})=1$. Every indexed observable $c^{(k)}_a(Z_i,U_{i,a})$ or $e^{(k)}_a(Z_i,U_{i,a})$ takes values in $[0,1]$ and is i.i.d.\ across calibration rows $i$, so the final remark in the proof of \cref{thm:main} delivers simultaneous validity over the product space. Evaluating at the posterior $Q\otimes\delta_k\otimes\delta_t$ yields
\[
\KL\bigl(Q\otimes\delta_k\otimes\delta_t\,\Vert\,P\otimes\pi\otimes\mathrm{Unif}\{\mathrm{clip},\mathrm{exc}\}\bigr)\;=\;\KL(Q\Vert P)+\log[(k+1)(k+2)]+\log2,
\]
and linearity of $\widehat F_n$ and $F$ in $Q$ completes the proof.
\end{proof}

\paragraph{Impossibility of unclipped mean bounds without a cap.}
Let $\mathcal P$ denote the class of all distributions on $[0,\infty)$ with finite expectation, and suppose $\hat u(C_1,\ldots,C_n)$ satisfies $\inf_{\nu\in\mathcal P}\P_\nu\{\E_\nu C\le\hat u\}\ge1-\delta$ at some fixed $n$. Mixing any base law $\nu_0\in\mathcal P$ with mass $\eps\ll 1/n$ at an arbitrarily large atom $M$ perturbs the $n$-sample total-variation distance by at most $n\eps$ while driving $\E C\ge\eps M\to\infty$, forcing $\P_{\nu_0}\{\hat u=\infty\}\ge1-\delta$ \citep{bahadur1956}.

\section{The selection penalty cannot be avoided}\label{app:price}

We now prove the two lower bounds of \cref{sec:lower}. \Cref{thm:codes} shows that simultaneous validity across independent recipes imposes a Kraft inequality on per-recipe kl-radii, making Shannon entropy the fundamental lower bound on expected certificate radius; \cref{prop:lower} removes the per-recipe restriction on exchangeable grids.

\subsection{Simultaneous validity is a Kraft budget (\texorpdfstring{Theorem~\ref*{thm:codes}}{Certificates are codes})}

Consider product Bernoulli environments in which each recipe $a\in\cA$ incurs i.i.d.\ $\mathrm{Bernoulli}(L(a))$ losses independently across recipes, with arbitrary true risks $(L(a))_{a\in\cA}\in[0,1]^{\cA}$. A \emph{per-recipe certificate} assigns to each $a$ a bound $U_a=\phi_a(\ell_{1:n}(a))$ depending only on recipe $a$'s own loss sequence $\ell_{1:n}(a)\in\{0,1\}^n$; both point-mass PAC-Bayes bounds and arbitrary weighted unions of per-recipe confidence sequences belong to this class. Write $\bar\phi_a(x)$ for a median of $\phi_a$ under a uniform random permutation of $n x$ ones and $n(1-x)$ zeros, and define the worst-case kl-radius spent on recipe $a$ by
\[
r_a\;=\;\min\Bigl\{n\kl\bigl(x\Vert\bar\phi_a(x)\bigr):x\in\bigl\{0,\tfrac1n,\ldots,1\bigr\},\ \bar\phi_a(x)<1\Bigr\},
\]
with the conventions $\kl(x\Vert\bar\phi_a(x))=0$ whenever $\bar\phi_a(x)\le x$ and $\min\emptyset=\infty$. Note why post-selection validity demands simultaneous coverage $\P\{\forall a,\,L(a)\le U_a\}\ge1-\delta$: if that event fails with probability exceeding $\delta$ under some product law, an omniscient stopping-and-selection rule at fixed horizon $\tau=n$ can simply output a recipe $a$ whose bound failed ($U_a<L(a)$).

\begin{theorem}[Certificates are codes]\label[theorem]{thm:codes}
Let $U=(U_a)_{a\in\cA}$ be any per-recipe certificate satisfying $\P\{L(a)\le U_a\text{ for all }a\}\ge1-\delta$, with $\delta\in(0,\frac12]$, under every independent Bernoulli loss law.
\begin{enumerate}[leftmargin=1.6em,label=\textup{(\roman*)}]
\item \emph{Kraft inequality:} Up to an $O(\log n)$ sample-resolution offset, the radii $(r_a)$ satisfy Kraft's inequality: $\sum_{a\in\cA}e^{-r_a}\le4\sqrt{2n}\,\delta$.
\item \emph{Achievability:} Conversely, for every probability mass function (code) $q$ on $\cA$, the point-mass certificates of \cref{thm:main} with prior $q$ are valid across all loss laws and spend $r_a=\log(1/q_a)+\pen{n}$.
\item \emph{Entropy benchmark and redundancy:} Under any fixed reporting rule whose selected recipe $\hat a$ has marginal law $\rho$, $\E\,r_{\hat a}\ge H(\rho)+\log(1/\delta)-\log(4\sqrt{2n})$, where $H(\rho)$ is Shannon entropy. Choosing $q=\rho$ in~\textup{(ii)} achieves $\E\,r_{\hat a}=H(\rho)+\pen{n}$, whereas any mismatched prior $q$ pays the additional redundancy $\KL(\rho\Vert q)$.
\end{enumerate}
\end{theorem}

\begin{proof}
\emph{Part~(i), Step~1: Adversarial risk assignment.} Let $S\subset\cA$ be any finite subset of recipes with $r_a<\infty$. Following the tightness construction of \citet[Thm.~4.4]{langford2005}, choose for each $a\in S$ an empirical mean $x_a\in\{0,\frac1n,\ldots,1\}$ with $\bar\phi_a(x_a)<1$ and a true risk
\[
p_a\in\bigl(\max\{\bar\phi_a(x_a),x_a\},\,1\bigr)\quad\text{if }x_a<1,\qquad p_a\in\bigl(\max\{\bar\phi_a(1),0\},\,1\bigr)\quad\text{if }x_a=1,
\]
setting $p_a=0$ for all $a\notin S$.

\emph{Step~2: Per-recipe failure probability.} For each $a\in S$, the miscoverage event $F_a=\{\phi_a(\ell_{1:n}(a))<p_a\}$ contains $\{\Lhat_n(a)=x_a\}\cap\{\phi_a(\ell_{1:n}(a))\le\bar\phi_a(x_a)\}$. Conditioned on $\Lhat_n(a)=x_a$, the i.i.d.\ Bernoulli sequence $\ell_{1:n}(a)$ is uniformly distributed over its permutations, so the second event occurs with conditional probability at least $\frac12$ by definition of the median $\bar\phi_a(x_a)$. Applying \cref{lem:binom} gives
\[
\alpha_a\;:=\;\P(F_a)\;\ge\;\tfrac12\,\P\{\Lhat_n(a)=x_a\}\;\ge\;\frac{e^{-n\kl(x_a\Vert p_a)}}{2\sqrt{2n}}.
\]

\emph{Step~3: Simultaneous coverage enforces a sum budget.} Because $(F_a)_{a\in S}$ depend on mutually independent loss sequences, simultaneous coverage requires
\[
1-\delta\;\le\;\prod_{a\in S}(1-\alpha_a)\;\le\;\exp\Bigl(-\sum_{a\in S}\alpha_a\Bigr)
\quad\Longrightarrow\quad
\sum_{a\in S}\alpha_a\;\le\;\log\frac1{1-\delta}\;\le\;2\delta
\]
for all $\delta\in(0,\frac12]$. Substituting Step~2 yields $\sum_{a\in S}e^{-n\kl(x_a\Vert p_a)}\le4\sqrt{2n}\,\delta$.

\emph{Step~4: Limiting radii.} Letting $p_a\downarrow\max\{\bar\phi_a(x_a),x_a\}$ when $x_a<1$ (or $p_a\uparrow1$ when $x_a=1$), continuity of $\kl(x_a\Vert\cdot)$ sends $n\kl(x_a\Vert p_a)\to n\kl(x_a\Vert\bar\phi_a(x_a))$. Choosing the minimizing $x_a$ for each $a\in S$ and taking the supremum over all finite subsets $S\subset\cA$ establishes~(i).

\emph{Part~(ii).} Validity holds by \cref{thm:main} with prior $q$. Because $U_a$ depends on $\ell_{1:n}(a)$ solely through $\Lhat_n(a)$, it is permutation-invariant ($\bar\phi_a=\phi_a$). If $q_a=0$, then $U_a\equiv1$ and $r_a=\infty$. If $q_a>0$, then $0<\eps_n(\delta_a)<\infty$; for every $x<1$, continuity and strict monotonicity of $\kl(x\Vert\cdot)$ on $[x,1)$ give $x<U_a<1$ with $n\kl(x\Vert U_a)=n\eps_n(\delta_a)$, while at $x=1$ we have $U_a=1$. Thus $r_a=n\eps_n(\delta_a)=\log(1/q_a)+\pen{n}$.

\emph{Part~(iii).} By~(i), $\tilde q_a=e^{-r_a}/(4\sqrt{2n}\,\delta)$ defines a sub-probability measure on $\cA$. Nonnegativity of Kullback--Leibler divergence (Gibbs' inequality) yields
\[
\E\,r_{\hat a}\;=\;\sum_{a\in\cA}\rho(a)\log\frac1{\tilde q_a}-\log\bigl(4\sqrt{2n}\,\delta\bigr)\;\ge\;H(\rho)-\log\bigl(4\sqrt{2n}\,\delta\bigr),
\]
while under~(ii), $\E\,r_{\hat a}=\sum_a\rho(a)\log(1/q_a)+\pen{n}=H(\rho)+\KL(\rho\Vert q)+\pen{n}$.
\end{proof}

\subsection{Arbitrary full-data certificates on exchangeable grids (\texorpdfstring{Proposition~\ref*{prop:lower}}{Exchangeable-recipe lower bound})}

\begin{proposition}[Sharp $\log M$ cost of empirical minimization]\label[proposition]{prop:lower}
Let $n\ge20$ and $2\le M\le e^{n/8}$, and suppose $M$ recipes incur independent $\mathrm{Bernoulli}(\frac12)$ losses on each calibration example. Let $\hat a\in\arg\min_a\Lhat_n(a)$ be an empirically optimal recipe, and let $U$ be \emph{any} measurable function of the full $M\times n$ calibration data satisfying $\P\{L(\hat a)\le U\}\ge1-\delta$. Then for every $\eta\in(0,\frac12]$, with probability at least $1-\delta-\eta$,
\[
n\,\kl\bigl(\Lhat_n(\hat a)\,\Vert\,\min\{U,1\}\bigr)\;\ge\;\log M-\tfrac12\log n-\log\bigl(4\sqrt2\log(1/\eta)\bigr).
\]
\end{proposition}

\begin{proof}
Realize the $M$ recipes as input-independent fair coin flips evaluated against an independent fair binary label $Y$, so that $n\Lhat_n(a)\sim\mathrm{Bin}(n,\frac12)$ i.i.d.\ across $a\in\{1,\ldots,M\}$ and $L(a)\equiv\frac12$. Set
\[
B\;=\;\log M-\log\bigl(\sqrt{2n}\log(1/\eta)\bigr),
\qquad
K\;=\;\bigl\{k<n/2:\ n\kl(k/n\Vert\tfrac12)\le B\bigr\},
\]
noting $B\le\log M\le n/8$ since $\sqrt{2n}\log(1/\eta)\ge1$. If $K=\emptyset$, then $B<n\kl(\frac{\lceil n/2\rceil-1}n\Vert\frac12)\le4/n<\log4$, making the claimed lower bound $B-\log4$ negative and hence trivial since $\kl\ge0$. Otherwise, let $k^\star=\min K$ and $t^\star=k^\star/n$.

\emph{Step~1: Order-statistic anti-concentration.} Because $n\kl(t^\star\Vert\frac12)\le B$, \cref{lem:binom} implies $\P\{\Lhat_n(a)\le t^\star\}\ge e^{-B}/\sqrt{2n}=\log(1/\eta)/M$ for each recipe $a$. By independence across the $M$ recipes,
\[
\P\Bigl\{\min_{1\le a\le M}\Lhat_n(a)>t^\star\Bigr\}\;\le\;\Bigl(1-\frac{\log(1/\eta)}{M}\Bigr)^M\;\le\;\eta.
\]

\emph{Step~2: Discretization step at $t^\star$.} Since $n\kl(0\Vert\frac12)=n\log2>n/8\ge B$, we have $k^\star\ge1$ and $k^\star-1\notin K$, meaning $n\kl(s\Vert\frac12)>B$ at $s=t^\star-\frac1n$. Convexity of $x\mapsto\kl(x\Vert\frac12)$ (with derivative $\log\frac{x}{1-x}$) gives
\[
n\bigl[\kl(s\Vert\tfrac12)-\kl(t^\star\Vert\tfrac12)\bigr]\;\le\;\log\frac{1-s}{s}.
\]
By Pinsker's inequality, $2(\frac12-t^\star)^2\le\kl(t^\star\Vert\frac12)\le B/n\le\frac18$, so $t^\star\ge\frac14$ and $s\ge\frac15$ for all $n\ge20$. Thus $\log\frac{1-s}{s}\le\log4$, yielding $n\kl(t^\star\Vert\frac12)>B-\log4$.

\emph{Step~3: High-probability intersection.} On the event $\{\min_a\Lhat_n(a)\le t^\star\}\cap\{L(\hat a)\le U\}$, which holds with probability at least $1-\delta-\eta$, we have $\Lhat_n(\hat a)\le t^\star<\frac12=L(\hat a)\le\min\{U,1\}$. Monotonicity of $\kl(x\Vert u)$ on $x\le u$ gives
\[
n\kl\bigl(\Lhat_n(\hat a)\,\Vert\,\min\{U,1\}\bigr)\;\ge\;n\kl\bigl(t^\star\Vert\tfrac12\bigr)\;>\;B-\log4.
\]
Moreover, since $\kl(t\Vert\frac12)\le4(t-\frac12)^2$, empirical selection optimism on this event satisfies $\frac12-\Lhat_n(\hat a)\ge\frac12-t^\star>\frac12\sqrt{(B-\log4)/n}$ whenever $B>\log4$.
\end{proof}

\section{Voting, allocation, and ranking}\label{app:voting}

We next prove the mode-amplification inequalities (\cref{prop:margin}), solve the continuous-budget water-filling optimization (\cref{thm:alloc}), and detail the paired empirical-Bernstein ranking sequence (\cref{rem:ranking}). Notation follows \cref{sec:sc}, with $B(u,v)$ denoting the Bhattacharyya exponent and $B$ the average trace budget.

\subsection{Plurality mode concentration (\texorpdfstring{Proposition~\ref*{prop:margin}}{Voting amplifies the mode})}
\begin{proof}
Fix an input $x$, abbreviate $p=p_x$, and let $A_1,\ldots,A_b\sim p$ be $b$ i.i.d.\ sampled answers.

\emph{Step~1: Pairwise score differences.} For any competitor $a\ne y$, set $S^a_i=\one\{A_i=a\}-\one\{A_i=y\}$. Whenever plurality fails to output $y$ (under any tie-breaking rule), at least one incorrect answer $a\ne y$ ties or outvotes $y$, so $\sum_{i=1}^b S^a_i\ge0$.

\emph{Step~2: Exact Chernoff saddlepoint.} For any $\lambda\ge0$,
\[
\E e^{\lambda S^a_1}\;=\;1-p(a)-p(y)+p(a)e^\lambda+p(y)e^{-\lambda}.
\]
When $p(y)\ge p(a)>0$, minimizing over $\lambda\ge0$ sets $e^{2\lambda}=p(y)/p(a)$, where $\E e^{\lambda S^a_1}=1-(\sqrt{p(y)}-\sqrt{p(a)})^2=e^{-B(p(y),p(a))}$ (which also holds by continuity at $p(a)=0$). Markov's inequality yields
\[
\P\Bigl\{\sum_{i=1}^b S^a_i\ge0\Bigr\}\;\le\;e^{-bB(p(y),p(a))}.
\]

\emph{Step~3: Forward and converse bounds.} Because $v\mapsto B(p(y),v)$ decreases on $[0,p(y)]$ and every wrong answer satisfies $p(a)\le p(a_2)$, a union bound over the at most $m-1$ wrong answers gives $\P\{\SC_b(x)\ne y\}\le(m-1)e^{-bs_x}$. Averaging over $X\sim\cD$ and bounding the conditional error by $1$ whenever $\Delta_X\le0$ (or when $(m_X-1)e^{-bs_X}>1$) yields the population risk inequality. Conversely, if a wrong answer $a^\star\ne y$ is the strict mode, recovering $y$ requires $\sum_{i=1}^b(\one\{A_i=y\}-\one\{A_i=a^\star\})\ge0$; exchanging $y$ and $a^\star$ in Step~2 bounds this event by $e^{-bB(p(a^\star),p(y))}$.

\emph{Step~4: Margin lower bound on the exponent.} Using $(\sqrt u-\sqrt v)^2=(u-v)^2/(\sqrt u+\sqrt v)^2\ge(u-v)^2/[2(u+v)]$ together with $-\log(1-z)\ge z$ gives $B(u,v)\ge\frac{(u-v)^2}{2(u+v)}\ge\frac{(u-v)^2}{2}$ whenever $u+v\le1$.
\end{proof}

\subsection{Oracle water-filling allocation (\texorpdfstring{Theorem~\ref*{thm:alloc}}{Oracle allocation})}

Let $A_x=m_x-1$ and let $s_x=B(p_x(y),p_x(a_2))$ on $\{\Delta_x>0\}$ as in \cref{prop:margin}.

\begin{theorem}[Oracle compute allocation is water-filling]\label[theorem]{thm:alloc}
Assume $\E\,m_X<\infty$. Over all measurable relaxed budget schedules $\beta(x)\ge1$ satisfying $\E\beta(X)\le B$, the exponential risk surrogate $\P\{\Delta_X\le0\}+\E[A_Xe^{-s_X\beta(X)}\one\{\Delta_X>0\}]$ is minimized by
\[
\beta_\lambda(x)\;=\;\begin{cases}
\max\bigl\{1,\,s_x^{-1}\log(A_xs_x/\lambda)\bigr\}, & \text{if }\Delta_x>0,\ A_x>0,\text{ and }s_x<\infty,\\[2pt]
1, & \text{otherwise},
\end{cases}
\]
where $\lambda>0$ is chosen so that $\E\beta_\lambda(X)=B$ whenever $B>1$ and $\P\{\Delta_X>0,\,A_X>0,\,s_X<\infty\}>0$ (and $\beta\equiv1$ otherwise). Rounding $\beta_\lambda(x)$ up to the nearest odd integer produces an integer vote schedule with mean budget at most $B+2$ and no larger surrogate risk.
\end{theorem}

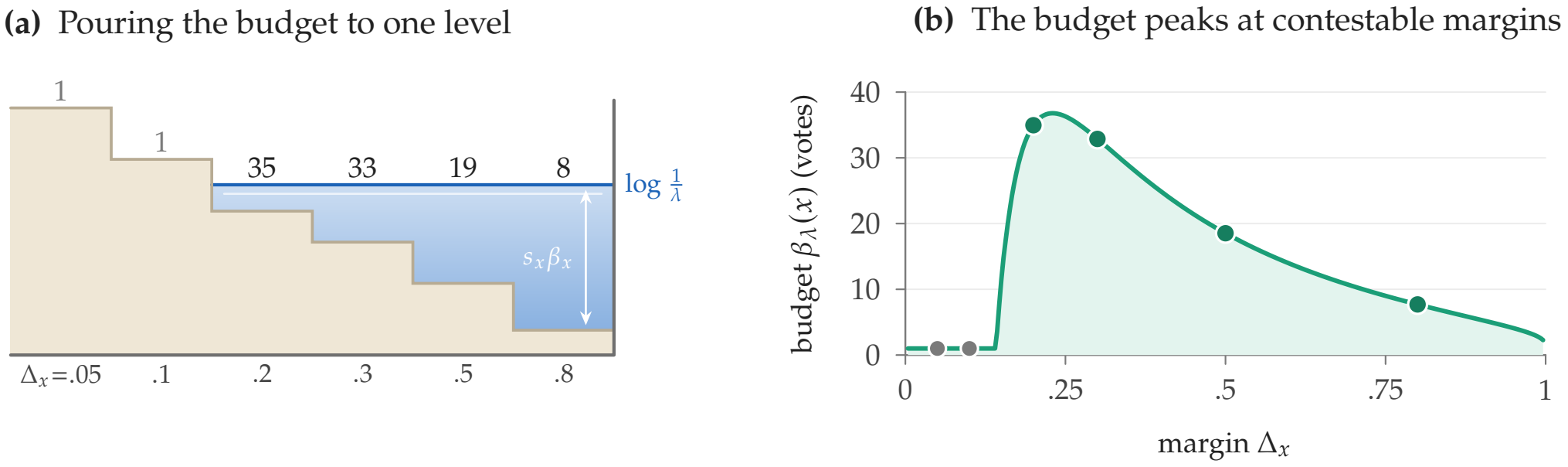
\begin{figure}[t]
\centering
\definecolor{Sand}{HTML}{EFE7D6}
\definecolor{SandEdge}{HTML}{B7AB93}
\begin{tikzpicture}[figbase, x=1pt, y=1pt,
  declare function={sb(\d)=-ln(1-(sqrt((1+\d)/2)-sqrt((1-\d)/2))^2);
                    budget(\d,\l)=max(1, ln(sb(\d)/\l)/sb(\d));}]
\def\u{24.5}    %
\def\k{9}       %
\pgfmathsetmacro{\lev}{4.605*\k}
\foreach \i/\f in {0/6.683,1/5.293,2/3.892,3/3.054,4/1.939,5/0.672}{
  \pgfmathsetmacro{\xl}{\i*\u}\pgfmathsetmacro{\xr}{(\i+1)*\u}\pgfmathsetmacro{\yt}{\f*\k}
  \fill[Sand] (\xl,0) rectangle (\xr,\yt);
}
\shade[top color=CoolMid!38!white, bottom color=CoolMid!88]
  (2*\u,\lev) -- (6*\u,\lev) -- (6*\u,0.672*\k) -- (5*\u,0.672*\k) -- (5*\u,1.939*\k) -- (4*\u,1.939*\k)
  -- (4*\u,3.054*\k) -- (3*\u,3.054*\k) -- (3*\u,3.892*\k) -- (2*\u,3.892*\k) -- cycle;
\draw[white, opacity=0.7, line width=0.5pt] (2*\u+3,\lev-2.2) -- (6*\u-3,\lev-2.2);
\draw[CoolInk, line width=0.8pt] (2*\u,\lev) -- (6*\u,\lev);
\draw[SandEdge, line width=0.7pt, line join=miter]
  (0,6.683*\k) -- (\u,6.683*\k) -- (\u,5.293*\k) -- (2*\u,5.293*\k) -- (2*\u,3.892*\k) -- (3*\u,3.892*\k)
  -- (3*\u,3.054*\k) -- (4*\u,3.054*\k) -- (4*\u,1.939*\k) -- (5*\u,1.939*\k) -- (5*\u,0.672*\k) -- (6*\u,0.672*\k);
\draw[Ink!65, line width=0.8pt] (0,0) -- (6*\u,0) -- (6*\u,62);
\node[font=\scriptsize, text=MidGray, anchor=south, inner sep=1.5pt] at (0.5*\u,6.683*\k) {1};
\node[font=\scriptsize, text=MidGray, anchor=south, inner sep=1.5pt] at (1.5*\u,5.293*\k) {1};
\node[font=\scriptsize, text=Ink, anchor=south, inner sep=1.5pt] at (2.5*\u,\lev) {35};
\node[font=\scriptsize, text=Ink, anchor=south, inner sep=1.5pt] at (3.5*\u,\lev) {33};
\node[font=\scriptsize, text=Ink, anchor=south, inner sep=1.5pt] at (4.5*\u,\lev) {19};
\node[font=\scriptsize, text=Ink, anchor=south, inner sep=1.5pt] at (5.5*\u,\lev) {8};
\draw[<->, >={Stealth[length=1.2mm,width=1mm]}, white, line width=0.5pt] (5.72*\u,0.672*\k+1.2) -- (5.72*\u,\lev-1.2);
\node[font=\tiny, text=white, anchor=east, inner sep=1pt] at (5.66*\u,2.6*\k) {$s_x\beta_x$};
\foreach \i/\m in {0/{\Delta_x{=}.05},1/{.1},2/{.2},3/{.3},4/{.5},5/{.8}}
  \node[font=\tiny, text=Ink!85, anchor=north, inner sep=2pt] at (\i*\u+0.5*\u,0) {$\m$};
\node[font=\tiny, text=CoolInk, anchor=west, inner sep=1.5pt] at (6*\u+1,\lev) {$\log\frac1\lambda$};
\node[paneltitle, anchor=base west] at (-2,78.25) {\ptitle{a}{Pouring the budget to one level}};
\begin{axis}[ygridplot, at={(218pt,0)}, anchor=south west,
  scale only axis, width=156pt, height=64pt,
  xmin=0, xmax=1, ymin=0, ymax=40, xtick={0,0.25,0.5,0.75,1}, xticklabels={0,.25,.5,.75,1},
  ytick={0,10,20,30,40},
  xlabel={margin $\Delta_x$}, ylabel={budget $\beta_\lambda(x)$ (votes)},
  title={\ptitle{b}{The budget peaks at contestable margins}}]
  \addplot[draw=none, fill=SageFill, domain=0.004:0.996, samples=300] {budget(x,0.01)} \closedcycle;
  \addplot[SageInk, line width=1.1pt, domain=0.004:0.996, samples=300] {budget(x,0.01)};
  \addplot[only marks, mark=*, mark size=2.3pt, mark options={fill=SageInk!80!black, draw=white, line width=0.6pt},
    samples at={0.2,0.3,0.5,0.8}] {budget(x,0.01)};
  \addplot[only marks, mark=*, mark size=2.1pt, mark options={fill=MidGray, draw=white, line width=0.6pt},
    samples at={0.05,0.1}] {budget(x,0.01)};
\end{axis}
\end{tikzpicture}%
\caption{\textbf{Geometry of oracle water-filling} (\cref{thm:alloc}; binary answer laws at shadow price $\lambda=0.01$). \textbf{(a)}~Each input sits on a floor of height $\log(1/(A_xs_x))$, and compute is poured up to the common water surface $\log(1/\lambda)$; the submerged depth equals $s_x\beta_\lambda(x)$, so the allocated vote budget is depth divided by $s_x$, while elevated floors above the surface keep the minimum $b=1$. \textbf{(b)}~Resulting vote schedule $\beta_\lambda(x)$ versus margin $\Delta_x$: budgets remain at $1$ near $\Delta_x\approx0$, peak sharply across contestable margins, and taper off once single traces are already reliable.}\label{fig:waterfill}
\end{figure}

At the optimum, every input allocated $\beta_\lambda(x)>1$ equalizes its marginal surrogate reduction $-\frac{d}{db}[A_xe^{-s_xb}]=A_xs_xe^{-s_x\beta_\lambda(x)}$ to the shadow price $\lambda$ (\cref{fig:waterfill}). Inputs with $\Delta_x\le0$ (wrong or tied mode), tiny positive margins ($A_xs_xe^{-s_x}\le\lambda$), or deterministic right answers ($s_x=\infty$) receive the minimum single trace, concentrating test-time compute on intermediate, contestable problems.

\begin{proof}
Write $S=\{x:\Delta_x>0,\,A_x>0,\,s_x<\infty\}$. Outside $S$, the objective is invariant to $\beta(x)$, so setting $\beta(x)=1$ off $S$ is optimal without consuming excess budget.

On $S$, for any Lagrange multiplier $\lambda>0$, the pointwise Lagrangian $h_{x,\lambda}(b)=A_xe^{-s_xb}+\lambda b$ ($b\ge1$) is strictly convex and coercive with derivative $\lambda-A_xs_xe^{-s_xb}$, uniquely minimized on $[1,\infty)$ by $\beta_\lambda(x)$. Moreover, $h_{x,\lambda}(\beta_\lambda(x))\le h_{x,\lambda}(1)$ implies $\lambda\beta_\lambda(x)\le\lambda+A_xe^{-s_x}\le\lambda+A_x$, so $\beta_\lambda(x)\le1+A_x/\lambda$ is integrable since $\E A_X\le\E m_X<\infty$.

Define $g(\lambda)=\E\beta_\lambda(X)$. By monotone and dominated convergence, $g$ is continuous and nonincreasing on $(0,\infty)$, with $\lim_{\lambda\to\infty}g(\lambda)=1$, and $\lim_{\lambda\downarrow0}g(\lambda)=\infty$ whenever $\P(S)>0$. Thus for any $B>1$ with $\P(S)>0$, there exists $\lambda>0$ with $g(\lambda)=B$. Pointwise optimality $A_Xe^{-s_X\beta(X)}+\lambda\beta(X)\ge A_Xe^{-s_X\beta_\lambda(X)}+\lambda\beta_\lambda(X)$ integrated over $X$ together with $\E\beta(X)\le B=\E\beta_\lambda(X)$ proves global optimality (with $\beta\equiv1$ trivial when $B=1$ or $\P(S)=0$). Finally, rounding $\beta_\lambda(x)$ up to the smallest odd integer $\bar b_\lambda(x)\le\beta_\lambda(x)+2$ cannot increase $e^{-s_xb}$.
\end{proof}

\begin{remark}[Certifying a deployable two-stage plug-in allocator]\label[remark]{rem:plugin}
Any adaptive rule that estimates margin parameters from an initial pilot batch of traces on input $X$ and then samples the remaining traces is itself a valid test-time recipe (\cref{def:recipe}) with random, input-dependent token cost $C_a(X,U)$. Our two-stage rule draws $b_0=3$ pilot traces, plugs in empirical vote-count estimates for $(\Delta_x,s_x,m_x)$ (\cref{app:recipes}), and allocates total traces via \cref{thm:alloc}, yielding a self-contained pipeline certified directly by \cref{cor:joint} (\cref{sec:ranking}).
\end{remark}

\subsection{Paired empirical-Bernstein sequences for ranking (\texorpdfstring{Remark~\ref*{rem:ranking}}{Paired ranking})}\label{app:paired}

To test whether recipe $a'$ strictly improves upon recipe $a$ when both run on shared calibration problems and coupled seeds, form the rescaled paired loss difference
\[
D_i\;=\;\tfrac12\bigl(1+\ell_a(Z_i,U_i)-\ell_{a'}(Z_i,U_i)\bigr)\in\bigl\{0,\tfrac12,1\bigr\},
\]
which is i.i.d.\ with mean $\mu_D=\frac12(1+L(a)-L(a'))$, so $L(a)>L(a')$ if and only if $\mu_D>\frac12$.

We apply the predictable plug-in empirical-Bernstein confidence sequence of \citet[Thm.~2]{waudbysmith2024}: defining
\[
\hat\mu_t=\frac{\frac12+\sum_{i\le t}D_i}{t+1},\qquad\hat\sigma^2_t=\frac{\frac14+\sum_{i\le t}(D_i-\hat\mu_i)^2}{t+1},\qquad\lambda_t=\min\Bigl\{\sqrt{\frac{2\log(2/\alpha)}{\hat\sigma^2_{t-1}\,t\log(1+t)}},\,\frac12\Bigr\},
\]
predictable variance increments $v_i=4(D_i-\hat\mu_{i-1})^2$,\footnote{Note the range-scaling factor $4$ required on $[0,1]$ by \citet[Thm.~2]{waudbysmith2024}; omitting that factor understates empirical variance and produces anti-conservative stopping times.} and $\psi_E(\lambda)=(-\log(1-\lambda)-\lambda)/4$, the running intersection over $t\ge1$ of
\[
\Bigl(\frac{\sum_{i\le t}\lambda_iD_i}{\sum_{i\le t}\lambda_i}\;\pm\;\frac{\log(2/\alpha)+\sum_{i\le t}v_i\psi_E(\lambda_i)}{\sum_{i\le t}\lambda_i}\Bigr)
\]
forms a simultaneous $(1-\alpha)$ confidence sequence for $\mu_D$. Whereas marginal certificates for true risks $0.02$ and $0.04$ ($M=4$) require $n\approx\margSepA$ examples to stop overlapping, the paired width scales with $\mathrm{Var}(D_i)$, collapsing whenever the two recipes agree on most inputs.

\paragraph{Streaming ranking protocol.}
In our ranking study (\cref{rem:ranking}), we declare $L(a)>L(a')$ at the first example $t$ where the lower confidence bound strictly exceeds $\frac12$, dividing $\alpha=0.05$ evenly across predeclared comparison pairs and streaming each recorded dataset in published order. Across 2{,}000 synthetic streams of length 2{,}000 under four baseline disagreement laws (including sparse-disagreement regimes), empirical familywise miscoverage stays between $\csCovMin$ and $\csCovMax$ at nominal $\alpha=0.05$. On the LLM benchmarks, Qwen's $b=9$ is certified superior to $b=1$ at $n=\sepNQG$ on GSM8K and $n=\sepNQM$ on MATH-500, while \texttt{gpt-4.1-mini}'s much smaller accuracy gap remains unseparated within the $1{,}319$ GSM8K and $500$ MATH-500 problems.

\FloatBarrier
\section{Choosing the prior and evaluating stochastic recipes}\label{app:practice}

\subsection{Principles for choosing the prior}\label{app:prior}

The prior $P$ encodes the practitioner's pre-calibration sweep design. By \cref{thm:codes}(iii), if a selection procedure reports $\hat a\sim\rho$, choosing $P$ to approximate $\rho$ minimizes the redundancy $\KL(\rho\Vert P)$. Four concrete architectures suffice for almost all test-time scaling workflows:
\begin{enumerate}[leftmargin=1.6em]
\item \emph{Small finite grids: uniform prior.} On modest sweeps ($M\lesssim30$) at realistic calibration sizes, the sample-concentration term $\Theta(\log n/n)$ dominates $\log M/n$ (\cref{tab:nstar}: expanding $M=4\to24$ shifts $n^\star$ by only \nstarGrowMin--\nstarGrowMax\%), making a flat prior simple and nearly lossless.
\item \emph{Unbounded or multiscale ladders: heavy-tailed complexity tilt.} Over countably infinite budget ladders $b\ge1$, a power-law prior $P(b)\propto(1+b)^{-2}$ charges $2\log(1+b)+\log(\pi^2/6-1)$ nats regardless of how high $b_{\max}$ is pushed: testing larger budgets is free unless a larger budget actually wins. \Cref{tab:prior} illustrates the analogous log-budget prior $\propto(1+\log_2 b)^{-2}$, tightening the zero-error certificate at $b=1$ from $\priorUniCheap$ to $\priorCplxCheap$ in exchange for a mild surcharge at $b=65$.
\item \emph{Structured multi-family sweeps: hierarchical priors (\cref{cor:structured}).} When testing experimental prompts, temperatures, or verifiers against a production baseline, assign $\pi_0=0.5$ to the default family and split the remaining $0.5$ across exploratory families. If the baseline wins, exploratory expansion costs nothing (\cref{tab:loc}); if an exploratory recipe wins, it pays only $\log(1/\pi_j)$ extra (for example, on our $\gridMQGrid$-recipe Qwen grid with $\pi_0=0.5$ on P0 and $0.25$ each on P1 and P2, selecting a P1 winner gives $\gridLocCertQGrid$ versus $\gridCertQGrid$ under a flat prior).
\item \emph{Support alignment.} Ensure $Q\ll P$ with finite $\KL(Q\Vert P)$: assign positive atoms $P(a)>0$ to any candidate that may be deployed deterministically, or deploy smooth posteriors (such as Gibbs mixtures) when $P$ is continuous. Any revision to $P$ motivated by calibration results must be certified on fresh calibration data.
\end{enumerate}

\begin{table}[!htbp]
\centering\small
\caption{\textbf{Effect of prior tilt across a budget ladder} ($n=256$, zero empirical errors, $\delta=0.05$), comparing a uniform prior on $\{1,3,5,9,17,33,65\}$ against a log-complexity prior $P(b)\propto(1+\log_2b)^{-2}$.}\label{tab:prior}
\begin{tabular}{@{}rcccc@{}}
\headtop
\headstrut \textbf{budget} & \textbf{uniform mass} & \textbf{complexity mass} & \textbf{uniform certificate} & \textbf{complexity certificate}\\
\headbottom
1 & 0.143 & 0.723 & 0.073 & 0.067\\
3 & 0.143 & 0.108 & 0.073 & 0.074\\
5 & 0.143 & 0.065 & 0.073 & 0.076\\
9 & 0.143 & 0.042 & 0.073 & 0.078\\
17 & 0.143 & 0.028 & 0.073 & 0.079\\
33 & 0.143 & 0.020 & 0.073 & 0.080\\
65 & 0.143 & 0.015 & 0.073 & 0.082\\
\bottomrule
\end{tabular}

\end{table}

\subsection{Inner Monte Carlo evaluation of stochastic recipes}\label{app:mc}

In \cref{def:transcript}, a single realized execution $(H_a(X_i,U_{i,a}),C_a(X_i,U_{i,a}))$ per calibration problem suffices to certify the marginal deployment risk $L(a)=\E_{Z,U}\ell_a(Z,U)$, because deployment itself runs recipe $a$ once per query. If an analyst additionally wishes to estimate the input-conditional expected loss $\ell^\star_a(Z_i)=\E_U[\ell_a(Z_i,U)\mid Z_i]$ by averaging $r$ independent executions $\bar\ell_{i,a}=r^{-1}\sum_{j=1}^r\ell_a(Z_i,U_{i,a,j})$ on each of the $n$ problems, the outer generalization gap still scales with the number of independent problems $n$, while for fixed $(n,r)$ on an $M$-recipe grid Hoeffding's inequality bounds the inner Monte Carlo discrepancy with probability at least $1-\eta$ simultaneously across all $a$ by
\[
\Bigl|\frac1n\sum_{i\le n}\bar\ell_{i,a}-\frac1n\sum_{i\le n}\ell^\star_a(Z_i)\Bigr|\;\le\;\sqrt{\frac{\log(2M/\eta)}{2nr}}.
\]

\FloatBarrier
\section{Experimental protocol}\label{app:recipes}

\paragraph{Models and decoding.}
We queried \texttt{gpt-4.1-mini} via the OpenAI Chat Completions API and served \texttt{Qwen2.5-7B-Instruct} locally in \texttt{bfloat16} with vLLM~0.25.1 on a single NVIDIA L40S GPU. Unless swept, decoding used temperature $0.7$ and top-$p$ $0.95$, with deterministic per-request seeds hashed from the problem index, prompt template, temperature, and trace index. Hard per-trace generation caps were set to 512 tokens on GSM8K and 1{,}024 tokens on MATH-500. The 192-example GSM8K pilot (\cref{tab:scale}, row~1; \cref{app:pilot}) used the same model, temperature, budget grid, token cap, and tie-breaker as the full \texttt{gpt-4.1-mini} sweep, differing in three recipe coordinates (\cref{def:recipe}): a concise system prompt (\emph{``Reason briefly and put the final numeric answer on a line formatted exactly as: Final answer: $\langle\text{number}\rangle$''}), a parser prioritizing the \texttt{Final answer:} line, and top-$p$ $1.0$.

\paragraph{Prompt templates.}
On GSM8K we evaluated three zero-shot chain-of-thought templates:
\begin{itemize}[leftmargin=1.5em]
\item \textbf{P0 (default):} \emph{``Solve the following grade school math problem. Think step by step, then write the final numeric answer on the last line in the exact form `\texttt{\#\#\#\#} $\langle\text{answer}\rangle$'.''}
\item \textbf{P1:} \emph{``You are a careful math tutor. Solve the problem step by step, keeping the reasoning brief, and end your reply with a line of the exact form `\texttt{\#\#\#\#} $\langle\text{final numeric answer}\rangle$'.''}
\item \textbf{P2:} \emph{``Read the problem and compute the answer. Show your key steps. The last line of your reply must be `\texttt{\#\#\#\#} $\langle\text{number}\rangle$' and nothing else.''}
\end{itemize}
On MATH-500 we used: \emph{``Solve the following math problem. Reason step by step, and put your final answer inside \texttt{\textbackslash boxed\{\}}.''} Each template was followed immediately by the problem statement.

\paragraph{Answer extraction and voting.}
GSM8K answers were extracted as the last number following \texttt{\#\#\#\#}, else following \emph{``answer is''}, else the final number in the trace, normalized to a canonical numeric string. MATH-500 answers were extracted from the final \texttt{\textbackslash boxed\{\}} expression under standard \LaTeX{} normalization (stripping whitespace, \texttt{\textbackslash left}/\texttt{\textbackslash right}, trailing units or percent signs, normalizing \texttt{\textbackslash dfrac}, and canonicalizing numerals); ground-truth answers underwent the identical normalizer. Plurality voting ignored unparsed traces, broke ties deterministically by lexicographic order (a rule independent of trace order), and scored an all-unparsed ballot as an error.

\paragraph{Candidate recipe families.}
\emph{Self-consistency} at budget $b$ takes the plurality vote over the first $b$ of $b_{\max}$ shared traces ($b_{\max}=17$ for Qwen on full GSM8K; $b_{\max}=9$ elsewhere). \emph{Early stopping} $\mathrm{ES}(\text{lead},\text{cap})$ draws traces sequentially and halts as soon as the front-runner leads the runner-up by \emph{lead} votes or \emph{cap} traces are reached, charging the exact tokens consumed up to stopping. The \emph{length router} assigns $b=9$ to prompts of at least 60 words and $b=1$ otherwise. \emph{Verifier best-of-3} feeds the first three P0 traces to the same model at temperature $0$, parses \texttt{BEST: $k$}, and adds the judge's tokens to total cost. \emph{Two-stage water-filling} draws $b_0=3$ pilot traces, estimates $\hat\Delta_x=(v_1-v_2)/3$ from the top two vote counts and $\hat m_x=\max\{\#\text{distinct answers},2\}$, sets $\hat s_x=\max\{\hat\Delta_x,0.2\}^2/2$ and $\hat A_x=\hat m_x-1$, and rounds $\beta_\lambda(x)$ of \cref{thm:alloc} up to an odd integer in $[3,17]$, calibrating $\lambda$ by bisection on unlabeled traces from the complementary split to match the target mean budget (\cref{rem:repairs}).

\paragraph{Sweep catalogs.}
The $\gridMQGrid$-recipe Qwen grid on the first 500 GSM8K problems ($250/250$ split) crosses templates $\{\text{P0},\text{P1},\text{P2}\}\times\text{temperatures }\{0.7,1.0\}\times\text{budgets }\{1,3,5,9\}$ alongside $\mathrm{ES}(3,9)$, the length router, and best-of-3 verification on $(\text{P0},0.7)$; the $\gridMGGrid$-recipe \texttt{gpt-4.1-mini} grid omits the router and verifier. In the hierarchical-prior ablation (\cref{tab:loc}, \cref{rem:repairs}), the default family $(\text{P0},0.7)$ comprises $\elocMdefault$ recipes (budgets $\{1,3,5,7,9,11,13,17\}$ across trace offsets plus nine early-stopping rules), while each of the five exploratory $(\text{template},\text{temperature})$ families comprises 54 recipes (budgets $1\text{--}9$ with offsets and nine early-stopping rules), totaling $M=\elocMa$; the $M=\elocMb$ and $\elocMc$ stress tests append seeded trace subsets of sizes $2\text{--}9$ within the exploratory families.

\paragraph{Compute footprint and reproducibility.}\label{app:repro}
All \nGenQwen{} local Qwen generations required $\gpuMinMeasured$ GPU-minutes on one L40S GPU (plus a few minutes for temperature-0 verification and container smoke tests), and all \nGenGpt{} \texttt{gpt-4.1-mini} API calls (\apiTokM M tokens) cost \$\apiCost{}. Every certificate, figure, and table in the paper is regenerated in minutes on a standard CPU from the archived raw traces and seeds via two self-contained build scripts (one writing all main-text \LaTeX{} macros, one building all appendix tables, with standalone scripts reproducing the pilot and synthetic stress tests bit-for-bit).

\FloatBarrier
\section{Additional results}\label{app:results}

\subsection{Complete per-recipe results on full-scale benchmarks}
\Cref{tab:full} reports empirical risks, held-out risks, 95\% certificates, and mean token counts for every candidate across the four full-scale evaluations of \cref{tab:scale}; \cref{tab:grid} lists the top eight configurations on each multi-template recipe grid.
\begin{table}[!htbp]
\centering\small
\caption{Per-recipe breakdown across all four full-scale benchmarks ($\delta=0.05$, uniform prior; \cref{tab:scale} reports the minimum-calibration-risk recipe with token-cost tie-breaking).}\label{tab:full}
\begin{tabular}{@{}llcccr@{}}
\headtop
\headstrut \textbf{evaluation} & \textbf{recipe} & \textbf{cal.\ risk} & \textbf{held-out risk} & \textbf{certificate} & \textbf{mean tokens}\\
\headbottom
\multirow{4}{*}{GSM8K, gpt-4.1-mini} & $b{=}1$ & 39/659 $=0.059$ & 0.052 & 0.138 & 162\\
 & $b{=}3$ & 27/659 $=0.041$ & 0.041 & 0.112 & 485\\
 & $b{=}5$ & 26/659 $=0.039$ & 0.039 & 0.109 & 808\\
 & $b{=}9$ & 26/659 $=0.039$ & 0.038 & 0.109 & 1454\\
\midrule
\multirow{7}{*}{GSM8K, Qwen2.5-7B} & $b{=}1$ & 67/659 $=0.102$ & 0.097 & 0.196 & 257\\
 & ES(2,5) & 46/659 $=0.070$ & 0.080 & 0.154 & 582\\
 & $b{=}3$ & 54/659 $=0.082$ & 0.085 & 0.171 & 767\\
 & ES(3,9) & 45/659 $=0.068$ & 0.073 & 0.152 & 909\\
 & $b{=}5$ & 46/659 $=0.070$ & 0.083 & 0.154 & 1277\\
 & $b{=}9$ & 44/659 $=0.067$ & 0.073 & 0.150 & 2298\\
 & $b{=}17$ & 41/659 $=0.062$ & 0.073 & 0.144 & 4338\\
\midrule
\multirow{4}{*}{MATH-500, gpt-4.1-mini} & $b{=}1$ & 79/250 $=0.316$ & 0.320 & 0.508 & 587\\
 & $b{=}3$ & 75/250 $=0.300$ & 0.304 & 0.491 & 1763\\
 & $b{=}5$ & 70/250 $=0.280$ & 0.296 & 0.470 & 2940\\
 & $b{=}9$ & 71/250 $=0.284$ & 0.300 & 0.474 & 5300\\
\midrule
\multirow{4}{*}{MATH-500, Qwen2.5-7B} & $b{=}1$ & 90/250 $=0.360$ & 0.352 & 0.553 & 570\\
 & $b{=}3$ & 89/250 $=0.356$ & 0.308 & 0.549 & 1713\\
 & $b{=}5$ & 82/250 $=0.328$ & 0.300 & 0.521 & 2847\\
 & $b{=}9$ & 75/250 $=0.300$ & 0.268 & 0.491 & 5132\\
\bottomrule
\end{tabular}

\end{table}

\begin{table}[!htbp]
\centering\small
\caption{Top eight configurations on the Qwen ($M=27$) and \texttt{gpt-4.1-mini} ($M=25$) GSM8K recipe grids ($n_{\mathrm{cal}}=250$, $\delta=0.05$).}\label{tab:grid}
\begin{tabular}{@{}llcccr@{}}
\headtop
\headstrut \textbf{grid} & \textbf{recipe (template/temperature/rule)} & \textbf{cal.\ risk} & \textbf{held-out} & \textbf{certificate} & \textbf{tokens}\\
\headbottom
Qwen, $M{=}27$ & P1/T1.0/$b{=}5$ & 0.056 & 0.072 & 0.198 & 755\\
 & P2/T1.0/$b{=}9$ & 0.056 & 0.056 & 0.198 & 1624\\
 & P0/T1.0/$b{=}9$ & 0.056 & 0.064 & 0.198 & 2337\\
 & P0/T1.0/$b{=}5$ & 0.060 & 0.068 & 0.204 & 1306\\
 & P1/T1.0/$b{=}9$ & 0.060 & 0.068 & 0.204 & 1355\\
 & P2/T0.7/$b{=}9$ & 0.060 & 0.060 & 0.204 & 1606\\
 & P0/T0.7/verifier bo3 & 0.064 & 0.076 & 0.211 & 840\\
 & P0/T0.7/ES(3,9) & 0.064 & 0.064 & 0.211 & 903\\
\midrule
gpt, $M{=}25$ & P1/T1.0/$b{=}9$ & 0.020 & 0.044 & 0.134 & 1224\\
 & P2/T1.0/$b{=}9$ & 0.020 & 0.048 & 0.134 & 1392\\
 & P0/T0.7/ES(3,9) & 0.024 & 0.056 & 0.142 & 527\\
 & P1/T1.0/$b{=}5$ & 0.024 & 0.040 & 0.142 & 677\\
 & P0/T0.7/$b{=}9$ & 0.024 & 0.056 & 0.142 & 1470\\
 & P2/T1.0/$b{=}3$ & 0.028 & 0.048 & 0.149 & 462\\
 & P1/T0.7/$b{=}5$ & 0.028 & 0.044 & 0.149 & 685\\
 & P2/T1.0/$b{=}5$ & 0.028 & 0.048 & 0.149 & 774\\
\bottomrule
\end{tabular}

\end{table}

\subsection{Pilot study and small-sample stability}\label{app:pilot}
\Cref{tab:pilot,tab:pilotsens,tab:pilotresamp} detail the 192-example \texttt{gpt-4.1-mini} GSM8K pilot ($96/96$ split), its anytime prefix trajectory from $n=32$ to $160$, and its selection distribution across 3{,}000 random subsets.

\begin{table}[!htbp]
\centering\small
\begin{minipage}[t]{0.49\linewidth}\centering
\captionof{table}{GSM8K pilot ($n=96$, $M=4$).}\label{tab:pilot}
\resizebox{\linewidth}{!}{\begin{tabular}{@{}lcccc@{}}
\headtop
\headstrut \textbf{budget} & \textbf{cal.\ risk} & \textbf{held-out risk} & \textbf{certificate} & \textbf{mean cal.\ tokens}\\
\headbottom
$b{=}1$ & 3/96 & 4/96 & 0.240 & 205\\
$b{=}3$ & 2/96 & 1/96 & 0.219 & 616\\
$b{=}5$ & 2/96 & 2/96 & 0.219 & 1024\\
$b{=}9$ & 2/96 & 2/96 & 0.219 & 1839\\
\bottomrule
\end{tabular}
}
\end{minipage}\hfill
\begin{minipage}[t]{0.49\linewidth}\centering
\captionof{table}{Anytime prefix trajectory on the pilot.}\label{tab:pilotsens}
\resizebox{\linewidth}{!}{\begin{tabular}{@{}rccccr@{}}
\headtop
\headstrut \textbf{$n_{\mathrm{cal}}$} & \textbf{selected $b$} & \textbf{cal.\ risk} & \textbf{held-out risk} & \textbf{certificate} & \textbf{mean tokens}\\
\headbottom
32 & 3 & 0.000 & 0.019 & 0.350 & 634\\
64 & 3 & 0.000 & 0.023 & 0.215 & 624\\
96 & 3 & 0.021 & 0.010 & 0.219 & 616\\
128 & 3 & 0.023 & 0.000 & 0.191 & 626\\
160 & 3 & 0.019 & 0.000 & 0.159 & 635\\
\bottomrule
\end{tabular}
}
\end{minipage}
\end{table}

\begin{table}[!htbp]
\centering\small
\caption{Selection stability of the pilot across 3{,}000 random calibration subsets (budgets $5$ and $9$ tie budget $3$ on calibration errors and yield to the token-cost tie-breaker).}\label{tab:pilotresamp}
\begin{tabular}{@{}rcccc@{}}
\headtop
\headstrut \textbf{$n_{\mathrm{cal}}$} & \textbf{$\P(\hat b=1)$} & \textbf{$\P(\hat b=3)$} & \textbf{mean held-out risk} & \textbf{median certificate}\\
\headbottom
32 & 0.482 & 0.518 & 0.028 & 0.350\\
64 & 0.198 & 0.802 & 0.022 & 0.264\\
96 & 0.065 & 0.935 & 0.018 & 0.193\\
128 & 0.013 & 0.987 & 0.017 & 0.174\\
160 & 0.001 & 0.999 & 0.015 & 0.159\\
\bottomrule
\end{tabular}

\end{table}

\subsection{Hierarchical versus flat priors on expanding sweeps}\label{app:loc}
\Cref{tab:loc} provides the exact numerical values plotted in \cref{fig:frontier}b (\cref{sec:grids}). Assigning $\pi_0=0.5$ to the default family $(\text{P0},0.7;\,\elocMdefault\text{ recipes})$ fixes the default winner's selection penalty at $\log2+\log\elocMdefault$ nats regardless of how broadly the five exploratory families expand, whereas a uniform prior charges $\log M$ across the entire pool.

\begin{table}[!htbp]
\centering\small
\caption{\textbf{Hierarchical priors insulate default-family bounds as exploratory sweeps expand} (GSM8K, Qwen, $n=250$, $\delta=0.05$; numerical values for \cref{fig:frontier}b). Rows $M\ge\elocMb$ append seeded trace-subset configurations to the exploratory families (\cref{rem:repairs}).}\label{tab:loc}
\begin{tabular}{@{}rccc@{}}
\headtop
\headstrut \textbf{recipes $M$} & \textbf{uniform: sel.\ cal.\,/\,held-out\,/\,bound} & \textbf{uniform: default bound} & \textbf{hierarchical}\\
\headbottom
\elocMa & \elocFlatCala\,/\,\elocFlatHelda\,/\,\elocFlatAudita & \elocFlatSamea & \elocLoc\\
\elocMb & \elocFlatCalb\,/\,\elocFlatHeldb\,/\,\elocFlatAuditb & \elocFlatSameb & \elocLoc\\
\elocMc & \elocFlatCalc\,/\,\elocFlatHeldc\,/\,\elocFlatAuditc & \elocFlatSamec & \elocLoc\\
\bottomrule
\end{tabular}
\end{table}

\subsection{Dyadic certification of uncapped and heavy-tailed costs}
\Cref{tab:envelope} applies \cref{cor:envelope} to realized token counts of Qwen's $b=9$ recipe on GSM8K ($n=\ncalQG$; empirical median \tokMedB{}, 99th percentile \tokPnnB{} tokens) using base scale $C_0=\cmaxZero$ without invoking the structural cap $9\times512=4{,}608$. Once $C_k$ crosses the bulk of the token distribution, the clipped-mean bound tightens rapidly while the exceedance certificate settles onto the zero-event floor $\approx0.036$. \Cref{tab:envelopeln} stress-tests the same bound on genuinely unbounded synthetic lognormal costs ($n=500$, median $\lnMedian$, mean $\lnMean$, 99th percentile $\lnPnn$, $C_0=256$): across $\lnReps$ independent replicates, all sixteen simultaneous clipped-mean and exceedance bounds achieve 100\% joint coverage of their exact analytical targets.

\begin{table}[!htbp]
\centering\small
\caption{Simultaneous dyadic cost certificates on realized Qwen GSM8K token counts ($C_0=\cmaxZero$, $n=\ncalQG$, $\delta=0.05$).}\label{tab:envelope}
\begin{tabular}{@{}rccc@{}}
\headtop
\headstrut \textbf{threshold $C_k$} & \textbf{penalty $\log[(k{+}1)(k{+}2)]+\log 2$} & \textbf{certified $\E\min\{C/C_k,1\}$} & \textbf{certified $\P\{C>C_k\}$}\\
\headbottom
512 & 1.39 & 1.000 & 1.000\\
1{,}024 & 2.48 & 1.000 & 1.000\\
2{,}048 & 3.18 & 0.979 & 0.732\\
4{,}096 & 3.69 & 0.688 & 0.054\\
8{,}192 & 4.09 & 0.410 & 0.036\\
16{,}384 & 4.43 & 0.251 & 0.036\\
32{,}768 & 4.72 & 0.161 & 0.037\\
65{,}536 & 4.97 & 0.110 & 0.037\\
\bottomrule
\end{tabular}

\end{table}

\begin{table}[!htbp]
\centering\small
\caption{Simultaneous dyadic cost certificates on unbounded lognormal costs ($n=500$, $C_0=256$, $\delta=0.05$), comparing one representative realization against exact population values.}\label{tab:envelopeln}
\begin{tabular}{@{}rccccc@{}}
\headtop
\headstrut \textbf{threshold $C_k$} & \textbf{penalty} & \textbf{certified $\E\min\{C/C_k,1\}$} & \textbf{true} & \textbf{certified $\P\{C>C_k\}$} & \textbf{true}\\
\headbottom
256 & 1.39 & 0.975 & 0.914 & 0.888 & 0.775\\
512 & 2.48 & 0.882 & 0.777 & 0.657 & 0.525\\
1{,}024 & 3.18 & 0.713 & 0.576 & 0.394 & 0.264\\
2{,}048 & 3.69 & 0.519 & 0.368 & 0.227 & 0.093\\
4{,}096 & 4.09 & 0.349 & 0.207 & 0.109 & 0.022\\
8{,}192 & 4.43 & 0.233 & 0.108 & 0.066 & 0.003\\
16{,}384 & 4.72 & 0.157 & 0.055 & 0.047 & 0.000\\
32{,}768 & 4.97 & 0.112 & 0.027 & 0.047 & 0.000\\
\bottomrule
\end{tabular}

\end{table}

\FloatBarrier
\subsection{Empirical winner entropy and code-length savings}\label{app:winners}
Across the 3{,}000 random $n=150$ calibration splits of \cref{sec:miscoverage}, \cref{tab:winners} reports the empirical Shannon entropy $\hat H=H(\hat\rho)$ of the winning recipe's selection distribution $\hat\rho$ alongside the uniform-prior redundancy $\log M-\hat H$ (\cref{thm:codes}(iii)). Because selection concentrates heavily on a handful of strong configurations, matching the prior to $\hat\rho$ saves up to $0.80$ nats over a uniform grid prior.

\begin{table}[!htbp]
\centering\small
\caption{Empirical Shannon entropy $\hat H$ (nats) of the selected winner across 3{,}000 random splits ($n=150$).}\label{tab:winners}
\begin{tabular}{@{}lrcccl@{}}
\headtop
\headstrut \textbf{sweep (GSM8K)} & \textbf{$M$} & \textbf{$\log M$} & \textbf{$\hat H$} & \textbf{$\log M-\hat H$} & \textbf{most frequent winners (share)}\\
\headbottom
gpt-4.1-mini, budgets & 4 & 1.386 & 1.204 & 0.182 & $b{=}3$ (0.44), $b{=}5$ (0.27)\\
Qwen, budgets & 5 & 1.609 & 1.315 & 0.294 & $b{=}9$ (0.37), $b{=}17$ (0.36)\\
gpt-4.1-mini, recipe grid & 25 & 3.219 & 2.630 & 0.589 & P1/T1.0/$b{=}3$ (0.15), P1/T1.0/$b{=}5$ (0.13)\\
Qwen, recipe grid & 27 & 3.296 & 2.499 & 0.797 & P2/T1.0/$b{=}9$ (0.23), P1/T1.0/$b{=}5$ (0.14)\\
\bottomrule
\end{tabular}

\end{table}

\FloatBarrier
\subsection{Controlled synthetic stress tests}\label{app:synthetic}
\begin{figure}[!htb]
\centering
\pgfplotstableread[row sep=crcr]{
p b1 b5 b17 b65\\
0.00 1.000000 1.000000 1.000000 1.000000\\
0.01 0.990000 0.999990 1.000000 1.000000\\
0.02 0.980000 0.999922 1.000000 1.000000\\
0.03 0.970000 0.999742 1.000000 1.000000\\
0.04 0.960000 0.999398 1.000000 1.000000\\
0.05 0.950000 0.998842 1.000000 1.000000\\
0.06 0.940000 0.998030 1.000000 1.000000\\
0.07 0.930000 0.996920 0.999999 1.000000\\
0.08 0.920000 0.995475 0.999998 1.000000\\
0.09 0.910000 0.993659 0.999995 1.000000\\
0.10 0.900000 0.991440 0.999989 1.000000\\
0.11 0.890000 0.988790 0.999975 1.000000\\
0.12 0.880000 0.985681 0.999950 1.000000\\
0.13 0.870000 0.982091 0.999904 1.000000\\
0.14 0.860000 0.978000 0.999828 1.000000\\
0.15 0.850000 0.973388 0.999705 1.000000\\
0.16 0.840000 0.968241 0.999514 1.000000\\
0.17 0.830000 0.962546 0.999229 1.000000\\
0.18 0.820000 0.956293 0.998814 1.000000\\
0.19 0.810000 0.949472 0.998228 1.000000\\
0.20 0.800000 0.942080 0.997419 1.000000\\
0.21 0.790000 0.934112 0.996327 1.000000\\
0.22 0.780000 0.925566 0.994883 1.000000\\
0.23 0.770000 0.916444 0.993009 0.999999\\
0.24 0.760000 0.906749 0.990618 0.999997\\
0.25 0.750000 0.896484 0.987615 0.999993\\
0.26 0.740000 0.885658 0.983899 0.999983\\
0.27 0.730000 0.874277 0.979362 0.999960\\
0.28 0.720000 0.862352 0.973895 0.999913\\
0.29 0.710000 0.849895 0.967386 0.999818\\
0.30 0.700000 0.836920 0.959723 0.999637\\
0.31 0.690000 0.823441 0.950800 0.999304\\
0.32 0.680000 0.809474 0.940515 0.998717\\
0.33 0.670000 0.795037 0.928776 0.997721\\
0.34 0.660000 0.780149 0.915500 0.996095\\
0.35 0.650000 0.764831 0.900621 0.993532\\
0.36 0.640000 0.749103 0.884088 0.989628\\
0.37 0.630000 0.732988 0.865866 0.983873\\
0.38 0.620000 0.716509 0.845946 0.975656\\
0.39 0.610000 0.699692 0.824334 0.964281\\
0.40 0.600000 0.682560 0.801064 0.949000\\
0.41 0.590000 0.665140 0.776189 0.929060\\
0.42 0.580000 0.647460 0.749790 0.903772\\
0.43 0.570000 0.629545 0.721968 0.872583\\
0.44 0.560000 0.611425 0.692844 0.835152\\
0.45 0.550000 0.593127 0.662564 0.791423\\
0.46 0.540000 0.574681 0.631290 0.741671\\
0.47 0.530000 0.556115 0.599200 0.686530\\
0.48 0.520000 0.537460 0.566485 0.626979\\
0.49 0.510000 0.518745 0.533349 0.564301\\
0.50 0.500000 0.500000 0.500000 0.500000\\
0.51 0.490000 0.481255 0.466651 0.435699\\
0.52 0.480000 0.462540 0.433515 0.373021\\
0.53 0.470000 0.443885 0.400800 0.313470\\
0.54 0.460000 0.425319 0.368710 0.258329\\
0.55 0.450000 0.406873 0.337436 0.208577\\
0.56 0.440000 0.388575 0.307156 0.164848\\
0.57 0.430000 0.370455 0.278032 0.127417\\
0.58 0.420000 0.352540 0.250210 0.096228\\
0.59 0.410000 0.334860 0.223811 0.070940\\
0.60 0.400000 0.317440 0.198936 0.051000\\
0.61 0.390000 0.300308 0.175666 0.035719\\
0.62 0.380000 0.283491 0.154054 0.024344\\
0.63 0.370000 0.267012 0.134134 0.016127\\
0.64 0.360000 0.250897 0.115912 0.010372\\
0.65 0.350000 0.235169 0.099379 0.006468\\
0.66 0.340000 0.219851 0.084500 0.003905\\
0.67 0.330000 0.204963 0.071224 0.002279\\
0.68 0.320000 0.190526 0.059485 0.001283\\
0.69 0.310000 0.176559 0.049200 0.000696\\
0.70 0.300000 0.163080 0.040277 0.000363\\
0.71 0.290000 0.150105 0.032614 0.000182\\
0.72 0.280000 0.137648 0.026105 0.000087\\
0.73 0.270000 0.125723 0.020638 0.000040\\
0.74 0.260000 0.114342 0.016101 0.000017\\
0.75 0.250000 0.103516 0.012385 0.000007\\
0.76 0.240000 0.093251 0.009382 0.000003\\
0.77 0.230000 0.083556 0.006991 0.000001\\
0.78 0.220000 0.074434 0.005117 0.000000\\
0.79 0.210000 0.065888 0.003673 0.000000\\
0.80 0.200000 0.057920 0.002581 0.000000\\
0.81 0.190000 0.050528 0.001772 0.000000\\
0.82 0.180000 0.043707 0.001186 0.000000\\
0.83 0.170000 0.037454 0.000771 0.000000\\
0.84 0.160000 0.031759 0.000486 0.000000\\
0.85 0.150000 0.026612 0.000295 0.000000\\
0.86 0.140000 0.022000 0.000172 0.000000\\
0.87 0.130000 0.017909 0.000096 0.000000\\
0.88 0.120000 0.014319 0.000050 0.000000\\
0.89 0.110000 0.011210 0.000025 0.000000\\
0.90 0.100000 0.008560 0.000011 0.000000\\
0.91 0.090000 0.006341 0.000005 0.000000\\
0.92 0.080000 0.004525 0.000002 0.000000\\
0.93 0.070000 0.003080 0.000001 0.000000\\
0.94 0.060000 0.001970 0.000000 0.000000\\
0.95 0.050000 0.001158 0.000000 0.000000\\
0.96 0.040000 0.000602 0.000000 0.000000\\
0.97 0.030000 0.000258 0.000000 0.000000\\
0.98 0.020000 0.000078 0.000000 0.000000\\
0.99 0.010000 0.000010 0.000000 0.000000\\
1.00 0.000000 0.000000 0.000000 0.000000\\
}{\tblSynthVote}%
\pgfplotstableread[row sep=crcr]{
n emp true upper oracle\\
64 0.2559 0.2974 0.6026 0.2824\\
128 0.2706 0.2920 0.5292 0.2824\\
256 0.2765 0.2887 0.4666 0.2824\\
512 0.2781 0.2865 0.4166 0.2824\\
1024 0.2802 0.2845 0.3808 0.2824\\
}{\tblSynthNsweep}%
\pgfplotstableread[row sep=crcr]{
m optimism guide fixedcov pbcov\\
2 0.0180 0.0520 0.9860 1.0000\\
4 0.0325 0.0736 0.9695 1.0000\\
8 0.0442 0.0901 0.9433 1.0000\\
16 0.0551 0.1041 0.8895 1.0000\\
32 0.0646 0.1164 0.7893 1.0000\\
64 0.0733 0.1275 0.6203 1.0000\\
128 0.0810 0.1377 0.3880 1.0000\\
}{\tblSynthMsweep}%
\begin{tikzpicture}[figbase]
\begin{axis}[paperplot, name=sa, scale only axis, width=88pt, height=78pt,
  xmin=0, xmax=1, ymin=0, ymax=1, xtick={0,0.5,1}, xticklabels={0,.5,1}, ytick={0,0.5,1}, yticklabels={0,.5,1},
  xlabel={single-trace correctness $p$}, ylabel={majority-vote risk},
  title={\ptitle{a}{Voting helps only if $p>\frac12$}}]
  \fill[WarmFill] (axis cs:0,0) rectangle (axis cs:0.5,1);
  \fill[CoolFill] (axis cs:0.5,0) rectangle (axis cs:1,1);
  \addplot[MidGray, line width=0.7pt] table[x=p, y=b1] {\tblSynthVote};
  \addplot[CoolMid, line width=0.9pt] table[x=p, y=b5] {\tblSynthVote};
  \addplot[CoolInk, line width=1.0pt] table[x=p, y=b17] {\tblSynthVote};
  \addplot[Ink, line width=1.1pt] table[x=p, y=b65] {\tblSynthVote};
  \draw[->, Ink!75, line width=0.5pt] (axis cs:0.66,0.47) -- (axis cs:0.66,0.035);
  \node[font=\tiny, text=Ink, anchor=west, inner sep=1pt, align=left] at (axis cs:0.68,0.40) {votes\\$b{=}1,5,$\\$17,65$};
\end{axis}
\begin{axis}[ygridplot, name=sb, at={(sa.south east)}, anchor=south west, xshift=46pt,
  scale only axis, width=88pt, height=78pt, xmode=log, log basis x=2,
  xmin=50, xmax=1300, ymin=0.2, ymax=0.86, xtick={64,256,1024}, xticklabels={64,256,1024},
  ytick={0.2,0.3,0.4,0.5,0.6}, yticklabels={.2,.3,.4,.5,.6},
  xlabel={calibration examples $n$}, ylabel={risk},
  title={\ptitle{b}{The certificate tightens}},
  legend style={at={(1.0,1.0)}, anchor=north east, font=\tiny, row sep=-0.8pt},
  legend image post style={xscale=0.62}]
  \addplot[CoolInk, line width=1.0pt, mark=*, mark size=1.3pt, mark options={draw=white, line width=0.4pt}] table[x=n, y=upper] {\tblSynthNsweep};
  \addplot[Ink, line width=0.7pt, mark=*, mark size=1.2pt, mark options={fill=white, draw=Ink, line width=0.5pt}] table[x=n, y=true] {\tblSynthNsweep};
  \addplot[WarmInk, line width=0.7pt, mark=square*, mark size=1.1pt, mark options={draw=white, line width=0.3pt}] table[x=n, y=emp] {\tblSynthNsweep};
  \addplot[MidGray, line width=0.6pt, dash pattern=on 2.2pt off 1.6pt] table[x=n, y=oracle] {\tblSynthNsweep};
  \legend{certificate, selected: true, selected: empirical, best recipe}
\end{axis}
\begin{axis}[ygridplot, name=sc, at={(sb.south east)}, anchor=south west, xshift=46pt,
  scale only axis, width=88pt, height=78pt, xmode=log, log basis x=2,
  xmin=1.7, xmax=150, ymin=0.3, ymax=1.05, xtick={2,8,32,128}, xticklabels={2,8,32,128},
  ytick={0.4,0.6,0.8,1.0}, yticklabels={.4,.6,.8,1},
  xlabel={recipes $M$}, ylabel={coverage},
  title={\ptitle{c}{Fixed bounds break}}]
  \addplot[CoolInk, line width=1.0pt, mark=*, mark size=1.3pt, mark options={draw=white, line width=0.4pt}] table[x=m, y=pbcov] {\tblSynthMsweep};
  \addplot[WarmInk, line width=1.0pt, mark=square*, mark size=1.2pt, mark options={draw=white, line width=0.3pt}] table[x=m, y=fixedcov] {\tblSynthMsweep};
  \node[font=\tiny, text=Ink, anchor=north east, inner sep=1pt] at (axis cs:128,0.975) {certificate};
  \node[font=\tiny, text=Ink, anchor=east, inner sep=1pt, align=right] at (axis cs:40,0.55) {fixed-recipe\\Hoeffding};
\end{axis}
\end{tikzpicture}%
\caption{\textbf{Synthetic ground-truth stress tests.} \textbf{(a)}~Majority voting reduces error for $p>\frac12$ and amplifies error for $p<\frac12$. \textbf{(b)}~Across 800 adaptive sweeps per sample size $n$, the PAC-Bayes certificate converges smoothly toward the selected recipe's exact risk while maintaining 100\% coverage. \textbf{(c)}~Across $M$ exchangeable fair-coin recipes ($n=256$), selection optimism erodes fixed-recipe Hoeffding coverage to \synthFixedCovMax{} by $M=\synthMmax$, whereas the PAC-Bayes certificate holds throughout.}\label{fig:synthetic}
\end{figure}
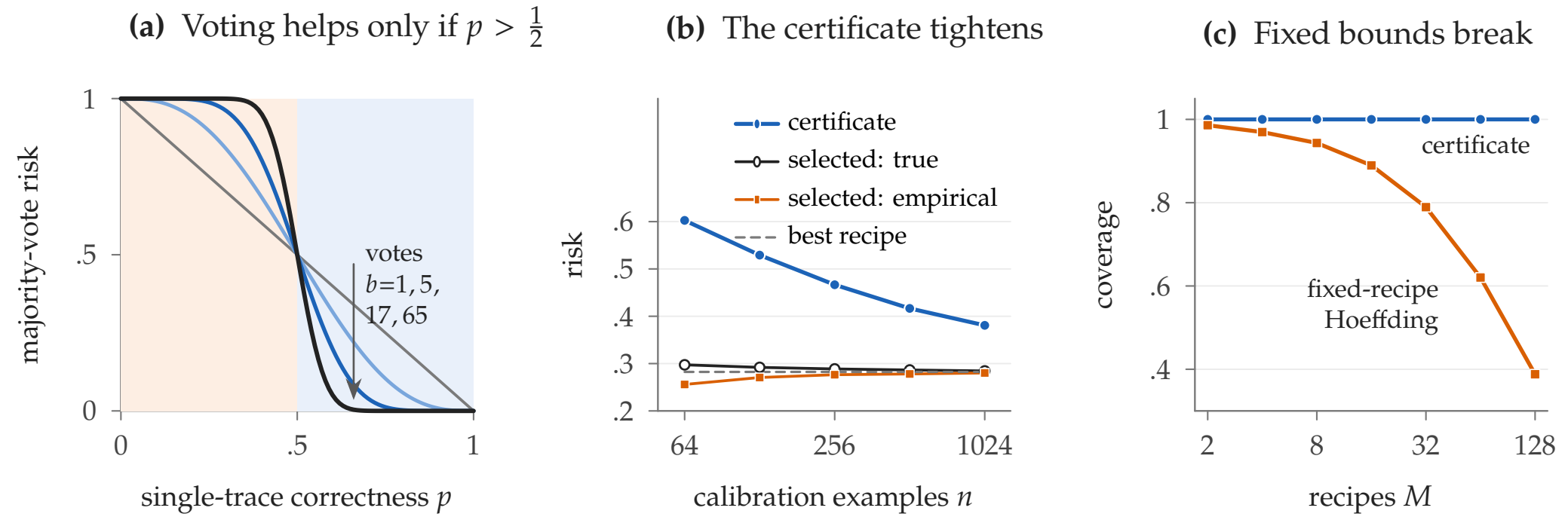

To probe exact population coverage where ground-truth risks are known analytically, we draw latent single-trace correctness probabilities $p_X\sim0.7\,\mathrm{Beta}(8,2)+0.3\,\mathrm{Beta}(2,5)$ (so $30\%$ of tasks lie in the negative-margin regime $p_X<\frac12$ where voting hurts; \cref{fig:synthetic}a), sweep majority-vote budgets $\{1,3,5,9,17,33,65\}$, and evaluate exact risks on $200{,}000$ fresh tasks (\cref{fig:synthetic}b, \cref{tab:synthn}). A second experiment fixes $M\in\{2,4,\ldots,128\}$ exchangeable recipes of exact risk $\frac12$ at $n=256$ to isolate \cref{prop:lower} (\cref{fig:synthetic}c, \cref{tab:synthm}): as $M$ grows, $\sqrt{\log M/n}$ selection optimism drives fixed-recipe Hoeffding coverage down from $\synthFixedCovMin$ to $\synthFixedCovMax$, whereas \cref{thm:main} maintains $1.000$ coverage across every sweep size.

\begin{table}[!htb]
\centering\small
\caption{Sample-size stress test on the Beta-mixture voting environment (means across 800 adaptive sweeps per $n$).}\label{tab:synthn}
\begin{tabular}{@{}rcccccc@{}}
\headtop
\headstrut \textbf{$n$} & \textbf{sel.\ empirical} & \textbf{sel.\ true} & \textbf{sel.\ certificate} & \textbf{oracle true} & \textbf{modal $\hat b$} & \textbf{coverage}\\
\headbottom
64 & 0.256 & 0.297 & 0.603 & 0.282 & 17 & 1.000\\
128 & 0.271 & 0.292 & 0.529 & 0.282 & 33 & 1.000\\
256 & 0.276 & 0.289 & 0.467 & 0.282 & 33 & 1.000\\
512 & 0.278 & 0.286 & 0.417 & 0.282 & 33 & 1.000\\
1024 & 0.280 & 0.285 & 0.381 & 0.282 & 65 & 1.000\\
\bottomrule
\end{tabular}

\end{table}

\begin{table}[!htb]
\centering\small
\caption{Sweep-breadth stress test across $M$ exchangeable recipes of exact risk $\frac12$ ($n=256$, $\delta=0.05$).}\label{tab:synthm}
\begin{tabular}{@{}rcccccc@{}}
\headtop
\headstrut \textbf{$M$} & \textbf{sel.\ empirical} & \textbf{optimism} & \textbf{fixed upper} & \textbf{fixed cov.} & \textbf{certificate} & \textbf{cert.\ cov.}\\
\headbottom
2 & 0.482 & 0.018 & 0.558 & 0.986 & 0.666 & 1.000\\
4 & 0.467 & 0.033 & 0.544 & 0.970 & 0.655 & 1.000\\
8 & 0.456 & 0.044 & 0.532 & 0.943 & 0.648 & 1.000\\
16 & 0.445 & 0.055 & 0.521 & 0.889 & 0.641 & 1.000\\
32 & 0.435 & 0.065 & 0.512 & 0.789 & 0.635 & 1.000\\
64 & 0.427 & 0.073 & 0.503 & 0.620 & 0.630 & 1.000\\
128 & 0.419 & 0.081 & 0.495 & 0.388 & 0.626 & 1.000\\
\bottomrule
\end{tabular}

\end{table}

\FloatBarrier
\section{Scope and broader impact}\label{app:limits}

\paragraph{Statistical boundary conditions.}
Like all calibration-based guarantees, our certificates hold with respect to the data-generating distribution $\cD$ from which the i.i.d.\ calibration problems are drawn and the specific answer-evaluation pipeline (including parsers or automated verifiers) embedded in the loss $\ell_a$. Significant distribution shift between benchmark calibration prompts and live deployment traffic, or systematic grading errors in reference labels, must be addressed separately through domain-matched sampling or human-audited labels.

\paragraph{Broader impact.}
Test-time compute scaling has made inference pipelines vastly more capable, but also far easier to overfit to small validation splits through repeated prompt, verifier, and budget tuning. Providing lightweight, anytime-valid certificates for both error rate and token cost replaces informal post-hoc error bars with auditable statistical guarantees, supporting transparent risk--compute reporting in safety- and cost-sensitive deployments.

\end{document}